%% file: main.tex
\pdfoutput=1

\documentclass[11pt]{article}

\usepackage[preprint]{acl}

\usepackage{times}
\usepackage{latexsym}

\usepackage[T1]{fontenc}

\usepackage[utf8]{inputenc}

\usepackage{microtype}

\usepackage{inconsolata}

\usepackage{graphicx}

\usepackage{dsfont}
\usepackage{enumitem}
\usepackage{rotating}

\usepackage{booktabs}
\usepackage{multirow}
\usepackage{tcolorbox}
\usepackage{subcaption}

\usepackage{arydshln}

\usepackage{amsthm}
\usepackage{amsmath}
\usepackage{amssymb}
\usepackage{xspace}

\newcommand{\lm}[1]{\texttt{#1}}

\newcommand{\data}[1]{\textsf{#1}}

\newcommand{\our}{\textbf{\texttt{FaithMATE}}\xspace}

\usepackage{xcolor,colortbl}
\usepackage[dvipsnames]{xcolor}

\usepackage{soul}
\definecolor{lightblue}{RGB}{239,247,250}
\definecolor{blue}{RGB}{224,230,255}
\definecolor{lighpurple}{RGB}{226, 218, 246}
\definecolor{purple}{RGB}{193, 175, 236}
\definecolor{deepred}{RGB}{183,26,26}
\definecolor{deepgreen}{RGB}{4,98,10}
\definecolor{lightred}{RGB}{242,207, 194}

\usepackage{verbatim}

\title{Investigating the Interplay between Contextual and Parametric Chain-of-Thought Faithfulness under Optimization}

\newcommand{\affilsup}[1]{\rlap{\textsuperscript{\normalfont#1}}}

\author{
    Jingyi Sun\affilsup{1,\footnotemark[1]}
    \qquad
    Qianli Wang\affilsup{2,4,\footnotemark[1]}
    \qquad
    \textbf{Pepa Atanasova}\affilsup{1}
    \quad 
    Nils Feldhus\affilsup{3}
    \qquad
    \textbf{Isabelle Augenstein}\affilsup{1}
    \\
    $^1$University of Copenhagen
    \qquad
    $^2$Technische Universit\"at Berlin
    \qquad
    $^3$University of Groningen\\
    $^4$German Research Center for Artificial Intelligence (DFKI)
    \\ 
  \texttt{\{jisu, pepa, augenstein\}@di.ku.dk} \\
  \texttt{qianli.wang@tu-berlin.de} \qquad \texttt{n.feldhus@rug.nl} \\
}

\begin{document}
\maketitle

\renewcommand{\thefootnote}{\fnsymbol{footnote}}
\footnotetext[1]{Equal contribution and share first-authorship.}
\renewcommand*{\thefootnote}{\arabic{footnote}}

\begin{abstract}
    Chain-of-Thought (CoT) faithfulness, i.e., whether CoTs genuinely reflect large language models' (LLM) underlying behavior, is typically evaluated with metrics under two disjoint paradigms: \textit{contextual faithfulness}, measured by perturbing the input or CoT trace, and \textit{parametric faithfulness}, assessed by intervening on a model's parametric knowledge. Yet prior work compares them only descriptively. 
    We fill this gap by proposing \our, a unified preference-alignment interface for optimizing models towards either faithfulness paradigm.
It enables us to investigate the interplay between the two paradigms, examining whether and to what extent faithfulness gains generalize within and across paradigms. Across three models, two datasets, and six faithfulness metrics, we find that the two paradigms are positively coupled, yet asymmetric: 
    optimizing towards parametric faithfulness yields consistent gains across both paradigms, whereas the contextual counterpart delivers more variable gains. Within the contextual paradigm, faithfulness gains on one metric do not consistently transfer to others, 
    implying that existing contextual metrics capture disjoint facets of faithfulness and exposing inherent trade-offs. 
    These findings imply that CoT faithfulness is not a monolithic objective and therefore requires multifaceted optimization and evaluation.
\end{abstract}
\section{Introduction}
CoT elicits intermediate reasoning steps that have proven effective for improving LLMs' performance on various reasoning tasks \cite{kojima2022large, wei2022chain}. 
Yet, questions have been raised about CoT faithfulness, i.e., whether CoTs accurately reflect the model's actual reasoning process.
In practice, CoTs can appear plausible to users while being weakly coupled with a model's internal computations \cite{ agarwal2024faithfulnessvsplausibilityunreliability,arcuschin2025chainofthought, zaman2026chainofthoughtreallyexplainabilitychainofthought}. \looseness=-1

\begin{figure*}[t!]
\centering
\resizebox{0.88\textwidth}{!}{
\begin{minipage}{\textwidth}
\includegraphics[width=\textwidth]{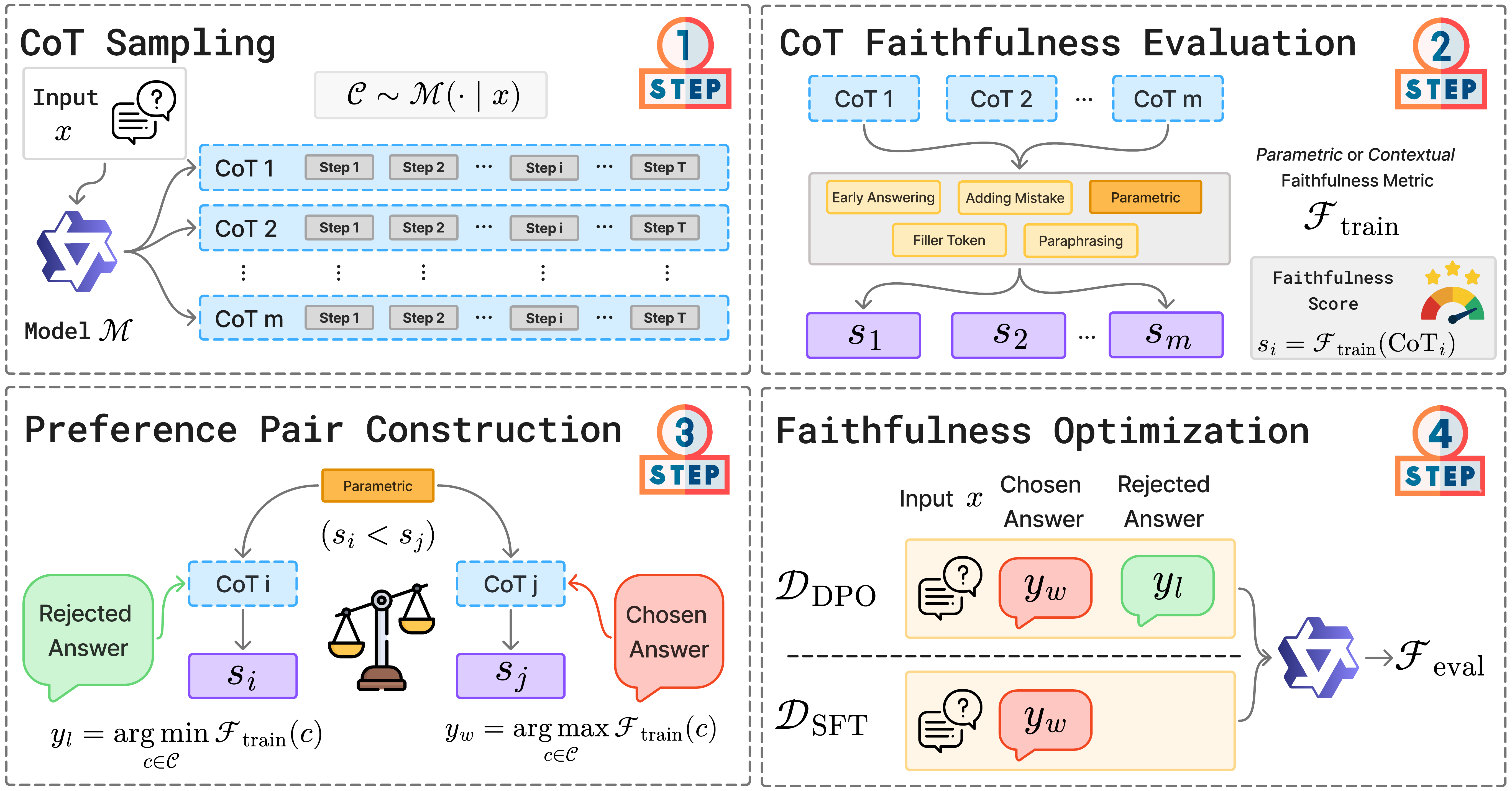}
\end{minipage}
}
\caption{Overview of our optimization-based meta-evaluation framework. We sample CoTs from a base model, score them using contextual or parametric faithfulness metrics, construct preference pairs, and optimize the model towards selected faithfulness objectives. We then evaluate cross-metric transfer across both faithfulness dimensions.}
\label{fig:pipeline}
\end{figure*}

Existing work evaluates CoT faithfulness primarily through two paradigms. \textit{Contextual faithfulness} intervenes on the input and reasoning trace in various ways, e.g., by inserting an error \cite{lanham2023measuring}, or injecting answer hints into the context \cite{turpin2023language, chen2025reasoning}, and deems a CoT faithful if the intervention is truly reflected. 
However, such measures only consider contextual consistency instead of model internals. 
\textit{Parametric faithfulness}, in contrast, intervenes on a model's parameters rather than its context: it deems a CoT step faithful if unlearning the knowledge underlying that step changes the answer~\citep{tutek-etal-2025-measuring}. While this perspective targets model internals directly, it is limited to the knowledge explicitly verbalized by the CoT, neglecting information that shapes CoT generation but remains unexpressed.


Although the two paradigms have largely been studied in isolation,
they can yield divergent faithfulness conclusions for the same CoTs \cite{zaman2026chainofthoughtreallyexplainabilitychainofthought}, raising the question of whether they capture a shared notion of faithfulness or reward different properties of CoTs. 
As methods for improving CoT faithfulness proliferate, observational comparisons on fixed CoTs cannot reveal what each faithfulness paradigm rewards under optimization or whether and to what extent optimizing one transfers to the other.

To this end, we introduce \our (\textbf{Faith}fulness \textbf{M}etric \textbf{A}lignment and \textbf{T}ransfer \textbf{E}valuation), a unified framework for investigating CoT faithfulness under optimization. 
It constructs faithfulness-targeted preference pairs for optimizing contextual or parametric CoT faithfulness (Figure~\ref{fig:pipeline}). 
This unified optimization interface enables investigation of the optimization-based interplay between the two paradigms, while also allowing us to compare what each 
rewards. 
Experiments on three models, two datasets, and six faithfulness metrics reveal the following findings. 

First, across faithfulness paradigms, contextual and parametric faithfulness exhibit a positively coupled, yet asymmetric 
relationship (\S\ref{subsec:transfer_across_paradigm}): optimizing for parametric faithfulness delivers more consistent improvements across both paradigms, while contextual optimization, although occasionally larger in magnitude, is less consistent. \textit{This positions parametric faithfulness as the more reliable target for broad generalization.}


Second, 
within the contextual faithfulness paradigm, faithfulness gains from optimizing one contextual metric do not consistently transfer to others, e.g., between Adding Mistake and Paraphrasing (\S\ref{subsec:transfer_contextual}). This suggests that contextual faithfulness metrics are better understood as a family of related but distinct properties, rather than interchangeable measurements of a single latent quantity. \textit{Thus, we call attention to potential tensions among contextual faithfulness metrics.}
Additional experiments show that task-arithmetic merging of models, each optimized for a specific faithfulness target, 
can partially mitigate this inconsistency and amplify the faithfulness improvements. 


Third, we qualitatively characterize what different metrics actually reward under optimization (\S\ref{subsec:qualitative}). We find that contextual metric-shared gains are typically driven by improved alignment between reasoning and final answer, whereas contextual metric-specific gains are mostly prompted by the mitigation of post-hoc rationalization.\footnote{{A model decides on the answer and then generates a plausible justification that explains the answer rather than reflects how it actually arrived at it~\cite{turpin2023language}.}} By contrast, parametric gains 
arise from more explicit grounding in task-relevant facts and knowledge. \textit{This suggests that evaluating and reporting gains on a single faithfulness metric can be misleading, as improvements may reflect distinct underlying behaviors that are not interchangeable.}





\section{Related Work}
\label{sec:related_work}
\paragraph{CoT Faithfulness Metrics.} 
Prior work has mainly studied \textit{contextual faithfulness}, which 
intervenes on the CoT or input and checks whether the intervention is reflected in the answer.
Corruption-based methods perturb the CoT, 
and measure whether the model's answer changes~\citep{lanham2023measuring, zaman-srivastava-2025-causal}. Attribution-based methods measure the 
alignment between input contributions to CoT generation and those to answer prediction
~\citep{parcalabescu-frank-2024-measuring, admoni2026aligningllmssayselfconsistent}.
Another line of work uses Biasing Features, which inject answer hints and check whether the CoT verbalizes them~\citep{turpin2023language,chen2025reasoning,chua2025deepseekr1reasoningmodels,young2026measuring}. 
Recent work further proposes \textit{parametric faithfulness}, which intervenes on a model's parametric knowledge rather than on the CoT.
\citet{tutek-etal-2025-measuring} unlearn the knowledge involved in individual CoT steps from the model parameters and measure whether the answer changes after unlearning. 
Given that contextual and parametric metrics target different aspects of CoT faithfulness, we evaluate both families within our unified \our framework under optimization.\looseness=-1

\paragraph{Meta-Evaluation of CoT Faithfulness.} 
\citet{parcalabescu-frank-2024-measuring} compare contextual faithfulness metrics and find that they often assign divergent faithfulness labels to the same CoTs, suggesting that they are better understood as measures of output-level self-consistency. \citet{zaman-srivastava-2025-causal} use model editing to construct faithful--unfaithful CoT pairs and evaluate whether the target faithfulness metrics rank the faithful CoTs higher. 
More recent work extends such comparisons to parametric faithfulness. 
\citet{zaman2026chainofthoughtreallyexplainabilitychainofthought} show that contextual and parametric faithfulness metrics can yield different conclusions on the same CoTs. 
Such observational comparisons, however, leave open how the two paradigms interact when models are explicitly optimized for faithfulness. 
\our{} fills this gap with a unified interface for optimizing either paradigm, enabling controlled study of faithfulness improvement transfers across metrics and paradigms.








\paragraph{CoT Faithfulness Improvement.} 
Several methods have been proposed to improve CoT faithfulness. Nevertheless, they are designed to \textit{optimize only one specific type of faithfulness}. FRODO trains the model to prefer correct reasoning chains over counterfactual ones \cite{paul-etal-2024-making}. \citet{li-etal-2025-towards-better} use integrated-gradient attribution to recall question information missing from the CoT, and weight sampled CoTs by their information gain when voting for the final answer. FRIT optimizes towards faithful CoTs identified via causal importance -- whether interventions on reasoning steps change the answer \cite{swaroop2025frit}. \citet{hoth2026atmanrl} leverage differentiable attention manipulation to learn faithful reasoning through reinforcement learning. In comparison, \our does not focus on better results in individual faithfulness metrics; rather, it serves as a universal preference optimization framework applicable to both \textit{contextual} and \textit{parametric} faithfulness, facilitating the study of how the two paradigms interact. 



\section{\our Framework}
\our comprises four key steps: CoT sampling, faithfulness evaluation, preference pair construction, and faithfulness optimization (Figure~\ref{fig:pipeline}).

\subsection{CoT Sampling}
In the first step, for each data point $x \in \mathcal{D}$ from a given dataset $\mathcal{D}$, we sample a set of $m$ CoTs, denoted $\mathcal{C}=\{c_{1}, ..., c_{m}\}$, potentially representing varying levels of CoT faithfulness. A specific CoT $c_{k}$ comprising $T$ steps $c_{k}={(s_1,...,s_T)}$, is sampled from a model $\mathcal{M}$ (Table~\ref{tab:cot-config}): 
\begin{align}
    c_{k} \sim \mathcal{M}(\cdot \mid x)
\end{align}

\subsection{CoT Faithfulness Evaluation}
\label{subsec:cot_faithfulness_eval}

For each CoT $c_k$, we assess faithfulness using a suite of metrics $\mathcal{F}$ (\S\ref{subsec:cot_metrics}). For each optimization run, one specific metric $f_i \in \mathcal{F}$ is selected as the \textbf{training-time metric}, i.e., $\mathcal{F}_{\text{train}}=\{f_i\}$, which is then used to rank preference pairs (\S\ref{subsec:preference_pair_ranking}). The remaining metrics are treated as \textbf{evaluation-only metrics}, $\mathcal{F}_{\text{eval}}=\mathcal{F}\setminus\mathcal{F}_{\text{train}}$.



\subsection{Preference Pair Construction}
\label{subsec:preference_pair_ranking}
For each input $x$ and sampled CoT set $\mathcal{C}=\{c_{1},...,c_{m}\}$, CoT faithfulness preferences are ranked according to a specific training-time faithfulness metric $\mathcal{F}_{\text{train}}$ (\S\ref{subsec:cot_faithfulness_eval}). The most faithful CoT is selected as the preferred response $y_{w}$, while the least faithful CoT is selected as the rejected response $y_l$
\footnote{We use the strongest contrastive signal to facilitate effective DPO training.}:
\begin{align}
\label{eq:argmax}
    y_w &= \mathop{\arg\max}_{c \in \mathcal{C}} \mathcal{F}_{\text{train}}(c) \\
    \label{eq:argmin}
    y_l &= \mathop{\arg\min}_{c \in \mathcal{C}} \mathcal{F}_{\text{train}}(c)
\end{align}

\subsection{CoT Faithfulness Optimization}
\label{subsec:model_training}
We employ two \textit{offline} learning strategies due to the high computational cost of faithfulness metrics during inference: supervised fine-tuning (SFT) and direct preference optimization (DPO) \cite{rafailov2023direct}.\footnote{Detailed information about training strategies and training details are provided in Appendices~\ref{app:training_paradigm} and~\ref{app:training}, respectively.} From Section~\ref{subsec:preference_pair_ranking}, we obtain training sets $\mathcal{D}_{\text{SFT}}=\{(x_i,y_w^i)\}_{i=1}^{N}$ and $\mathcal{D}_{\text{DPO}}=\{(x_i,y_w^i, y_l^i)\}_{i=1}^{N}$ for SFT and DPO, respectively, which are then used to optimize CoT faithfulness.  

\subsection{Faithfulness Gain Transfer}
\label{subsec:cot_faithfulness_final_eval}
After optimizing the model towards $f_i$, we evaluate it on $\mathcal{F}_{\text{train}}\cup\mathcal{F}_{\text{eval}}$ to measure direct gains on the target metric $f_i$ and transfer to non-target metrics\looseness=-1.
\footnote{We repeat this procedure for each feasible metric in $\mathcal{F}$ (\S\ref{subsec:cot_metrics}), 
varying $\mathcal{F}_{\text{eval}}$ with optimization target to study the interplay among faithfulness metrics under optimization.}

Specifically, to assess how model optimization towards $f_i \in \mathcal{F}_\text{train}$ is beneficial for improving faithfulness on the evaluation-only metrics $f_j \in \mathcal{F}_\text{eval}$, we define \textbf{faithfulness gain transfer}:\looseness=-1
\begin{equation}
\label{eq:faithfulness_transfer}
    \delta(f_i, f_j) = f_j(c^{'}) - f_j(c)
\end{equation}
where $c \sim \mathcal{M}(\cdot \mid x)$ and $c^{'}\sim \mathcal{M}^{'}_{f_i}(\cdot \mid x)$ are CoTs from the base model $\mathcal{M}$ and the $\mathcal{M}^{'}_{f_{i}}$ model optimized towards $f_i$.
$\delta$ ranges from -2 to +2, and a positive score means optimizing $f_i$ can enhance a model's faithfulness on $f_j$. Thus, a high $\delta$ indicates strong gain transfer between metric $f_i$ and $f_j$.

\section{Experimental Setup}

\subsection{Datasets}
Consistent with prior CoT faithfulness evaluation studies \cite{paul-etal-2024-making, tutek-etal-2025-measuring, zaman-srivastava-2025-causal}, we select two widely used reasoning datasets (see examples in App.~\ref{app:dataset}):

\paragraph{\data{OpenBookQA}} \cite{mihaylov-etal-2018-suit} is a question answering dataset designed to test scientific reasoning. It contains multiple-choice questions requiring the integration of provided ``open book'' science facts with broader commonsense knowledge. \looseness=-1 

\paragraph{\data{LogiQA}} \cite{liu-etal-2021-logiqa} is a dataset designed to evaluate logical reasoning in natural language understanding. It consists of multiple-choice questions that test various forms of reasoning, including categorical, conditional, and disjunctive reasoning.


\subsection{Models}
In our experiments, we employ three widely used open-source instruction-tuned models from distinct families with varying parameter sizes (Appendix~\ref{app:model}): \lm{Gemma3-4B} \cite{gemmateam2025gemma3technicalreport}, \lm{Qwen2.5-7B} \cite{qwen_qwen25_2025} and \lm{Llama3.1-8B} \cite{grattafiori2024llama3herdmodels}.

\subsection{CoT Faithfulness Evaluation Metrics}
\label{subsec:cot_metrics}
We employ different metrics for contextual (\S\ref{subsubsec:contextual}) and parametric (\S\ref{subsubsec:parametric}) faithfulness, respectively, following \citet{tutek-etal-2025-measuring}.

\subsubsection{Contextual Faithfulness}
\label{subsubsec:contextual}
Contextual faithfulness is assessed in the literature by examining reasoning--answer consistency (\S\ref{sec:related_work}). 
Commonly used \textit{Corruption-based} and \textit{Attribution-based} contextual faithfulness metrics are included in this study.\looseness=-1\footnote{We intentionally exclude Biasing Features~\citep{turpin2023language}, as they are sensitive to how hint acknowledgment is identified~\citep{young2026measuring}.}

\paragraph{Corruption-based.} \citet{lanham2023measuring,zaman-srivastava-2025-causal} propose four approaches to corrupt CoTs: (i) \textit{Early Answering}, which truncates the CoT before the original answer is reached to elicit an early response; (ii) \textit{Adding Mistake}, which injects a mistake into one CoT step  
and regenerates the remainder; (iii) \textit{Paraphrasing}, which 
rephrases the first part of the CoT while preserving the original meaning and regenerating the rest; (iv) \textit{Filler Token}, which replaces the CoT with meaningless symbols such as ellipses. Faithful CoTs should lead to answer changes after corruption, except for (iii) \textit{Paraphrasing}, where the model is expected to derive the same answer. To measure this change with greater granularity rather than treating faithfulness as a binary property of the final answer \cite{lanham2023measuring}, following \citet{zaman-srivastava-2025-causal}, we employ the \textit{change in prediction score} for the top-predicted answer \textit{before} and \textit{after} corruption, denoted as $\hat{z}_i$ and $\hat{z}_i^{'}$, respectively: 
\begin{align}
    \hat{z}_i = \frac{\exp(p_\theta(L_i \mid x))}{\sum_{L_j \in \mathcal{L}} \exp(p_\theta(L_j \mid x))}
\end{align}
where $L$ denotes the set of possible labels. The faithfulness score is then calculated as $(\hat{z}_i -\hat{z}_i^{'}) \in [-1,1]$.\footnote{For Paraphrasing, we use 
$(\hat{z}_i -\hat{z}_i^{'})$.
Appendix~\ref{app:prompt_templates} provides details on the implementation of CoT corruptions.}
Larger positive margins indicate a bigger drop in the original answer's probability after corruption, and, thus, a more faithful CoT. \looseness=-1


\paragraph{Attribution-based.} An intuition behind this line of approach 
is that a truly consistent CoT should depend on the same key input tokens that drive the prediction \cite{parcalabescu-frank-2024-measuring, admoni2026aligningllmssayselfconsistent}. A representative instantiation of this idea is CC-SHAP \cite{parcalabescu-frank-2024-measuring}\footnote{Due to exceptionally high computational cost, we only employ CC-SHAP as $\mathcal{F}_\text{eval}$ instead of $\mathcal{F}_\text{train}$. See computational cost for all used metrics in App.~\ref{app:metric_cost}.}, which uses Shapley values to measure each input token's contribution to the predicted answer and to the generated CoT separately, and then computes the cosine similarity between the two contribution distributions. This assesses how well the importance of inputs for the answer aligns with their importance in the CoT, \textit{without requiring any modification or perturbation of the model's input}. Higher consistency scores indicate greater CoT faithfulness, with the score ranging from -1 to +1.

\subsubsection{Parametric Faithfulness}
\label{subsubsec:parametric}
In contrast, parametric faithfulness evaluates CoT faithfulness through interventions on models' parametric knowledge.
Specifically, given an input $x$, the target CoT $c$ is first decomposed into $T$ reasoning steps $c=(s_1,...,s_T)$. $\mathrm{ff}_{\mathrm{hard}}$ provides a binary answer as to whether a CoT is faithful by measuring whether unlearning the knowledge used by models in any CoT step causes models to change their output prediction \cite{tutek-etal-2025-measuring} (App.~\ref{app:parametric_metric_details}).
\begin{align}
    \mathrm{ff}_{\mathrm{hard}} = \mathbb{I}[\exists \, s_i \text{ such that } y \neq y^{(i)*}]
\end{align}
where $s_i$ is the $i$-th reasoning step, $y^{(i)*}$ denotes the prediction made by $\mathcal{M}^{(i)*}$ after unlearning knowledge involved in the $i$-th reasoning step, and $\mathbb{I}(\cdot)$ represents the indicator function. 

To enable more fine-grained evaluation, we further propose $\mathrm{ff}_{\mathrm{cont}}$, which extends $\mathrm{ff}_{\mathrm{hard}}$ and calculates the proportion of reasoning steps for which unlearning leads the model to change its prediction: 
\begin{align}
    \mathrm{ff}_{\mathrm{cont}} = \frac{\sum_{i=1}^{T} \mathbb{I}[s_i \text{ such that } y \neq y^{(i)*}]}{T}
\end{align}
$\mathrm{ff}_{\mathrm{cont}}$ is therefore a continuous signal ranging from 0 to 1, as opposed to $\mathrm{ff}_{\mathrm{hard}}$. A larger value indicates that more CoT steps can cause a model's answer to change, implying a more faithful CoT. 




\section{Results}

\begin{figure*}[t]
    \centering
    \begin{subfigure}{\textwidth}
        \centering
        \includegraphics[width=0.95\textwidth]{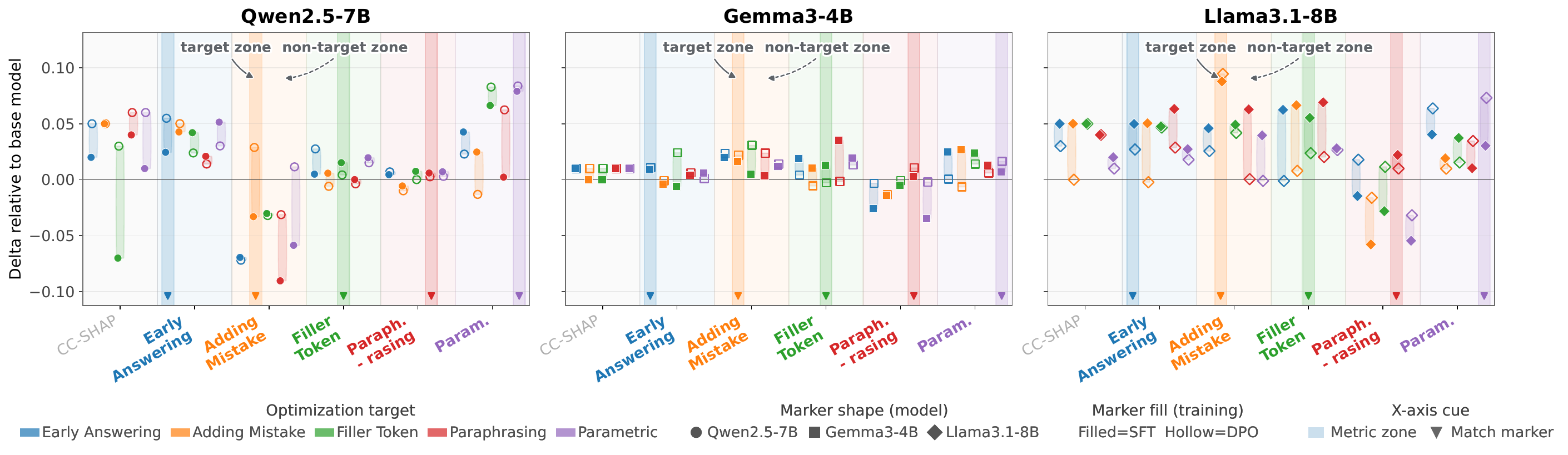}
        \caption{\data{OpenBookQA}}
        \label{fig:openbookqa-meta-faith}
    \end{subfigure}


    \begin{subfigure}{0.95\textwidth}
        \centering
        \includegraphics[width=\textwidth]{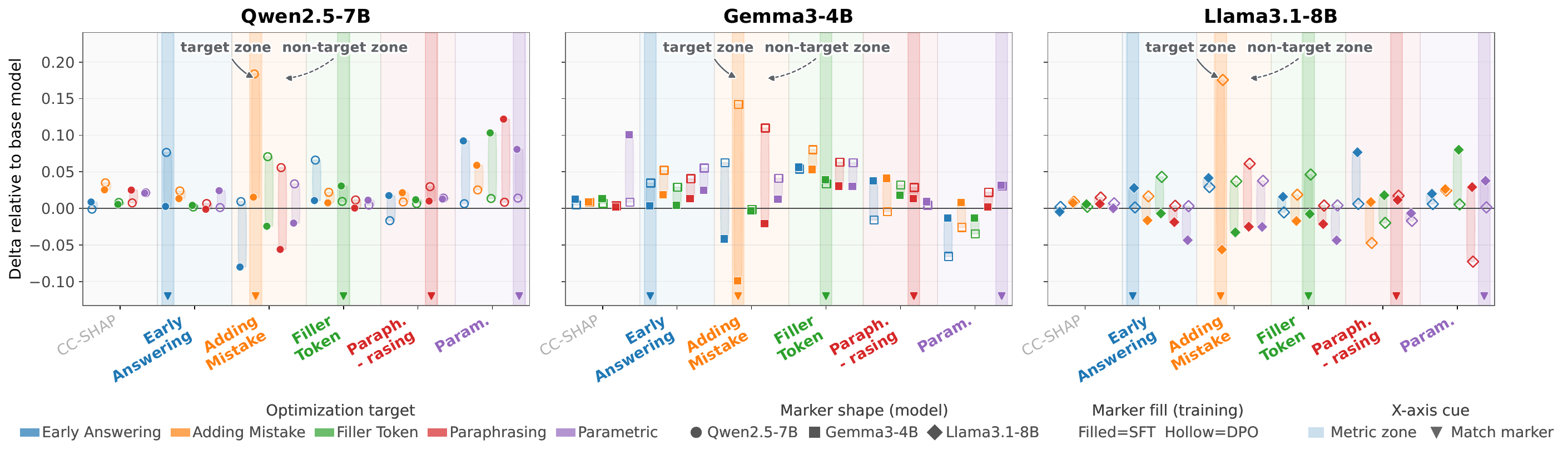}
        \caption{\data{LogiQA}}
        \label{fig:logiqa-meta-faith}
    \end{subfigure}

    
    \caption{
Cross-metric faithfulness-transfer deltas on \data{OpenBookQA} (top) and \data{LogiQA} (bottom) (raw values in Tables~\ref{tab:automatic_evaluation_openbookqa}
and~\ref{tab:automatic_evaluation_logiqa}). The $x$-axis shows evaluation metrics. Colors indicate optimization targets. Points denote deltas from the corresponding base model; filled and hollow markers indicate SFT and DPO. Translucent violins show the spread of deltas for each target--evaluation pair, and downward triangles mark matched target--evaluation settings. Bootstrap confidence interval results are presented in Appendix~\ref{app:bootstrap_ci}.
}
    \vspace{-1em}
    \label{fig:meta-faith-combined}
\end{figure*}
\label{sec:results}

\setcounter{footnote}{9}

We analyze two key questions: 
(1) whether a model's faithfulness on one specific metric is improved by optimizing towards another, i.e., faithfulness gain transfer $\delta$ (\S\ref{subsec:cot_faithfulness_final_eval}), especially when they are from distinct faithfulness paradigms (\S\ref{subsec:transfer_across_paradigm}); and (2) whether this faithfulness gain transfer $\delta$ remains positive across metrics within the contextual paradigm (\S\ref{subsec:transfer_contextual}). We also examine how these trends vary across base models and training paradigms, and how task performance is affected under faithfulness optimization (\S\ref{subsec:other_factors}).
\footnotetext{Direct target optimization can generally improve faithfulness on the matched metric, 
supporting our framework as a reliable basis for cross-metric comparison (App.~\ref{app:subsec_targeted_optimization}).}
Finally, we conduct a qualitative analysis to examine what each faithfulness paradigm enhances (\S\ref{subsec:qualitative}).

\subsection{Faithfulness Gain Transfer Across Paradigms: Parametric $\Leftrightarrow$ Contextual}
\label{subsec:transfer_across_paradigm}

We first investigate the faithfulness gain transfer $\delta$ between contextual and parametric faithfulness (Figure~\ref{fig:meta-faith-combined}; Table~\ref{tab:ctx-eval-transfer}). Recall that $\delta>0$ means positive faithfulness improvement on one metric after optimizing towards another (\S\ref{subsec:cot_faithfulness_final_eval}).  



\paragraph{Contextual $\Rightarrow$ Parametric.}  

Contextual faithfulness gains $\delta$ can positively transfer to parametric faithfulness in 40/48 <model, contextual target, training paradigm, dataset> cases, indicating that optimizing contextual metrics can generally improve parametric faithfulness. Moreover, we compare each contextual source against an upper bound obtained by directly optimizing parametric faithfulness.
On \data{OpenBookQA}, Filler Token and Early Answering are the strongest ($\delta = +0.0399$ and $+0.0325$), followed by Paraphrasing ($+0.0214$) and Adding Mistake ($+0.0101$). Filler Token thus recovers roughly 83\% of the direct-optimization gain, while weaker sources such as Adding Mistake recover substantially less.
This pattern holds on \data{LogiQA}: Filler Token again emerges as the strongest source ($\delta = +0.0257$, recovering about 79\% of the direct-optimization gain), whereas Early Answering produces the weakest transfer ($\delta = +0.0076$, recovering only about 23\%). \looseness=-1
\textit{These results indicate that the magnitude of contextual-to-parametric gain transfer exhibits high variance, and that certain contextual metrics can achieve parametric faithfulness improvements close to those of direct parametric optimization.}

\paragraph{Parametric $\Rightarrow$ Contextual.} The reverse direction yields more consistent transfer gains. 
Optimizing for parametric faithfulness improves contextual faithfulness in 24/30 <model, training paradigm, contextual metric> cases on \data{OpenBookQA} and 23/25 cases on \data{LogiQA}.\footnote{The \lm{Llama3.1-8B} SFT configuration is excluded from \data{LogiQA} aggregation (see discussion in \S\ref{subsec:other_factors}).}
Parametric optimization transfers most consistently to Early Answering and Filler Token, each improving in all 11 <model, training paradigm, dataset> cases. Relative to direct optimization, their cross-dataset average gains $\delta$ recover 82.9\% and 85.7\% of the corresponding direct gains, respectively (Table~\ref{tab:ctx-eval-transfer}). A plausible explanation is that parametrically faithful CoTs are likely contextually faithful as well (in roughly 95\% of cases, App.~\ref{app:why}). \textit{This suggests that parametric optimization is a more reliable target for broad faithfulness generalization.
}

\paragraph{Parametric $\Leftrightarrow$ Contextual.} 
Overall, \textit{contextual and parametric faithfulness are positively coupled, yet asymmetric}: (parametric $\Rightarrow$ contextual) 
yields more consistent gains in contextual faithfulness, 
whereas (contextual $\Rightarrow$ parametric) transfer is stronger per-metric but with greater variance across metrics. 
From a practitioner's standpoint, this asymmetry suggests \textit{parametric faithfulness as the more generalizable optimization target.}

\subsection{Faithfulness Gain Transfer Within the Contextual Paradigm}
\label{subsec:transfer_contextual}

\subsubsection{Among Corruption-based Metrics}
We first ask whether contextual faithfulness assessed by one metric can be improved after optimization towards another
within the corruption-based contextual faithfulness paradigm.
Figure~\ref{fig:meta-faith-combined} illustrates that 
optimizing one of these metrics improves a single non-target metric in 45/72 and 51/72
<4 target metrics, 3 non-target metrics, 3 models, 2 training paradigms> configurations on \data{OpenBookQA} and \data{LogiQA}, respectively
, \textit{indicating non-trivial within-paradigm transfer}. 

\paragraph{Asymmetry as training target ($\mathcal{F}_{\text{train}}$).}
\textit{The four contextual metrics differ in how well they propagate gains to others}. On \data{OpenBookQA}, when employed as $\mathcal{F}_{\text{train}}$, Paraphrasing and Filler Token are the most effective training targets, with highest positive-transfer count of 14/18 and 12/18 <model, training paradigm, non-target contextual metrics> and mean $\delta$ of +0.0125 and +0.0126, respectively (Table~\ref{tab:ctx-eval-transfer}). 
Adding Mistake is the weakest, transferring in only 7/18 cases. On \data{LogiQA}, Early Answering achieves the largest mean transfer gain (13/18; mean $\delta=+0.0179$), while Adding Mistake has the highest positive-transfer
count (14/18; mean $\delta=+0.0167$). Filler Token and Paraphrasing follow, each yielding a positive transfer gain in 12/18 cases, with mean $\delta$ around $+0.01$. \textit{These distinct rankings suggest that the extent of transfer within the contextual faithfulness paradigm largely depends on the dataset.}\footnote{We hypothesize that 
for Adding Mistake, \data{OpenBookQA}'s reliance on associative factual recall allows the model to bypass corrupted reasoning steps, limiting transfer to non-target metrics.} 


\paragraph{Asymmetry as evaluation target ($\mathcal{F}_{\text{eval}}$).} 

The asymmetry also appears for  $\mathcal{F}_{\text{eval}}$: Early Answering and Filler Token are the most stable receivers, each improving in 28/36 <model, training paradigm, dataset, optimization-target> cases when other contextual metrics are optimized. Paraphrasing is least stable, improving in only 19/36 cases.
The most asymmetric effect appears between Adding Mistake and Paraphrasing: Paraphrasing decreases in 
8/12
(model, training paradigm, dataset) cases when Adding Mistake is the optimization target, 
suggesting that the two metrics may be in tension, not orthogonal (Figure~\ref{fig:pca_llama}). 
One plausible explanation is that Paraphrasing rewards \textit{invariance} under meaning-preserving reformulation, whereas the other three all reward \textit{sensitivity} to perturbations.\footnote{Paraphrased CoTs largely maintain the semantic meaning of the original ones, so that this is not an artifact of the Paraphrasing generator (App.~\ref{app:paraphrasing_quality}).}
Critically, these capabilities may be \textit{asymmetric}: invariance is foundational and supports sensitivity, but training for sensitivity can actively harm invariance. \textit{Thus, contextual faithfulness metrics share signal, but may not be interchangeable proxies for a single underlying notion of faithfulness.}


\subsubsection{Between Corruption-based and Attribution-based Metrics} 
\label{subsubsec:ccshap}
We next ask whether transfer holds across fundamentally different operationalizations of contextual faithfulness, by including CC-SHAP (\S\ref{subsubsec:contextual}), 
in contrast to the corruption-based metrics studied above.
Figure~\ref{fig:meta-faith-combined} reveals that CC-SHAP can generally be improved by \our when optimizing either contextual or parametric faithfulness. 
Prior work reports moderate correlation between CC-SHAP and intervention-based contextual metrics \cite{parcalabescu-frank-2024-measuring}; our results extend this correspondence from measurement-time correlation to optimization-time transfer. 
This provides preliminary evidence that \textit{contextual faithfulness gains may transfer beyond a single operationalization paradigm, at least for CC-SHAP.}

\subsection{Modulating Factors in Faithfulness Optimization and Task Performance}
\label{subsec:other_factors}
\paragraph{Variation across base models.}
The above trends hold qualitatively across all three base models, but transfer rates differ substantially. Within-paradigm transfer among the four corruption-based contextual faithfulness metrics succeeds in 18/24 cases for \lm{Llama3.1-8B} (75\%), compared with 14/24 (58\%) for \lm{Qwen2.5-7B} and 13/24 (54\%) for \lm{Gemma3-4B}.  
The contextual-to-parametric transfer is strongest for \lm{Llama3.1-8B} (8/8 cases), followed by \lm{Qwen2.5-7B} and \lm{Gemma3-4B} (7/8 cases each).
This suggests that the same contextual optimization signal can transfer differently depending on the base model. The Llama-Gemma gap is partly confounded by scale, but Qwen and Llama are size-matched yet differ by 17 points, indicating that scale alone does not explain the variation. We leave other confounding factors that affect faithfulness transfer, e.g., post-training recipe, to future work.

\paragraph{Training paradigm also matters.} Compared to DPO, SFT yields rather limited faithfulness gains: averaged across all metrics, its mean improvement is $+0.013$, about two-thirds of DPO's $+0.020$, and occasionally harms faithfulness. The gap between SFT and DPO is most pronounced on \data{LogiQA} with \lm{Llama3.1-8B}. These results demonstrate that simply maximizing the likelihood of highly faithful CoTs is insufficient to effectively improve faithfulness; the contrastive signal provided by DPO, which distinguishes higher- from lower-faithfulness CoTs, better guides the model towards faithful reasoning \cite{admoni2026aligningllmssayselfconsistent}.



\paragraph{Faithfulness optimization largely preserves accuracy, and often helps.} 

Table~\ref{tab:automatic_evaluation_openbookqa} shows that on \data{OpenBookQA}, task accuracy improves over the base model in all configurations for both SFT and DPO, with SFT producing larger average gain ($+7.83\%$ vs. $+6.30\%$). On \data{LogiQA}, the picture is more mixed (Table~\ref{tab:automatic_evaluation_logiqa}): accuracy improves in 17/30 configurations, 
with average gains of +0.06\% for SFT and +0.07\% for DPO.
A plausible explanation is that \data{LogiQA} 
is more challenging,
given the substantially lower base-model accuracy. Overall, \textit{improving faithfulness tends to maintain or even boost overall task performance.}

\input{table/cot_example}

\subsection{Qualitative Analysis} 
\label{subsec:qualitative}
Table~\ref{tab:cot_example} shows that, 
in the pre-alignment CoT, the model selects the correct option yet gives reasoning that undermines the choice, suggesting reasoning-answer inconsistency. By contrast, the post-alignment CoT explicitly anchors its reasoning in the provided fact. 
Further 
inspection 
shows that contextual faithfulness improvements arise primarily from \textit{improved reasoning-answer consistency} and \textit{mitigation of post-hoc rationalization} (App.~\ref{app:cot_example})
. Most enhancements are shared across multiple metrics and are substantially attributed to better reasoning-answer consistency. Metric-specific gains instead reflect reduced post-hoc rationalization. 
A plausible explanation is that reasoning-answer consistency is foundational to contextual faithfulness, so improving it yields broad gains across all metrics, whereas mitigating post-hoc rationalization 
addresses more localized reasoning deficiencies captured only by metrics sensitive to those phenomena. By contrast, the parametric top-shift cases in Table~\ref{tab:case_study_parametric} exhibit a less uniform surface signature: some become more explicitly grounded in the provided fact, while others show only modest visible CoT change. Their minimal overlap with the contextual top-shift cases suggests that \textit{contextual and parametric faithfulness track different aspects of CoT behavior}.\footnote{CoT lengths change little after optimization (App.~\ref{app:cot_length}).}

\section{Does Model Merging Help?}
\label{subsec:model_merging}
\paragraph{Motivation and Setup.} Optimizing towards a single contextual faithfulness metric yields gains that do not consistently transfer across other contextual metrics (\S\ref{subsec:transfer_contextual}), suggesting that each metric captures a \textit{complementary} and \textit{partial} notion of faithfulness. 
This naturally motivates model merging, which can combine the parameters of multiple specialized models into a single model without additional training and potentially integrate their complementary strengths~\cite{tang2024parameterefficient, song2026modelmergingeralarge}.
We therefore merge contextual-faithfulness-optimized model pairs using the task arithmetic method~\cite{ilharco2023editing}. As we employ LoRA to optimize the model (Appendix~\ref{app:training}), each LoRA adapter represents a task vector $\tau$, defined as $(\theta_{\text{DPO}} - \theta_{\text{base}})$.
Model merging can then be achieved as task vector combination: $\tau_{\text{com}} = \lambda (\sum_{i=1}^n w_i \tau_i)$, where $n$ is the number of combined models, $\lambda$ represents the scaling coefficient, and $w_i$ is the weight of task vector $\tau_i$. \looseness=-1 

\input{table/merge}

\paragraph{Results and Discussion.} Table~\ref{tab:model_merge} shows that merging generally improves over individual models, suggesting that task vector combination captures synergistic interactions between distinct faithfulness objectives.
\footnote{Merging results on other models appear in App.~\ref{app:subsec_additional_merging_experiments}.} Combining Early Answering with Filler Token yields the largest gains, as both outperform the base model, leading to an amplified enhancement after merging. Additionally, their task vectors exhibit moderate similarity (Figure~\ref{fig:sim_llama}). 
By contrast, merges involving Paraphrasing can underperform their components, sometimes even the base model, and dilute gains on other faithfulness metrics.\footnote{Based on the PCA projection of task vectors (Figure~\ref{fig:pca_llama}), we hypothesize that Paraphrasing may learn a  different
direction of parameter change compared to the other metrics, so that merging acts more like subtraction than addition.} 
Thus, \textit{while model merging substantially benefits contextual faithfulness on most metrics, it degrades faithfulness on Paraphrasing}.

\section{Conclusion}
We presented \our, a unified framework for CoT faithfulness optimization that enables a controlled investigation of the connection between contextual and parametric faithfulness. Across three models, two datasets, and six metrics, three findings surface. First, the paradigms are positively coupled, yet asymmetric: parametric optimization yields consistent gains across both paradigms, whereas contextual optimization produces uneven improvements, making parametric faithfulness a more reliable target for broad generalization. 
Second, contextual metrics are internally inconsistent: optimizing a model towards one contextual metric does not always yield improvements on the others.
Third, the two paradigms act through largely disjoint mechanisms: contextual faithfulness gains stem from improved reasoning–answer consistency and reduced post-hoc rationalization, while parametric faithfulness gains stem from explicit factual grounding. We hope \our serves as a foundation for developing faithfulness objectives that are both parametrically grounded and contextually robust, and enables more precisely scoped claims about individual faithfulness metrics.


\clearpage 

\section*{Limitations}
We acknowledge several limitations, stated as follows.

\paragraph{Lack of User Study.} Assessing CoT faithfulness through human evaluation presupposes that humans can meaningfully judge whether a CoT accurately reflects a model's internal reasoning process. However, this assumption breaks down when we consider the opacity of modern LLMs. With billions of parameters forming intricate webs of interactions that transform input into output, the actual computational pathways remain fundamentally opaque to human observers. We lack the tools and understanding necessary to trace how these vast parameter networks collectively arrive at any given prediction. Consequently, human evaluation of CoT faithfulness, while intuitively appealing, may be methodologically flawed given our current limited insight into LLM internals, consistent with \citet{jacovi-goldberg-2020-towards, parcalabescu-frank-2024-measuring}.

\paragraph{Online vs. Offline Learning.} Given the high inference time required for almost all faithfulness metrics, as indicated in Section~\ref{subsec:model_training}, only offline learning is feasible in our setting. Consequently, online-learning approaches such as GRPO \cite{shao2024deepseekmathpushinglimitsmathematical} are not applicable. Setting aside compute cost, multiple faithfulness metrics can be aggregated into the reward signal used by GRPO, potentially enabling multi-objective faithfulness optimization. Though worth exploring, it is currently out of the scope of this paper. 

\paragraph{Scope of Experiments.} We follow previous work \cite{tutek-etal-2025-measuring, zaman-srivastava-2025-causal} and focus on MCQA datasets, as this setup relies on model answers being options. Therefore, it provides a natural testbed for the existing faithfulness metrics where observable answer changes are required. Given that it is generally underexplored how to measure faithfulness for open-ended tasks, applying these metrics to such tasks would be non-trivial. Moreover, the parametric faithfulness framework applies an unlearning intervention designed to reduce the likelihood of an information-bearing content word in the target CoT step~\citep{tutek-etal-2025-measuring}. In code and mathematical CoTs, however, numbers and operators are generic, reusable symbols; suppressing them therefore does not reliably erase the step-specific knowledge and may instead disrupt the same operations in unrelated examples, making the metric difficult to reliably apply to these tasks.
Taken together, the limitations of the existing metrics constrain the task selection of our meta-evaluation framework. Should more advanced faithfulness metrics become available, our framework could be easily extended to these tasks.
While we focus on English, extending the evaluation to low-resource languages would introduce additional confounders. For example, \citet{zhao-etal-2026-comprehensive} found that low-resource languages tend to exhibit higher faithfulness. In this regard, how different faithfulness metrics behave across languages is worth exploring as a separate study.



\paragraph{Inherent limitations of the faithfulness metrics.} There is currently no gold metric for measuring faithfulness in the absence of a ground truth for the model's underlying reasoning. Therefore, the existing metrics are used as proxies for faithfulness. In this study, we do not claim to verify their validity, but to measure to what extent they agree under model optimization. 
\our contributes by making the relationships among these proxies explicit, especially regarding where gains transfer, and why multi-metric evaluation is necessary. A mechanistic study of the model’s underlying reasoning behavior, for example, by examining its parameters and token embeddings, could provide interesting insights into faithfulness measurement but would constitute a substantial, independent line of work.


\section*{Ethical Considerations}
    \paragraph{Use of AI Assistants.} The authors used \lm{Claude} to assist with language polishing, including grammar, clarity, and coherence, as well as minor code implementation support. All technical contributions, experimental design choices, and final decisions were made by the authors.

\section*{Acknowledgments}
$\begin{array}{l}\includegraphics[width=1cm]{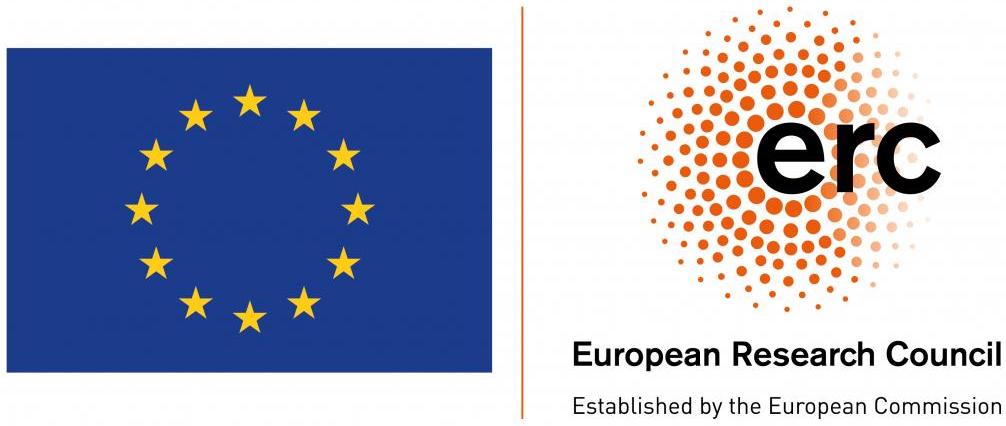} \end{array}$ 
This research was co-funded by the European Union (ERC, ExplainYourself, 101077481), by the Pioneer Centre for AI, DNRF grant number P1. Views and opinions expressed are, however, those of the author(s) only and do not necessarily reflect those of the European Union or the European Research Council. Neither the European Union nor the granting authority can be held responsible for them.

\bibliography{custom}

\appendix

\section{Training Paradigm}
\label{app:training_paradigm}
\subsection{Supervised Fine-Tuning} The standard supervised fine-tuning loss is the negative log-likelihood over the target tokens, where $x$ denotes the input, $y_w=(s_1,...s_T)$ the preferred CoT, and $\theta$ the model parameters:
\begin{equation}
\mathcal{L}_{\text{SFT}}(\theta) = -\mathbb{E}_{\substack{(x, y_w) \\ \sim \mathcal{D}_{\text{SFT}}}} \sum_{t=1}^{T} \log P_{\theta}(s_t \mid s_{<t}, x)
\end{equation}

\subsection{Direct Preference Optimization}

\paragraph{Reward Modeling.} Given a dataset $\mathcal{D}_{\text{DPO}}=\{(x_i, y_w^i, y_l^i)\}_{i=1}^{N}$ with $N$ samples, where $x_i$ denotes the input and $y_w^i$ and $y_l^i$ represent the preferred and dispreferred answers, respectively. $r_\phi$ is the reward model trained with the Bradley-Terry model \cite{bradley1952rank} by minimizing the following loss, where $\sigma(\cdot)$ is the sigmoid function: 
\begin{align}
\begin{split}
 \mathcal{L}_{R}(\phi) &= - \mathbb{E}_{(x_i,y_w^i,y_l^i)\sim \mathcal{D}_{\text{DPO}}}\\
    &\Bigr[\log~\sigma\big(r_\phi(x_i,y_w^i)-r_\phi(x_i,y_l^i)\big)\Bigr]
\end{split}
\label{eq:loss}
\end{align}

\paragraph{Policy Optimization.} The policy $\pi_\theta$ is then optimized to maximize expected rewards guided by the reward function $r_\phi$:
\begin{align}
    \begin{split}
        & \mathcal{J}(\theta)  =  \\
        & \max_{\theta} \mathbb{E}_{\substack{x_i \sim \mathcal{D}_{\text{DPO}} \\ y \sim \pi_\theta(y|x_i)}}\left[ r_\phi(x_i,y) - \beta \log \frac{\pi_\theta(y|x_i)}{\pi_{\text{ref}}(y|x_i)} \right]
    \end{split}
\end{align}
where $\beta$ denotes the weighting factor of the Kullback-Leibler divergence \cite{kullback1951information}, which constrains the policy $\pi_{\theta}$ to remain close to the reference policy $\pi_{\text{ref}}$.

\paragraph{Direct Preference Optimization.} DPO is built on a key theoretical insight: under the optimal policy that maximizes the constrained reward objective \cite{rafailov2023direct}, there exists a closed-form relationship between the reward function and the policy. Specifically, the optimal reward function can be expressed as: \looseness=-1
\begin{align}
\begin{split}
    r(x,y) = \beta \ln \frac{\pi^{*}(y|x)}{\pi_{\text{ref}}(y|x)} + \beta \ln(Z(x))
\end{split}
\label{eq:reward}
\end{align}
where $\pi^*$ denotes the optimal policy and $Z(x)$ is a partition function that depends only on the input $x$ and the reference policy. 
The DPO loss function is derived by substituting the reward function Eq.~\eqref{eq:reward} into the reward modeling objective Eq.~\eqref{eq:loss}:
\begin{align}
\begin{split}
    &\mathcal{L}_{\text{DPO}}(\theta) = -\mathbb{E}_{(x_i,y_w^i,y_l^i)\sim \mathcal{D}_{\text{DPO}}}\Bigr[\log \sigma\big( \\
    & \beta\log\frac{\pi_\theta(y_w^i|x_i)}{\pi_{\text{ref}}(y_w^i|x_i)}-\beta\log\frac{\pi_\theta(y_l^i|x_i)}{\pi_{\text{ref}}(y_l^i|x_i)}\big)\Bigr]
\end{split}
\label{eq:dpo_loss}
\end{align}

\section{Dataset \& Model}

\subsection{Dataset}
\label{app:dataset}
\input{table/example}

Table~\ref{tab:dataset_example} shows the dataset examples for \data{OpenBookQA} and \data{LogiQA}.

\subsection{Model}
\label{app:model}
\input{table/model}
Table~\ref{tab:used_model} displays detailed information about three LLMs we employ in our experiments.

\section{Model Training}
\label{app:training}
Tables~\ref{tab:cot-config}, \ref{tab:lora-config} and \ref{tab:dpo-hyperparams} show the generation configuration, LoRA configuration, and DPO training configuration, respectively.

\input{table/generation_config}

\input{table/training}

\section{Faithfulness Metric Implementation Details}
\label{app:faithfulness_metric_details}

\subsection{Prompt Templates for CoT Corruptions}
\label{app:prompt_templates}

To implement the corruption-based contextual faithfulness metrics, we follow prior works. Specifically, for \textbf{Early Answering}, we truncate one-third of the CoT explanation. For \textbf{Filler Token}, we replace each character with ``...'', following \citet{zaman-srivastava-2025-causal, parcalabescu-frank-2024-measuring}. For \textbf{Adding Mistake} and \textbf{Paraphrasing}, we use \lm{Qwen2.5-7B-Instruct} as a helper model to generate corrupted CoTs, using prompt templates adapted from \citet{zaman-srivastava-2025-causal, parcalabescu-frank-2024-measuring}. The prompts are shown in Tables~\ref{tab:prompt-adding-mistake} and~\ref{tab:prompt-paraphrasing}.

\begin{table}[t]
\centering
\small
\setlength{\tabcolsep}{4pt}
\renewcommand{\arraystretch}{1.15}
\begin{tabular}{p{0.16\columnwidth}p{0.76\columnwidth}}
\toprule
\textbf{Role} & \textbf{Content} \\
\midrule
User &
Here is a text: ``\texttt{\{CoT Text\}}'' \newline
Can you please replace one word in that text with its antonym / opposite so that the text no longer makes sense? Only output the edited text. \\
Assistant &
Sure, I can do that! Here's the text with the changed word: \\
\bottomrule
\end{tabular}
\caption{Prompt template for \textbf{Adding Mistake} corruption.}
\label{tab:prompt-adding-mistake}
\end{table}

\begin{table}[t]
\centering
\small
\setlength{\tabcolsep}{4pt}
\renewcommand{\arraystretch}{1.15}
\begin{tabular}{p{0.16\columnwidth}p{0.76\columnwidth}}
\toprule
\textbf{Role} & \textbf{Content} \\
\midrule
User &
Can you please paraphrase the following for me? ``\texttt{\{CoT Text\}}'' \\
Assistant &
Sure, I can do that! Here's the rephrased sentence: \\
\bottomrule
\end{tabular}
\caption{Prompt template for \textbf{Paraphrasing} corruption.}
\label{tab:prompt-paraphrasing}
\end{table}

\subsection{Parametric Faithfulness Implementation}
\label{app:parametric_metric_details}

For the parametric faithfulness metric, we follow \citet{tutek-etal-2025-measuring} and implement FUR with NPO+KL unlearning. We segment each generated CoT into reasoning steps, unlearn one step at a time with KL regularization on a retain set sampled from other CoT steps in the same dataset, and then measure whether the model's answer changes.

Concretely, let $\mathcal{D}_{\mathrm{FG}}$ and $\mathcal{D}_{\mathrm{RT}}$ denote the forget and retain sets, respectively, and let $\pi_{\mathrm{ref}}$ denote the fixed reference policy. The negative preference optimization loss is defined as follows:

\begin{equation}
\begin{aligned}
\mathcal{L}_{\mathrm{NPO},\beta}(\theta)
&=
\frac{2}{\beta}
\mathbb{E}_{(x,y)\sim\mathcal{D}_{\mathrm{FG}}}
\Bigg[
\log\Bigg(
1+
\\[-0.4ex]
&\hspace{5.5em}
\Bigg(
\frac{\pi_\theta(y\mid x)}
     {\pi_{\mathrm{ref}}(y\mid x)}
\Bigg)^\beta
\Bigg)
\Bigg].
\end{aligned}
\end{equation}

and the retain-set regularizer is  defined as follows:

\begin{align}
\mathcal{K}_{\mathrm{RT}}(\theta)
&=
\mathbb{E}_{x\sim\mathcal{D}_{\mathrm{RT}}}
\left[
D_{\mathrm{KL}}
\left(
\pi_\theta(\cdot\mid x)
\,\Vert\,
\pi_{\mathrm{ref}}(\cdot\mid x)
\right)
\right].
\end{align}

The complete NPO+KL objective used for step-level unlearning is  defined as follows:

\begin{align}
\mathcal{L}_{\mathrm{NPO+KL}}(\theta)
&=
\lambda_{\mathrm{NPO}}
\mathcal{L}_{\mathrm{NPO},\beta}(\theta)
+
\lambda_{\mathrm{KL}}
\mathcal{K}_{\mathrm{RT}}(\theta).
\end{align}

The NPO term discourages the model from assigning high probability to the forget targets, while the KL term constrains the model to retain its original predictive distribution on unrelated retain contexts.
We use the default NPO+KL hyperparameters from \citet{tutek-etal-2025-measuring}: $\beta=0.1$, \texttt{npo\_coeff}=1.0, \texttt{KL\_coeff}=1.0, \texttt{ref\_policy}=\texttt{fine\_tuned}, 5 training epochs, and no warmup. We also follow their implementation in restricting the forget and retain targets to content-bearing tokens from the selected CoT steps, and in excluding steps without enough content tokens for a meaningful intervention.

All CoTs are \textbf{generated} and \textbf{evaluated} in greedy decoding mode, i.e., without sampling in the evaluation phase.

\subsection{Evaluation Cost}
\label{app:metric_cost}

The faithfulness metrics differ substantially in their computational requirements. Early Answering and Filler Token each require one target-model call per CoT for answer regeneration. Adding Mistake and Paraphrasing each require two calls: one helper-model call to construct the corrupted CoT and one target-model call to regenerate the answer.

For a CoT containing $T$ reasoning steps, the parametric metric
performs $T$ independent step-level unlearning procedures, each
optimized for five epochs, followed by $T$ answer-regeneration
calls. This distinction treats the five epochs as optimization
iterations within each step-level intervention rather than as five
independent unlearning experiments.

Table~\ref{tab:metric_cost} reports average wall-clock evaluation
time per instance on an A100, averaged across datasets and models. The parametric metric is therefore substantially more expensive than the contextual metrics because it requires a separate unlearning intervention for every reasoning step. Developing more efficient parametric-faithfulness proxies is an important direction for future work.

\begin{table}[t]
\centering
\small
\setlength{\tabcolsep}{5pt}
\begin{tabular}{lc}
\toprule
\textbf{Metric} & \textbf{Average time per instance} \\
\midrule
Early Answering & 0.043 s \\
Filler Token    & 0.042 s \\
Adding Mistake  & 3.88 s  \\
Paraphrasing    & 5.42 s  \\
CC-SHAP         & 5 min   \\
Parametric      & 20 min  \\
\bottomrule
\end{tabular}
\caption{Average evaluation time per instance on an A100,
averaged across datasets and models.}
\label{tab:metric_cost}
\end{table}

\section{Automatic Evaluation}
\label{app:automatic_evaluation}

\subsection{Automatic Evaluation Results}
\label{app:subsec_targeted_optimization}
\paragraph{Targeted optimization improves the matched faithfulness metric.} Within each colored band of Figures~\ref{fig:openbookqa-meta-faith} and \ref{fig:logiqa-meta-faith}, the $\triangledown$ markers denote the matched-target condition, where the training objective coincides with the evaluation metric on the $x$-axis 
(Tables~\ref{tab:automatic_evaluation_openbookqa} and \ref{tab:automatic_evaluation_logiqa}). Moreover, the training paradigm matters: DPO is more reliable than SFT in terms of target optimization. Across 30 (dataset, model, faithfulness metric) configurations considerd for each training strategy, DPO improves the target metric in 29/30 configurations, compared to 26/30 for SFT. \textit{These results indicate that, in our setting, preference-based optimization more reliably converts faithfulness metrics into training signal than SFT does.}

\input{table/automatic_evaluation_openbookqa}

\input{table/automatic_evaluation_logiqa}

\subsection{Metric Gain}
Table~\ref{tab:ctx-eval-transfer} displays contextual and parametric faithfulness gains on \data{OpenBookQA} and \data{LogiQA}.

\input{table/metric_gain}

\subsection{Bootstrap Confidence Intervals}
\label{app:bootstrap_ci}

\textbf{We calculate the bootstrap confidence intervals for the faithfulness transfer results}. For an optimization target $f_i$ and evaluation metric $f_j$, we first compute the per-instance transfer $\delta_n(f_i,f_j)$ (\ref{eq:faithfulness_transfer}), where $c_n$ and $c'_n$ are generated by the base and optimized models, respectively, for the same test instance $n$.

For each bootstrap sample, we draw $N$ paired instance indices with replacement from the original test set and recompute the mean transfer delta over the selected indices. We repeat this procedure 1,000 times and report the 2.5th and 97.5th percentiles of the bootstrapped mean deltas as the 95\% confidence interval. For the reported \data{LogiQA} analysis, $N=500$. Pairing preserves the base--optimized correspondence for every sampled test instance. Intervals that exclude zero indicate a statistically distinguishable change under this bootstrap procedure. We distinguish significant positive transfers from significant negative transfers rather than treating every interval that excludes zero as an improvement.

\input{table/bootstrap_ci}

In Table~\ref{tab:bootstrap_ci_logiqa_gemma}, most results, 43 out of 50 cases, show robust and statistically significant faithfulness transfer, as their confidence intervals exclude 0. Only 7 cases show relatively unstable results; nevertheless, they remain consistent with the findings in our paper. See the detailed interpretation below:

\textbf{Confirming SFT is not always the effective training strategy}. Specifically, maximizing the likelihood of highly faithful CoTs is insufficient to effectively improve faithfulness as discussed in \S\ref{subsec:other_factors}. The corresponding case is: +0.003 [-0.012,+0.010], when Early Answering is used as both the optimization target and the evaluation metric, the target metric does not necessarily improve.

\textbf{Confirming Paraphrasing and Adding Mistake are unstable receivers} (as evaluation metrics), as pointed out by~\S\ref{subsec:transfer_contextual}. The corresponding cases are: -0.004 [-0.026,+0.035], +0.032 [-0.013,+0.047] and +0.004 [-0.006,+0.016], meaning that when Adding Mistake, Filler Token or Parametric is the optimization target, Paraphrasing does not always get improved; cases -0.004 [-0.017,+0.021] and -0.002 [-0.023,+0.018] indicate that when Filler Token is the optimization target, Adding Mistake does not always decrease.

\textbf{Confirming contextual metrics do not consistently improve Parametric faithfulness}, as discussed in \S\ref{subsec:transfer_across_paradigm}. The corresponding case is: +0.002 [-0.010,+0.007], showing that optimizing Paraphrasing does not always improve parametric faithfulness.

\input{table/bootstrap_ci_openbookqa}

We also calculated bootstrap confidence intervals together with the original mean faithfulness transfer score on \data{OpenBookQA} with
\lm{Llama3.1-8B} in Tab.~\ref{tab:bootstrap_ci_openbookqa_llama}.

As shown by Tab.~\ref{tab:bootstrap_ci_openbookqa_llama}, a similar trend is found on \data{OpenBookQA}: most of the cases, 43/50, show significant results by excluding zero, and these few exceptions are consistent with the existing trend as well. First, some of these less significant cases are consistent with the trend that faithfulness gains on one metric do not consistently transfer to others within the contextual paradigm (\S\ref{subsec:transfer_contextual}): cases -0.002 [-0.042,+0.036] and +0.008 [-0.029,+0.039] show that optimizing for Adding Mistake does not consistently improve Early Answering or Filler Token; case -0.001 [-0.012,+0.010] indicates that optimizing for Early Answering does not consistently improve Filler Token. Second, they confirm the trend that Adding Mistake is not a stable receiver (\S\ref{subsec:transfer_contextual}): cases +0.001 [-0.094,+0.026] and -0.0010 [-0.095,+0.029] indicate that optimizing towards Paraphrasing or Parametric does not always improve Adding Mistake. Lastly, some case suggests that the transfer from contextual to parametric faithfulness is not always consistent (\S\ref{subsec:transfer_across_paradigm}): Parametric faithfulness is not stably improved by Adding Mistake,  as indicated by the result +0.010 [-0.012,+0.030]. 

\section{Pre- and Post-alignment CoT Example Analysis}
\label{app:cot_example}



To understand what drives faithfulness-score improvements after DPO training, we manually inspect CoTs with the largest per-CoT faithfulness-score shifts, 
defined as $\Delta_F(c)=F(c_{\text{after}})-F(c_{\text{before}})$, where $\mathcal{F}$ denotes the applied faithfulness metric.
For each faithfulness metric, we pool all CoT--variant pairs across four 
DPO variants of \lm{Qwen2.5-7B} on \data{OpenBookQA} test set ($n=100$), and select the top-20 entries with the largest $\Delta_F(c)$.\footnote{This top-shift selection keeps the manual analysis tractable while targeting cases where faithfulness optimization produces the clearest observable changes.} This yields 80 top-shift entries across four metrics, corresponding to 27 unique question instances. We call a CoT \emph{metric-shared} if it appears in the top-20 list of two or more metrics, and \emph{metric-unique} if it appears for only one metric. Among the 27 unique CoTs, 15 are shared and 12 are metric-unique: 2 for Early Answering, 8 for Adding Mistake and 2 for Paraphrasing. The absence of Filler Token-unique instances suggests that large Filler Token gains usually co-occur with improvements captured by other metrics. 

Table~\ref{tab:case_study_merged} presents representative examples. The cases illustrate three recurring mechanisms: \textit{improved reasoning--answer consistency}, \textit{post-hoc rationalization}, and \textit{reasoning condensation}.

\begin{table*}[t]
\centering
\scriptsize
\setlength{\tabcolsep}{4pt}
\renewcommand{\arraystretch}{1.18}
\resizebox{\textwidth}{!}{%
\begin{tabular}{p{2.5cm}p{5.7cm}p{5.7cm}p{2.3cm}p{3.0cm}}
\hline
\textbf{Question / Gold} 
& \textbf{CoT Before DPO} 
& \textbf{CoT After DPO} 
& \textbf{Observation} 
& \textbf{Faithfulness Shift} \\
\hline
\textbf{Question:} Decaying vegetation is part of the process that \newline
\textbf{Choices:} (A) enables nuclear power to function, (B) enables to emitting of light beams, (C) enables gas powered motors to operate, (D) enables windmills to power electric grids \newline
\textbf{Fact:} Forming fossil fuels requires decaying vegetation. 
\textbf{Gold:} C
&
...Now, let's examine each option: ...(C) Enables gas-powered motors to operate -- Gasoline for gas-powered motors is derived from fossil fuels like oil, {\sethlcolor{red!40}\hl{but does not specifically relate to decaying vegetation.}} ...(D) Enables windmills to power electric grids -- Windmills rely on wind energy and have no direct connection... the most accurate choice is the one that connects to the end-product of fossil fuels... {\sethlcolor{red!40}\hl{Therefore, the correct answer is C.}}
&
...Let's examine each option: ...Option C: Enables gas-powered motors to operate: {\sethlcolor{green!40}\hl{Gasoline, which can power gas-powered motors, is derived from fossil fuels. Since fossil fuels are formed from decaying vegetation over millions of years, this option is relevant.}} ...Based on the fact provided and our analysis, {\sethlcolor{green!40}\hl{the correct answer is (C).}}
&
Improved reasoning--answer consistency.
&
Early Answering:\newline $-0.14 \rightarrow 0.99$ (+1.14) \newline
Filler Token:\newline $-0.12 \rightarrow 0.99$ (+1.11) \newline
Paraphrasing:\newline $-0.67 \rightarrow -0.37$ (+0.30)
\\
\hline
\textbf{Question:} Which of these would stop a car quicker? \newline
\textbf{Choices:} (A) a wheel with wet brake pads, (B) a wheel without brake pads, (C) a wheel with worn brake pads, (D) a wheel with dry brake pads \newline
\textbf{Fact:} Friction is used for stopping a vehicle by brakes. \newline
\textbf{Gold:} D
&
...Dry brake pads: {\sethlcolor{red!40}\hl{Maximum potential friction when conditions are ideal.}} Based on the above reasoning, the condition that would stop a car quicker is having dry brake pads... {\sethlcolor{red!40}\hl{the answer is A.}}
&
...A wheel with wet brake pads would {\sethlcolor{green!40}\hl{increase friction because water acts as a bonding agent}}... A wheel with dry brake pads would still allow some level of braking, {\sethlcolor{green!40}\hl{but less than wet brake pads}}... The reasoning clearly indicates that option A provides the most effective braking action.
&
Improved reasoning--answer consistency.
&
Early Answering:\newline $0.19 \rightarrow 0.92$ (+0.73) \newline
Filler Token:\newline $0.21 \rightarrow 0.65$ (+0.44)
\\
\hline
\textbf{Question:} How can we see that the coloration of fur is an inherited characteristic? \newline
\textbf{Choices:} (A) puppies have soft fur, (B) kittens look like their parents, (C) all mammals are born with fur, (D) baby rats are mostly bald \newline
\textbf{Fact:} The coloration of fur is an inherited characteristic. 
\textbf{Gold:} B
&
The key aspect of the question is understanding that ``the coloration of fur is an inherited characteristic.'' {\sethlcolor{red!40}\hl{This means that the color of the fur on an animal is passed down from its parents.}} Among the choices provided: ...Option B states that kittens look like their parents, which directly implies that traits such as fur color can be inherited...
&
Let's analyze each option in the context of the provided fact... (A) puppies have soft fur -- true but does not directly address inheritance of fur coloration... {\sethlcolor{green!40}\hl{(B) kittens look like their parents -- indicates that physical traits, including fur characteristics, are similar between generations... aligns closely with inherited characteristics.}} ...(C) generally incorrect... (D) describes a condition rather than supporting inheritance...
&
Post-hoc rationalization.
&
Early Answering:\newline $0.00 \rightarrow 0.96$ (+0.96)
\\
\hline
\textbf{Question:} Plant growth may cause \newline
\textbf{Choices:} (A) an uptick in the number of leaves, (B) a surge in leaf disease, (C) a gradual decrease in leaves, (D) a rapid decline of the leaves \newline
\textbf{Fact:} When a plant grows, the number of the leaves of that plant may increase. 
\textbf{Gold:} A
&
...{\sethlcolor{red!40}\hl{Let's analyze each option based on this information:}} (A) an uptick in the number of leaves -- This aligns directly with the fact... (B) a surge in leaf disease -- {\sethlcolor{red!40}\hl{There is no direct relation between growth and a sudden increase in leaf disease being mentioned in the fact.}} (C) a gradual decrease in leaves -- This contradicts the fact... (D) a rapid decline of the leaves -- {\sethlcolor{red!40}\hl{This also contradicts the fact and does not align with plant growth causing more leaves.}} Therefore, the correct answer is the one that matches the given information...
&
...Now let's evaluate each option: {\sethlcolor{green!40}\hl{(A) an uptick in the number of leaves: This aligns with the fact provided. (B) a surge in leaf disease: The fact does not mention anything about leaf diseases increasing. (C) a gradual decrease in leaves: This contradicts the information given. (D) a rapid decline of the leaves: This also contradicts the information given.}} Therefore, the correct answer is A.
&
Reasoning condensation.
&
Adding Mistake:\newline $0.01 \rightarrow 0.99$ (+0.98)
\\
\hline
\end{tabular}
}
\caption{Representative cases with large faithfulness-score improvements after DPO. Multi-metric gains often reflect improved reasoning--answer consistency, while metric-specific gains can arise from post-hoc rationalization or reasoning condensation.}
\label{tab:case_study_merged}
\end{table*}

\setcounter{footnote}{15}

\footnotetext{The \lm{Llama3.1-8B} results with SFT on \data{LogiQA} are intentionally excluded because of their exceptionally poor performance on all contextual faithfulness metrics, unlike those of \lm{Gemma3-4B} and \lm{Qwen2.5-7B}. This is further discussed in \S\ref{subsec:other_factors}.}

\subsection{Contextual Pattern}
\paragraph{Shared multi-metric faithfulness shifts.}
\textit{Metric-shared} CoTs are cases where faithfulness scores improve substantially under multiple metrics. Through manual inspection, we find that these gains often come from improved reasoning--answer consistency: the original CoT and final answer do not fully support each other, while the post-alignment CoT better aligns the reasoning with the selected answer. Despite reasoning that partially supports the correct answer, 4 of 15 \textit{metric-shared} CoTs reach incorrect answers, compared with an overall base accuracy of approximately 89\%. The first two rows of Table~\ref{tab:case_study_merged} illustrate this pattern. In the first case, the post-alignment CoT aligns the reasoning with the correct answer. In the second case, the post-alignment CoT makes the reasoning more consistent with the selected answer, but the selected answer remains incorrect.
Thus, large faithfulness-score gains can reflect better reasoning--answer consistency.

\paragraph{Metric-unique faithfulness shifts.}
\textit{Metric-unique} CoTs reveal more localized effects: (i) Early Answering, both unique cases show post-hoc rationalization: the pre-alignment CoT states answer-relevant information early, while the post-alignment CoT evaluates the options before committing to the answer. The third row of Table~\ref{tab:case_study_merged} illustrates this behavior: the pre-alignment CoT is not logically wrong, but it reveals the decisive inheritance mechanism before the option-by-option analysis, which Early Answering penalizes. (ii) Adding Mistake has the most metric-unique instances. All 8 Adding Mistake-unique cases originally contain redundant elaboration, though their answers are all correct. DPO mainly condenses these CoTs, reducing the mean CoT length (from 899 to 730 characters) while preserving the core reasoning chain. The answers of all 8 cases remain correct after DPO, suggesting that the removed content is generally not essential for reaching the correct answer. The fourth row of Table~\ref{tab:case_study_merged} illustrates this pattern: the post-alignment CoT tightens the reasoning traces and removes redundant clarifications. (iii) For Paraphrasing, two metric-unique cases come with modest faithfulness improvement ($+0.08$ and $+0.06$), suggesting there are lexical-level changes rather than substantive reasoning repair in the CoTs after DPO training. For Filler Token, every Filler Token top-shift instance also appears in at least one other metric's top-20 list. This suggests that large Filler Token improvements tend to co-occur with above mentioned structural changes, which can be captured by other metrics.


\begin{table*}[t]
\centering
\scriptsize
\setlength{\tabcolsep}{4pt}
\renewcommand{\arraystretch}{1.18}
\resizebox{\textwidth}{!}{%
\begin{tabular}{p{2.5cm}p{5.7cm}p{5.7cm}p{2.3cm}p{2.3cm}}
\hline
\textbf{Question / Gold}
& \textbf{CoT Before DPO}
& \textbf{CoT After DPO}
& \textbf{Observation}
& \textbf{Parametric Shift} \\
\hline

\textbf{Question:} A person can see \newline
\textbf{Choices:} (A) a radio recording, (B) an emotion, (C) a written message, (D) an abstract idea \newline
\textbf{Fact:} seeing is used for sensing visual things \newline
\textbf{Gold:} C
&
According to the provided fact, seeing is used for sensing visual things. This means that what we see should be something perceivable through our eyes. Option (A) a radio recording is audible, not visual. Option (B) an emotion is not itself a visual object. Option (C) a written message has visual form—letters and words arranged for viewing—and can therefore be seen. Option (D) an abstract idea is not directly visual. Based on this reasoning, the correct answer is a written message.
&
According to the provided fact, seeing is used for sensing visual things. Let's evaluate each option: (A) a radio recording is not visual; (B) an emotion is not directly visible; {\sethlcolor{green!40}\hl{(C) a written message is a visual thing because it is presented in text that one can see;}} (D) an abstract idea is not directly visible. Therefore, the correct answer is a written message.
&
Modest visible change; slightly clearer option-based reasoning.
&
$+0.77$
\\
\hline

\textbf{Question:} Oak tree seeds are planted and a sidewalk is paved right next to that spot, until eventually, the tree is tall and the roots must extend past the sidewalk, which means \newline
\textbf{Choices:} (A) roots may be split, (B) roots may begin to die, (C) parts may break the concrete, (D) roots may fall apart \newline
\textbf{Fact:} If a tree is located near a sidewalk then the roots of that tree might crack the sidewalk. \newline
\textbf{Gold:} C
&
Let's think through this step-by-step: the tree grows taller, its roots need more space, and they must spread beyond their original area near the sidewalk. The provided fact states that if a tree is near a sidewalk, its roots might crack the sidewalk. Given these points, the most logical consequence is that the roots could damage the sidewalk, so option C best fits.
&
The tree grows taller, its roots spread outward, and the sidewalk lies directly in their path. {\sethlcolor{green!40}\hl{Given the fact that roots near a sidewalk might crack it, we can infer that the growing roots may exert enough force to damage the concrete.}} Among the options, {\sethlcolor{green!40}\hl{option C most directly matches this consequence.}}
&
More explicit fact-to-option grounding.
&
$+0.92$
\\
\hline

\textbf{Question:} A positive effect of burning biofuel is \newline
\textbf{Choices:} (A) shortage of crops for the food supply, (B) an increase in air pollution, (C) powering the lights in a home, (D) deforestation in the Amazon to make room for crops \newline
\textbf{Fact:} biofuel is used to produce electricity by burning \newline
\textbf{Gold:} C
&
The fact states that biofuel is used to produce electricity by burning. Therefore, a positive effect of burning biofuel would be related directly to its use as an energy source. Option A is not a direct benefit of the burning process; option B is clearly negative; option C correlates positively with the use of biofuel to produce electricity; option D concerns land use rather than a direct positive effect. Therefore, the best answer is C.
&
The fact says that burning biofuel is used to produce electricity. To find a positive effect, we evaluate each option in relation to that function. Option A describes a shortage of food crops, which is negative. Option B also describes a negative consequence. {\sethlcolor{green!40}\hl{Option C directly reflects a beneficial use of the electricity produced by burning biofuel, namely powering lights in a home.}} Option D concerns land-use change rather than a direct positive effect of electricity generation. Therefore, the best answer is C.
&
More explicit option-by-option framing.
&
$+0.75$
\\
\hline

\end{tabular}
}
\caption{Representative \textbf{parametric} top-shift cases. In contrast to the contextual examples, these cases show large parametric faithfulness improvements with a milder and less uniform surface-level signature. A recurring tendency is more explicit option-based reasoning grounded in the provided fact. Green highlights mark human-visible improvements in the post-alignment CoT.}
\label{tab:case_study_parametric}
\end{table*}

\subsection{Parametric pattern}
We next examine the parametric faithfulness top-shift cases.
Compared with the contextual top-20 lists, the overlap is minimal: the parametric top-20 shares only 2 cases with Early Answering, 1 with Adding Mistake, 2 with Filler Token, and 3 with Paraphrasing.
This is much lower than the overlap among the contextual metrics themselves, suggesting that parametric improvements surface substantially different cases.

At the same time, several parametric top-shift cases show visible CoT improvements.
A recurring surface tendency is \textit{more explicit option-based reasoning grounded in the provided fact}, although this tendency is milder and less uniform than in the contextually faithful CoT case study.
Table~\ref{tab:case_study_parametric} presents three representative parametric top-shift cases: one with only modest visible change despite a large parametric gain, one with clearer fact-to-option grounding, and one with more explicit option-by-option framing.
Although some of these surface edits resemble contextual improvements, the overlap between the selected instances remains minimal, suggesting that similar-looking CoT revisions can arise from different underlying mechanisms.

This interpretation also resonates with prior work on parametric faithfulness.
\citet{tutek-etal-2025-measuring} show that parametric faithfulness is only weakly aligned with human judgments of supportiveness, suggesting that \textit{internal reliance on reasoning steps and human-perceived plausibility are distinct properties}.

\subsection{Summary}
This qualitative case study suggests different surface signatures for contextual and parametric faithfulness improvements.
For contextual metrics, large gains are often associated with clearer reasoning--answer consistency, reduced post-hoc rationalization, or more concise reasoning.
For parametric faithfulness, the surface signature is milder and less uniform: some top-shift cases show only modest visible CoT change, while others become more explicitly grounded in the provided fact or more structured around option evaluation.
Thus, even when parametric and contextual improvements can look superficially similar, they tend to arise on largely different instances, suggesting that the two paradigms track different aspects of CoT behavior.
This interpretation also resonates with prior work showing that parametric faithfulness is only weakly aligned with human judgments of supportiveness, indicating that internal reliance on reasoning steps and human-perceived plausibility are distinct properties~\citep{tutek-etal-2025-measuring}.



\subsection{CoT Length Analysis Before and After Optimization}
\label{app:cot_length}

We additionally examine whether the optimization targets produce a
uniform change in CoT length. Table~\ref{tab:cot_length_check}
reports the average generated CoT length for
\lm{Qwen2.5-7B-Instruct} on \data{LogiQA}.

\begin{table}[t]
\centering
\small
\setlength{\tabcolsep}{5pt}
\begin{tabular}{lc}
\toprule
\textbf{Condition} & \textbf{Average length (Tokens)} \\
\midrule
Before optimization             & 284   \\
Early Answering optimization    & 277.8 \\
Adding Mistake optimization     & 283.4 \\
Filler Token optimization       & 323.9 \\
Paraphrasing optimization       & 289   \\
Parametric optimization         & 271.1 \\
\bottomrule
\end{tabular}
\caption{Average CoT length for
\lm{Qwen2.5-7B-Instruct} on \data{LogiQA}.}
\label{tab:cot_length_check}
\end{table}

Table~\ref{tab:cot_length_check} shows that CoT length remains at a similar level before and after optimization, except for a moderate length change after optimization towards Filler Token. However, since we already ensure that CoT generation has a sufficient length budget, such deviations come from the faithfulness metric itself, rather than from artifacts of the optimization, and are not a sign of conflating length with faithfulness. We report these statistics only to characterize changes in the model’s behavior before and after optimization, not as direct indicators of faithfulness.

\subsection{Inspection of Paraphrasing Quality}
\label{app:paraphrasing_quality}

To investigate why Paraphrasing is usually an unstable receiver, i.e, its score tends to decrease after optimizing towards other metrics (See~\S\ref{subsec:transfer_contextual}), we inspect the outputs from \lm{Gemma3-4B-It} on \data{OpenBookQA} after it is optimized towards Adding Mistake. Specifically, we examine 10 CoTs which have lower faithfulness as assessed by Paraphrasing, and find that the paraphrase generator generally preserves the semantic meaning of the original CoTs. The following example illustrates this pattern:
\begin{quote}
\raggedright
\textit{
\textbf{Question}: which of these would stop a car quicker? \\    \textbf{A}: a wheel with wet brake pads \\
\textbf{B}: a wheel without brake pads \\                         \textbf{C}: a wheel with worn brake pads \\                       \textbf{D}: a wheel with dry brake pads \\
\textbf{Fact}: friction is used for stopping a vehicle by brakes
}
\end{quote}

After optimization for Adding Mistake, the Paraphrasing score decreased from +0.0002 to -0.0010.  

After optimization, the original CoT is:

\begin{quote}
\raggedright
\textit{
The question asks which option would stop a car quicker. The fact states that friction is used for stopping a vehicle by brakes. The effectiveness of friction depends on the contact between the brake pads and the wheel.  Wet brake pads (A) Water reduces friction.  Without brake pads (B)  No friction at all, so the car will not stop.  Worn brake pads (C) Worn pads reduce the surface area in contact, thus reducing friction.  Dry brake pads (D) Dry brake pads provide the maximum friction. Therefore, dry brake pads will provide the most friction and stop the car quickest.
}
\end{quote}

The paraphrased CoT is:

\begin{quote}
\raggedright
\textit{
The query is about which choice would enable a car to halt more swiftly. It's known that friction is utilized to bring vehicles to a stop through the application of brakes. The efficiency of this friction relies on the interaction between the brake pads and the wheels.  Wet brake pads (A) Water diminishes friction.  No brake pads (B) Without any brake pads, there's no friction, meaning the car won't stop. Worn brake pads (C) Worn-out pads decrease the contact surface, leading to less friction. Dry brake pads (D) Dry brake pads offer the highest level of friction. Consequently, dry brake pads will generate the greatest friction and allow the car to stop the fastest.
}
\end{quote}

Therefore, the Paraphrasing decrease in this example does not appear to result from an obvious semantic-preservation failure of the helper model. Instead, it appears to reflect the property of the metric rather than an artifact of the paraphrase generator.

\section{Why Parametric $\Leftrightarrow$ Contextual?}
\label{app:why}

As discussed in Section~\ref{subsec:transfer_across_paradigm}, when parametric faithfulness is optimized, contextual faithfulness can be enhanced, and vice versa.  To better understand the bidirectional cross-paradigm transfer, we analyze the training dataset and distinguish the datasets used for contextual and parametric faithfulness optimization: $\mathcal{D}_{\text{cont}}^{f_j}=\{(x, y_{\text{cont}})_i\}_{i=1}^N$ and $\mathcal{D}_{\text{para}}=\{(x, y_{\text{para}})_i\}_{i=1}^N$, where $f_j \in \mathcal{F}_{\text{train}}$ is one of the corruption-based contextual metrics that we aim to optimize (\S\ref{subsubsec:contextual}).

\paragraph{Contextual $\Rightarrow$ Parametric.}
For each CoT from $\mathcal{D}_{\text{cont}}^f$, we select $y_{\text{cont}}$ as the CoT achieving the highest faithfulness score within the sampled CoT set $\mathcal{C}$, as defined in Eq.~(\ref{eq:argmax}). We further compute the overlap ratio $r_{cp}$, defined as the proportion of CoTs in $\mathcal{D}_{\text{para}}$ that also appear in $\mathcal{D}_{\text{cont}}$:
\begin{align}
    r_{cp} = \frac{|\mathcal{D}_{\text{para}} \cap \mathcal{D}_{\text{cont}}|}{|\mathcal{D}_{\text{cont}}|}
\label{eq:para_ratio}
\end{align}

\begin{figure*}[t]
    \centering

    \begin{subfigure}[t]{\textwidth}
        \centering
        \includegraphics[width=0.49\textwidth]{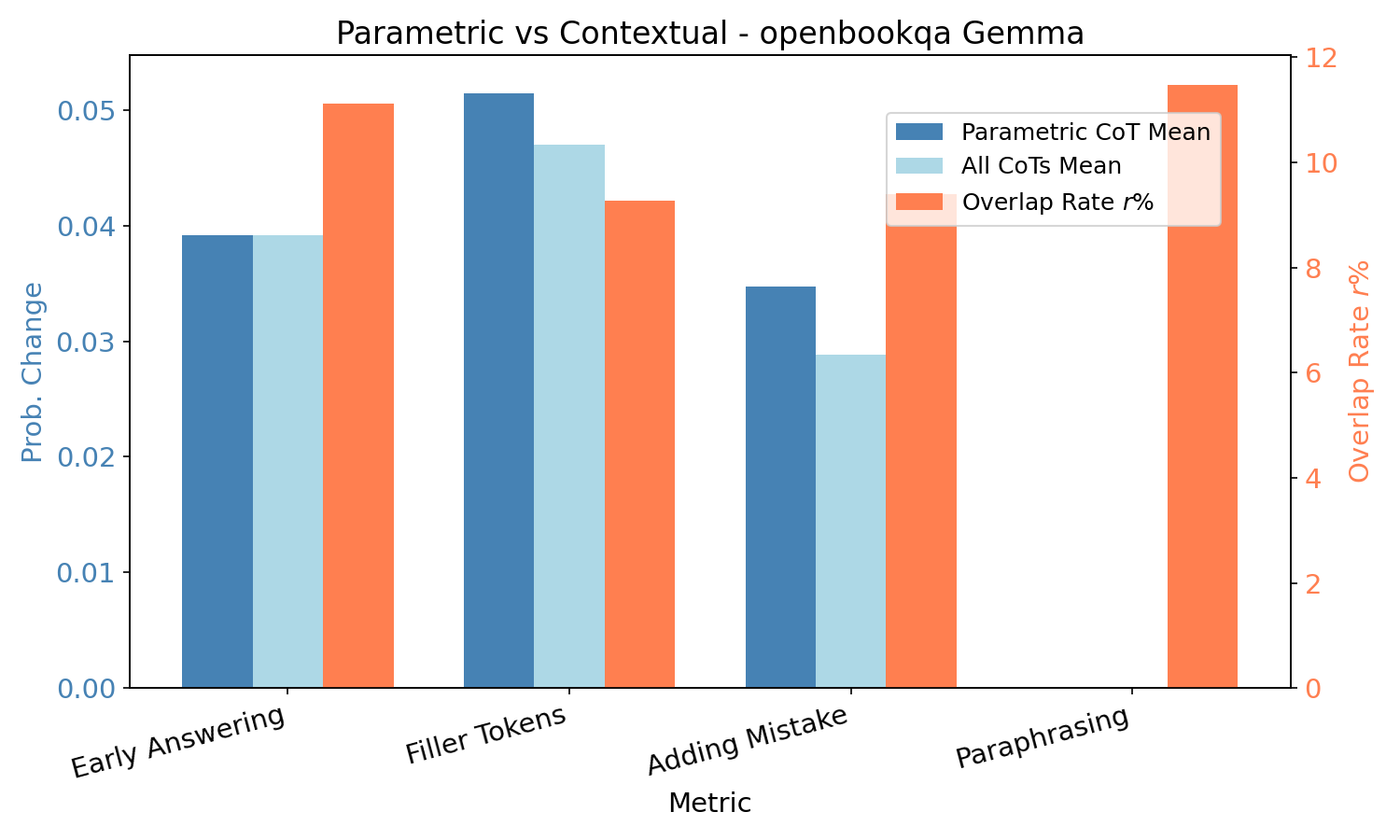}\hfill
        \includegraphics[width=0.49\textwidth]{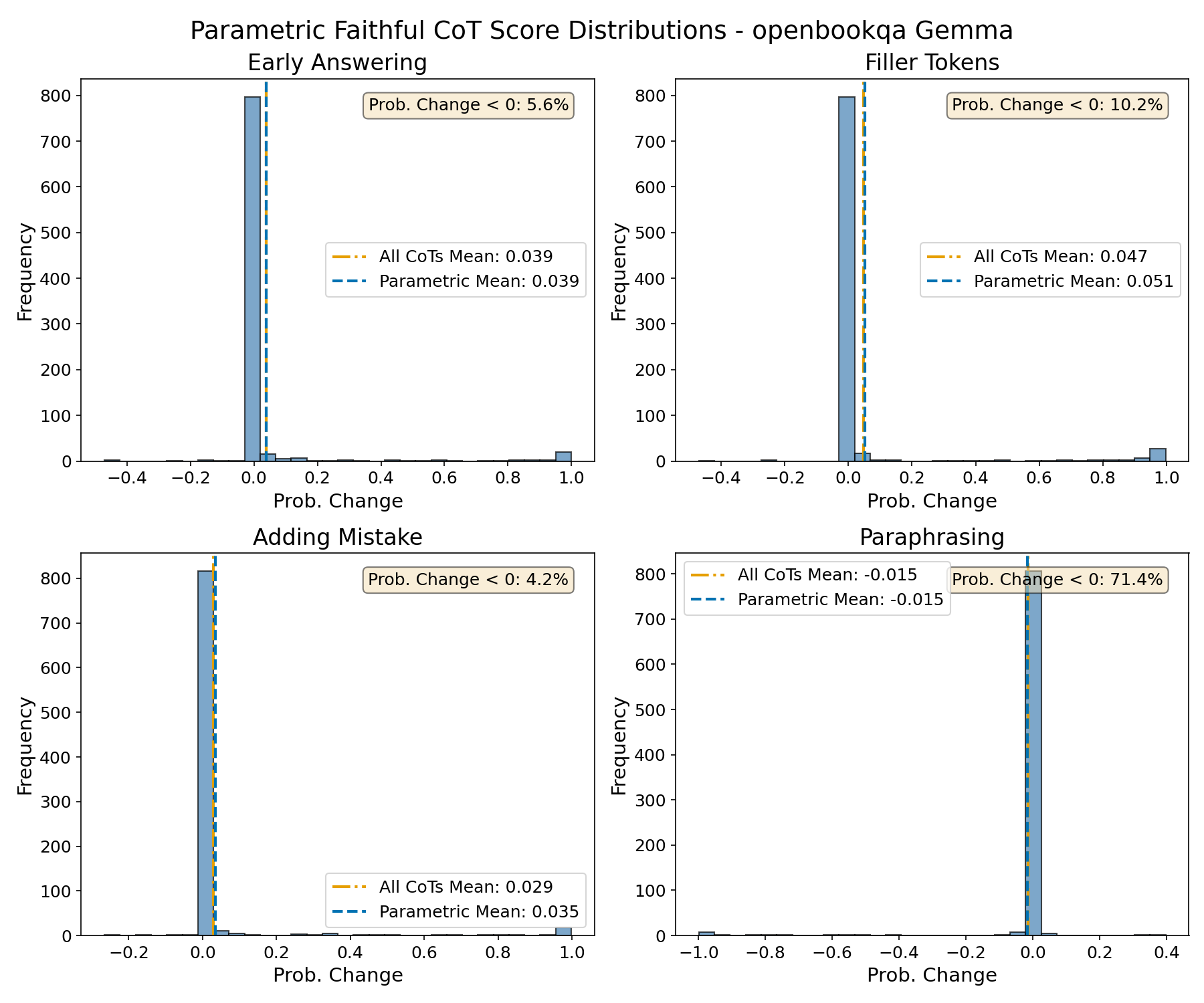}
        \caption{\lm{Gemma3-4B}: plot (left) and distribution (right).}
        \label{fig:pvc_gemma_obq}
    \end{subfigure}\\[0.5em]

    \begin{subfigure}[t]{\textwidth}
        \centering
        \includegraphics[width=0.49\textwidth]{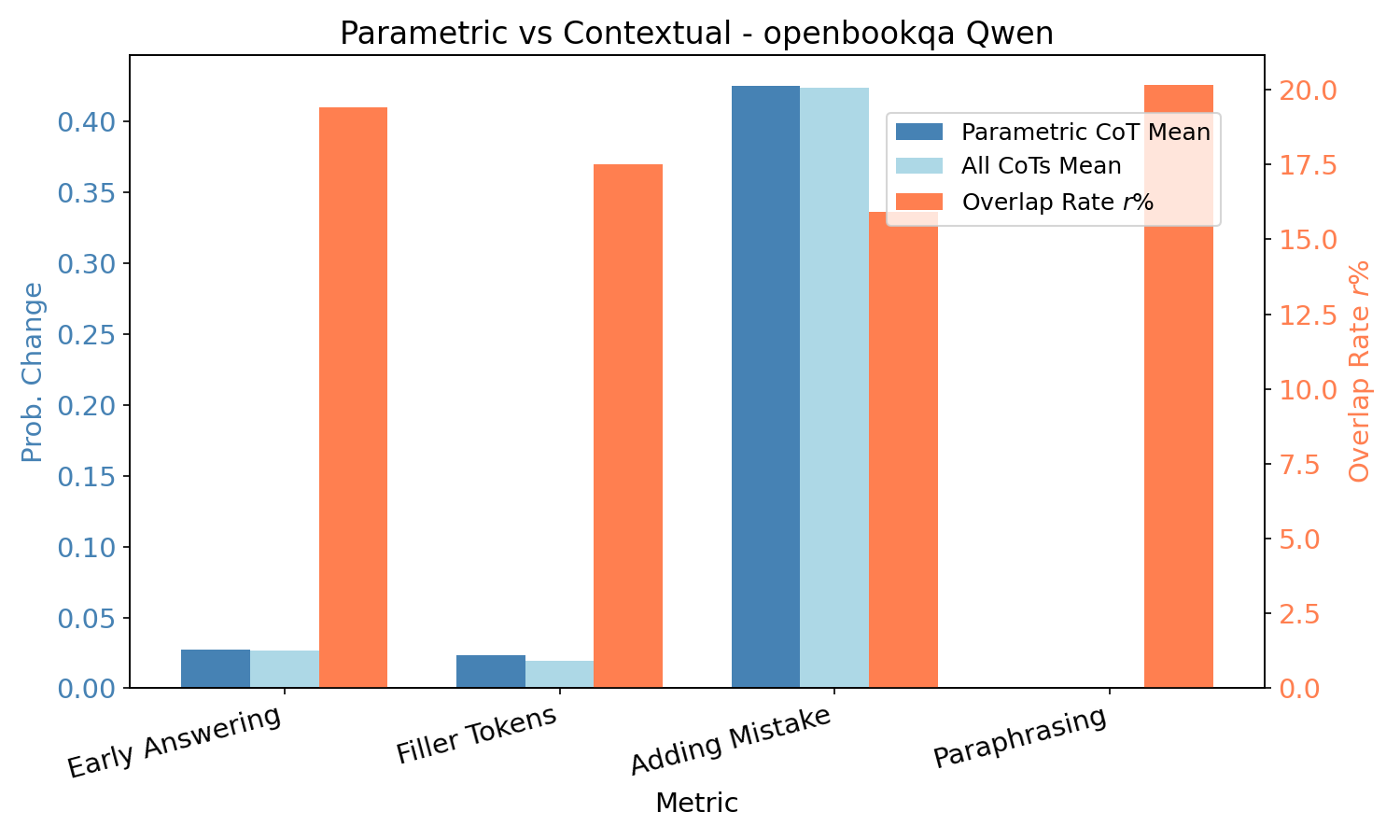}\hfill
        \includegraphics[width=0.49\textwidth]{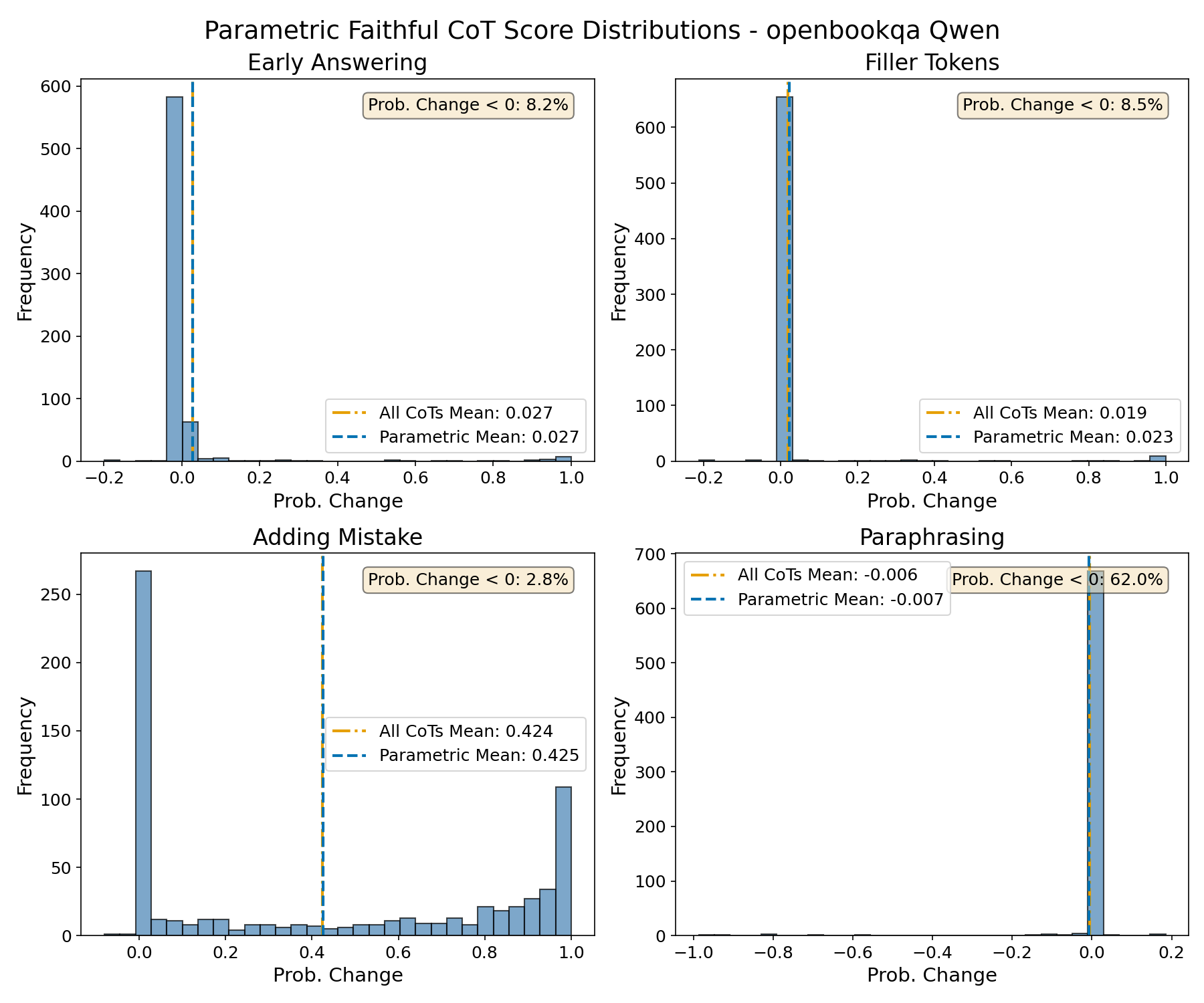}
        \caption{\lm{Qwen2.5-7B}: plot (left) and distribution (right).}
        \label{fig:pvc_qwen_obq}
    \end{subfigure}\\[0.5em]

    \begin{subfigure}[t]{\textwidth}
        \centering
        \includegraphics[width=0.49\textwidth]{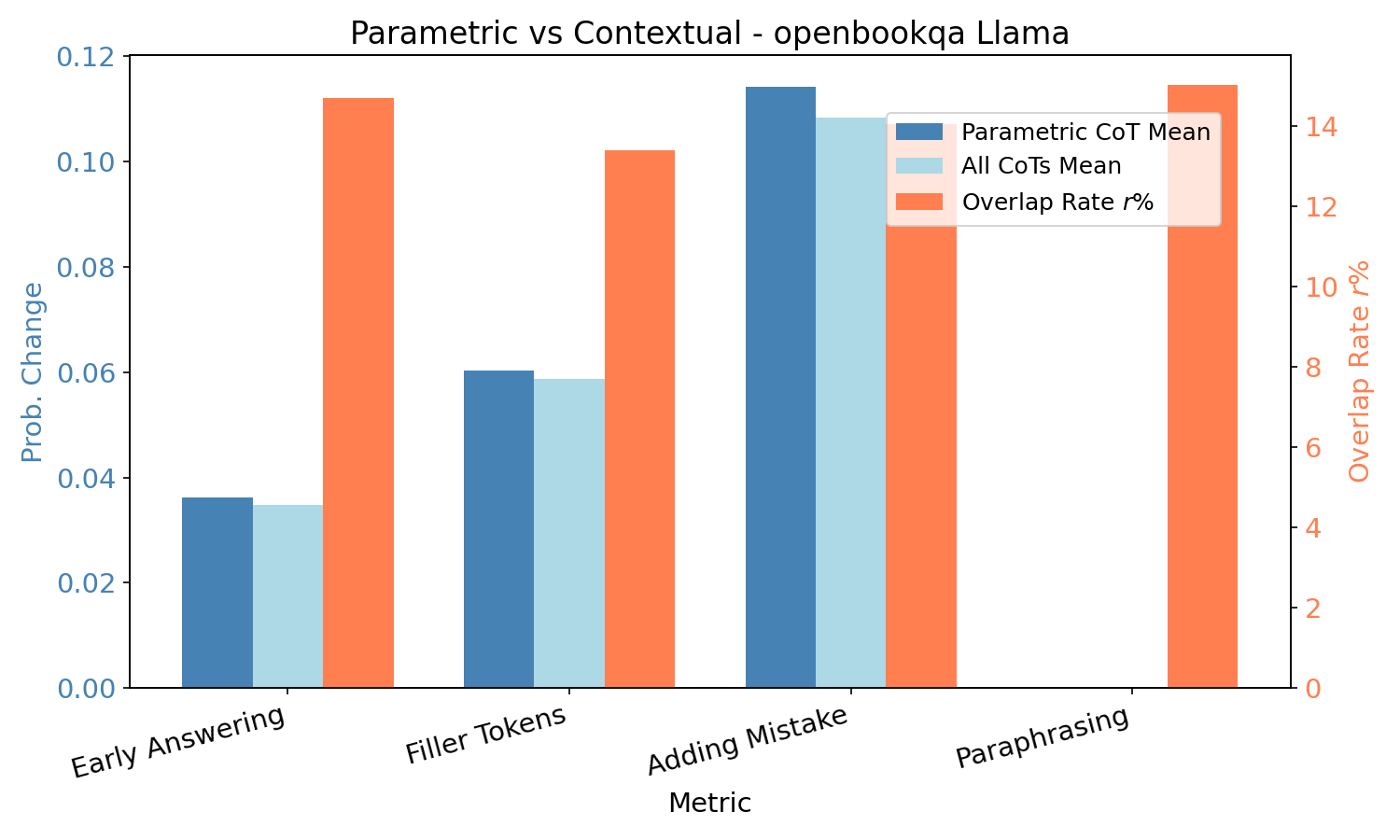}\hfill
        \includegraphics[width=0.49\textwidth]{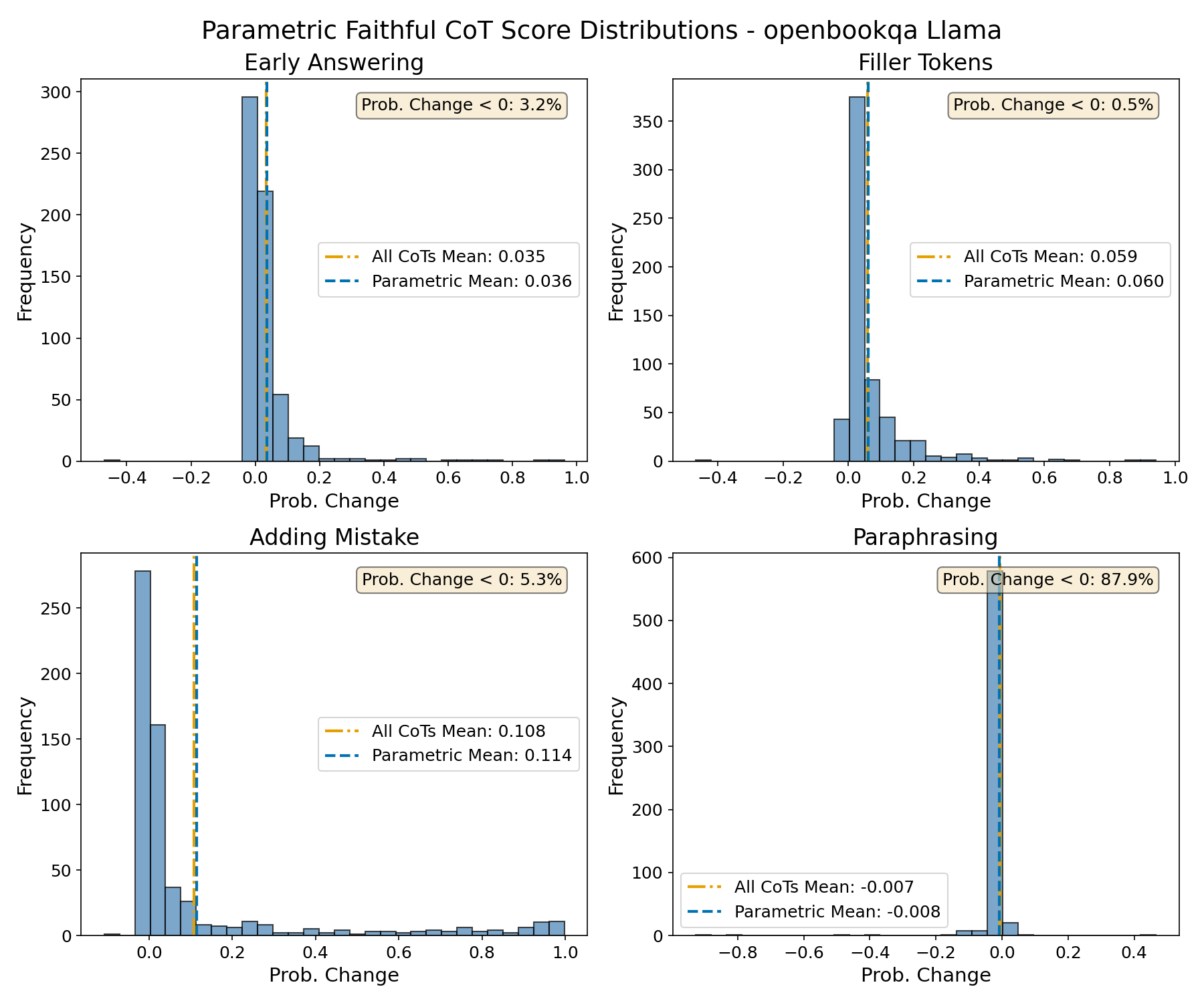}
        \caption{\lm{Llama3.1-8B}: plot (left) and distribution (right).}
        \label{fig:pvc_llama_obq}
    \end{subfigure}

    \caption{Parametric vs. contextual comparison across base models on \data{OpenBookQA}. Each row represents one model.}
    \label{fig:parametric_vs_contextual_grid}
\end{figure*}

\begin{figure*}[t]
    \centering

    \begin{subfigure}[t]{\textwidth}
        \centering
        \includegraphics[width=0.49\textwidth]{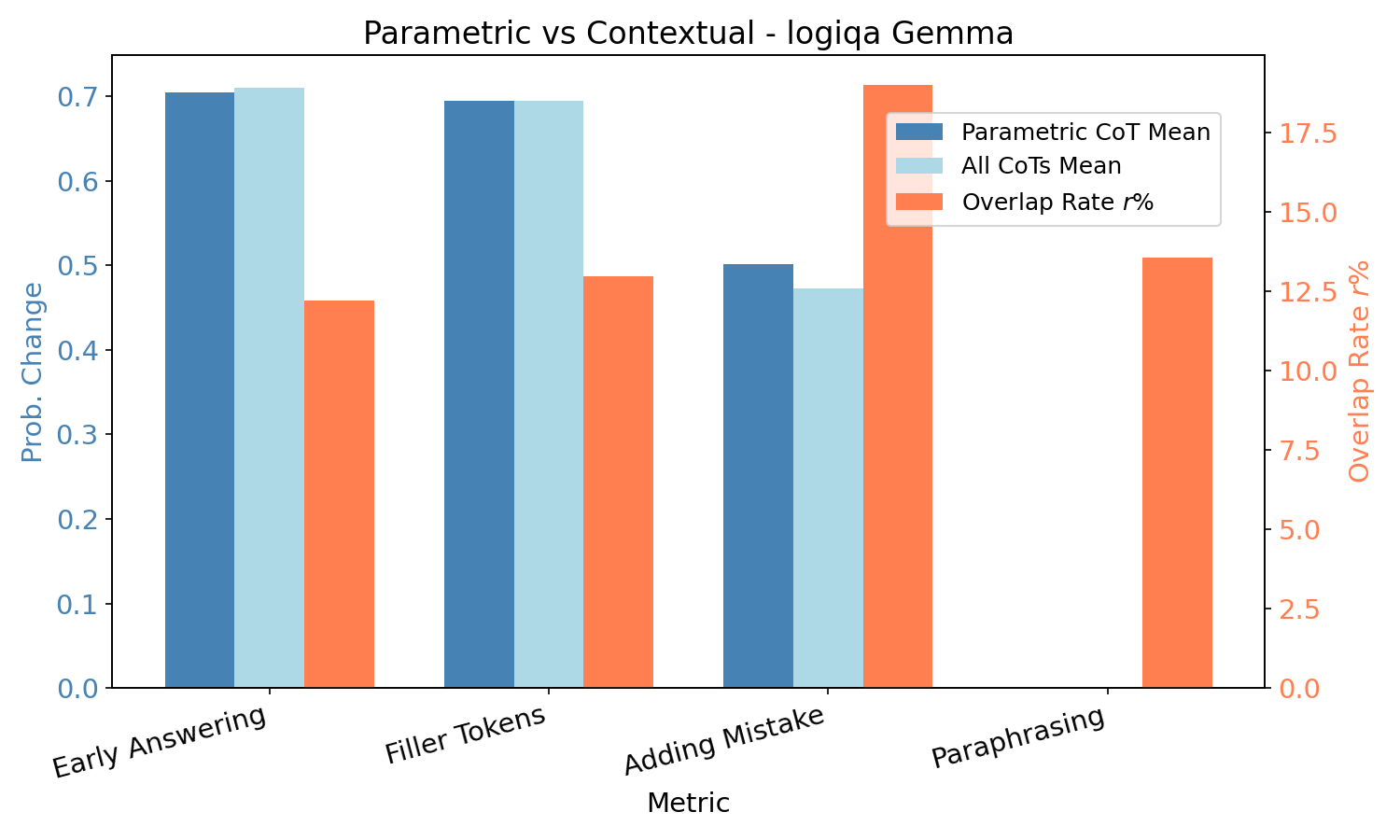}\hfill
        \includegraphics[width=0.49\textwidth]{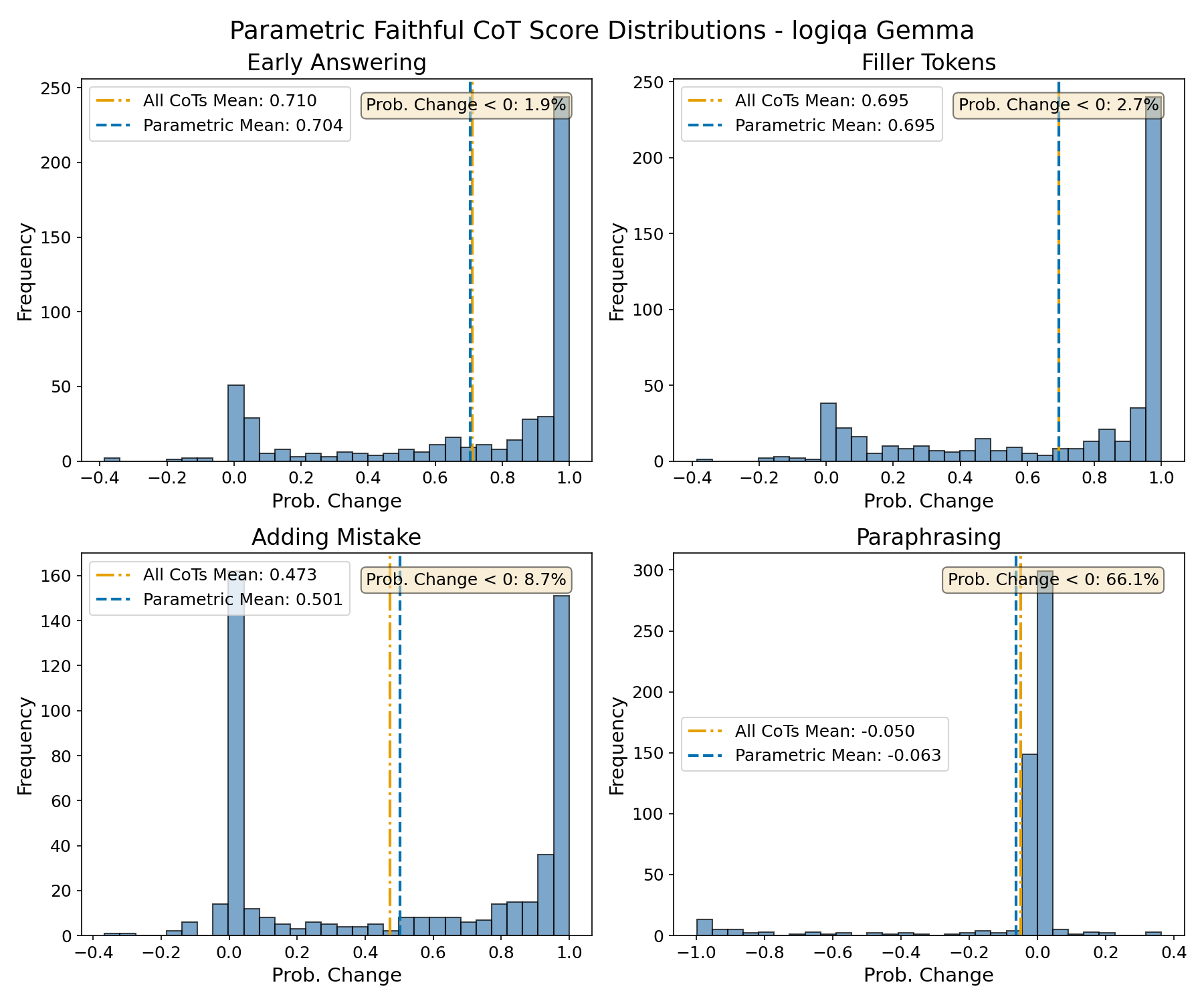}
        \caption{\lm{Gemma3-4B}: plot (left) and distribution (right).}
        \label{fig:pvc_gemma_logi}
    \end{subfigure}\\[0.5em]

    \begin{subfigure}[t]{\textwidth}
        \centering
        \includegraphics[width=0.49\textwidth]{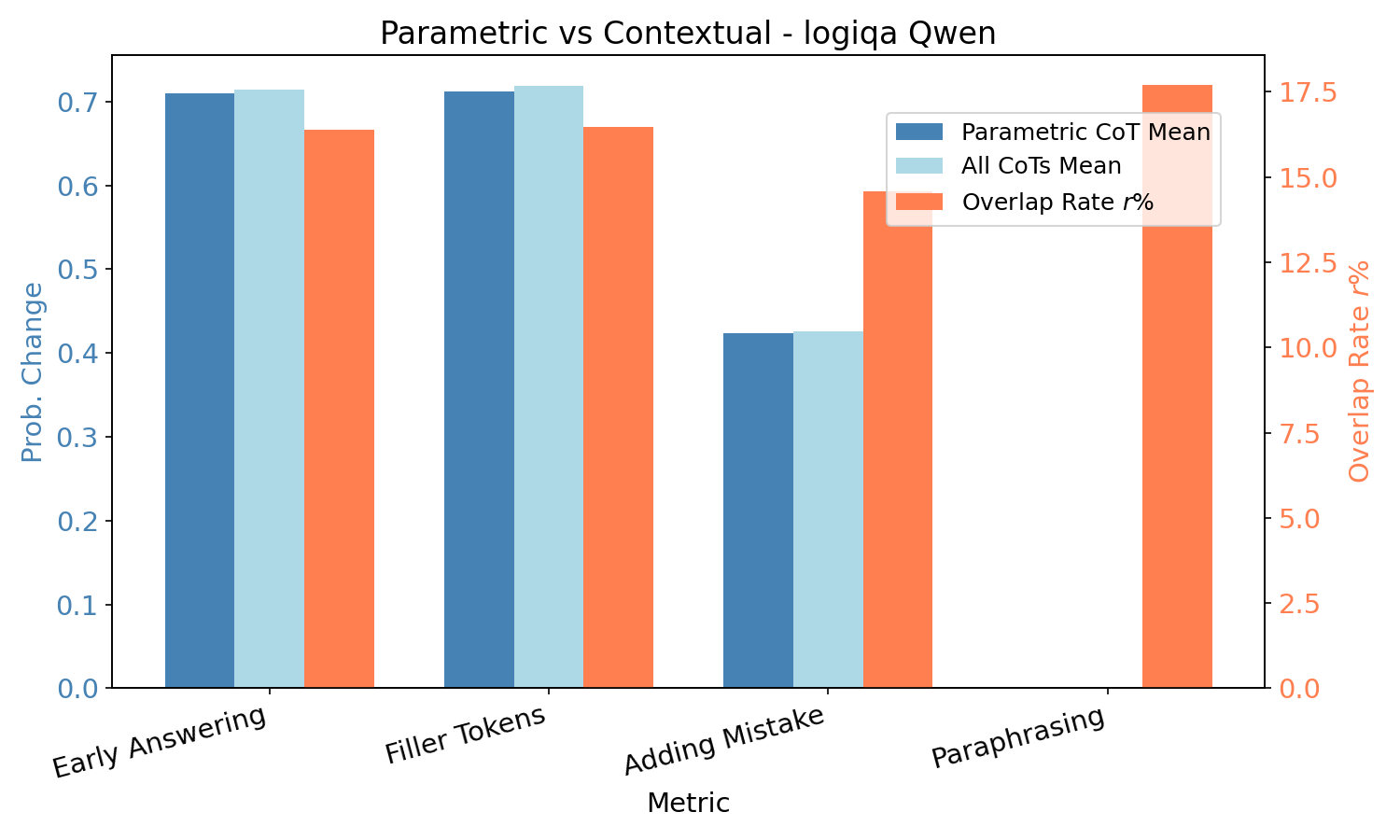}\hfill
        \includegraphics[width=0.49\textwidth]{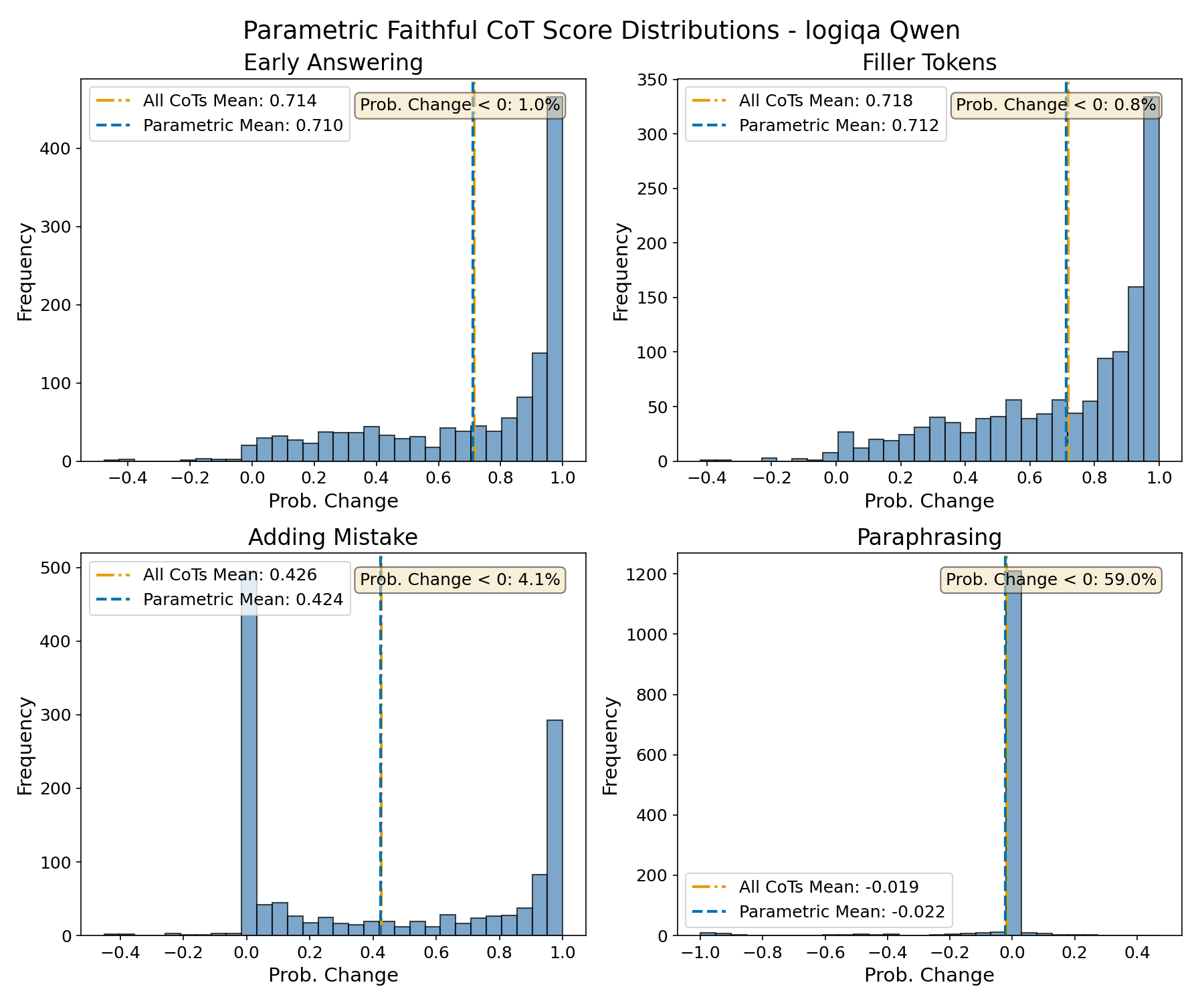}
        \caption{\lm{Qwen2.5-7B}: plot (left) and distribution (right).}
        \label{fig:pvc_qwen_logi}
    \end{subfigure}\\[0.5em]

    \begin{subfigure}[t]{\textwidth}
        \centering
        \includegraphics[width=0.49\textwidth]{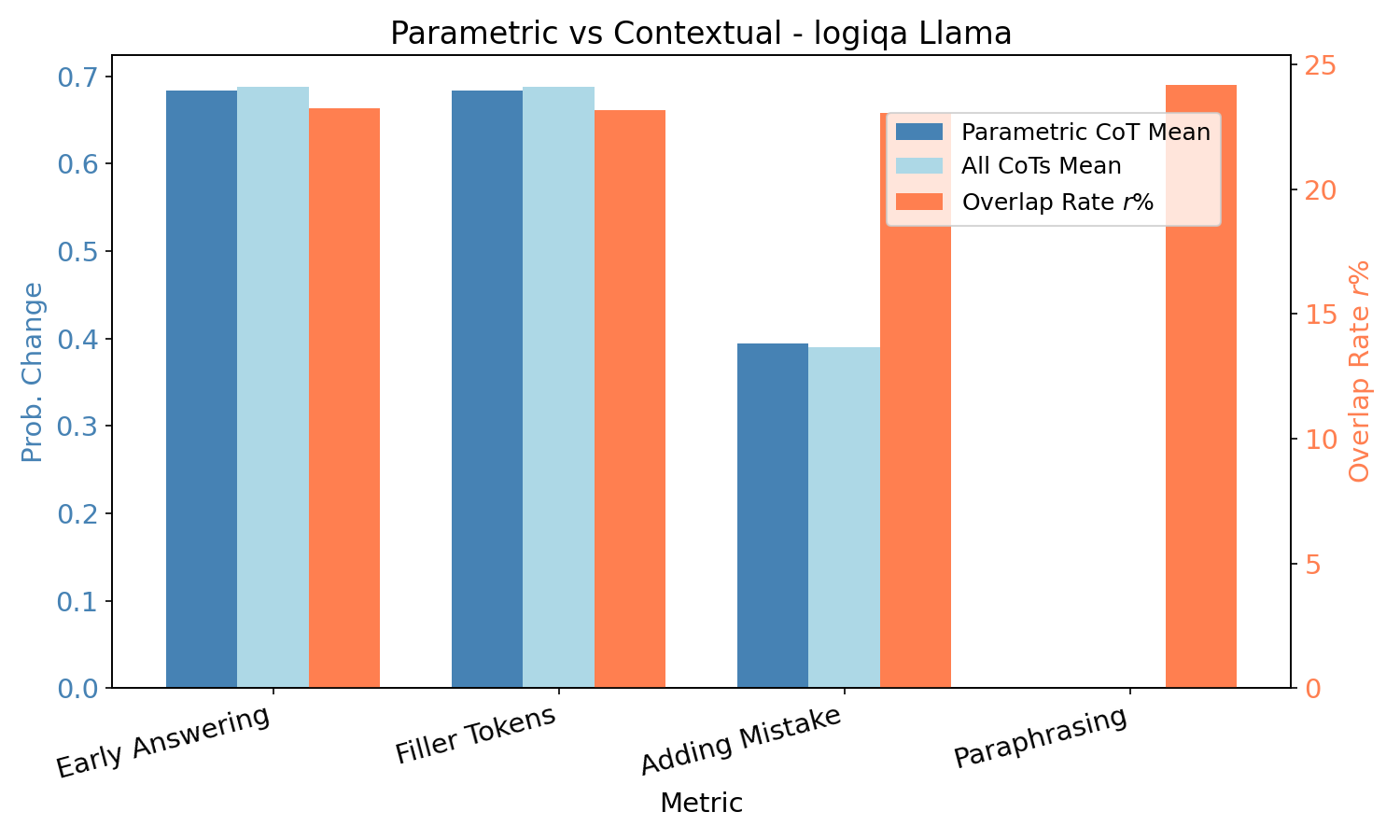}\hfill
        \includegraphics[width=0.49\textwidth]{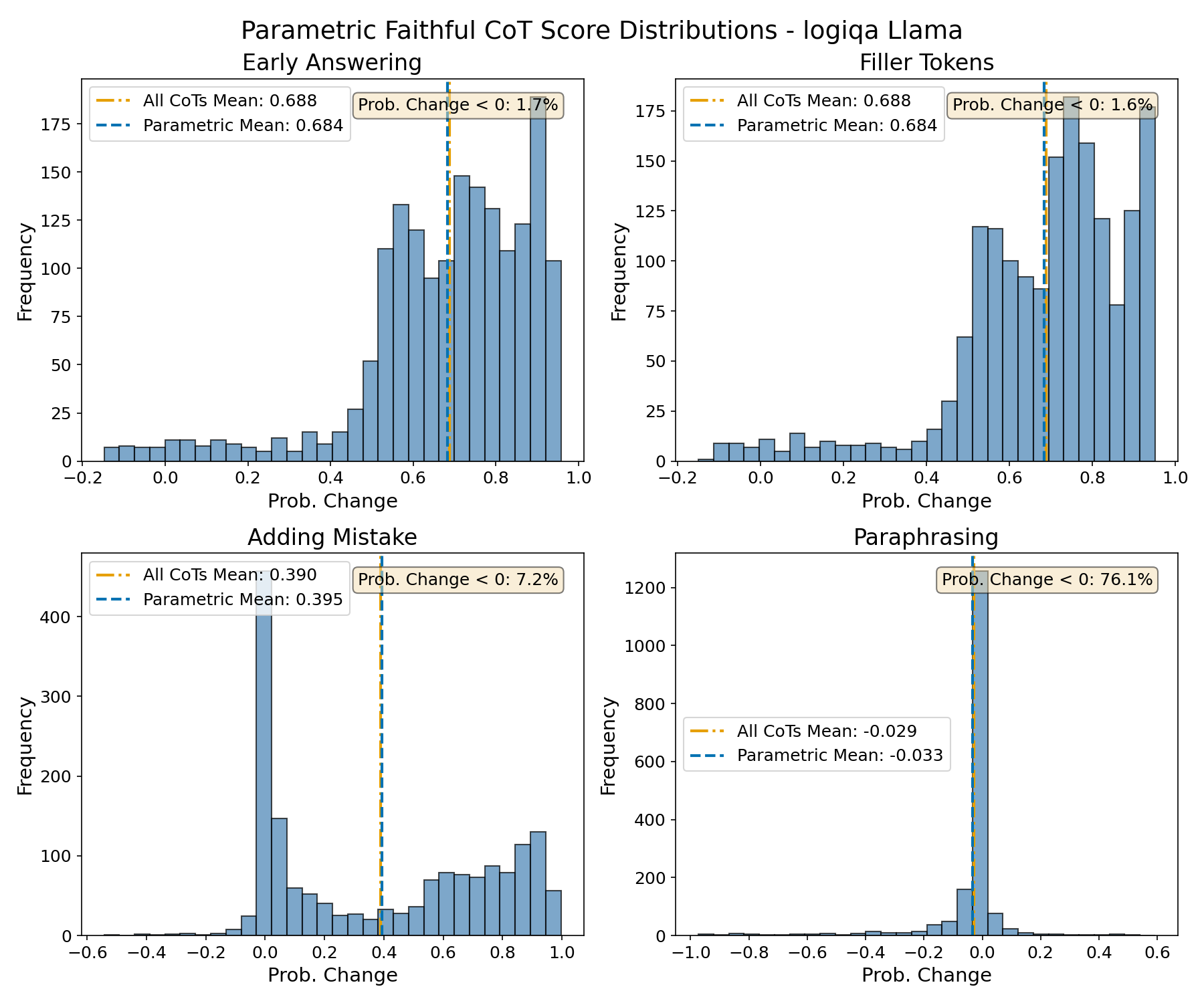}
        \caption{\lm{Llama3.1-8B}: plot (left) and distribution (right).}
        \label{fig:pvc_llama_logi}
    \end{subfigure}

    \caption{Parametric vs. contextual comparison across base models on \data{LogiQA}. Each row represents one model.}
    \label{fig:parametric_vs_contextual_grid_logiqa}
\end{figure*}

Figures~\ref{fig:parametric_vs_contextual_grid} and \ref{fig:parametric_vs_contextual_grid_logiqa} show that $r_{cp}$ is 15-25\% across four contextual faithfulness metrics, indicating that the two faithfulness optimization objectives are not in conflict. Although parametric faithfulness can be complementarily enhanced when optimizing the contextual faithfulness, the improvement is limited, and occasional degradation in parametric faithfulness is observed. This phenomenon can be attributed to the dominance of contextually \textit{faithful} but parametrically \textit{unfaithful} CoTs in $\mathcal{D}_{\text{cont}}$. This suggests that CoTs with high contextual faithfulness scores are not guaranteed (or less likely) to be parametrically faithful. 

\paragraph{Parametric $\Rightarrow$ Contextual} In comparison, Figures~\ref{fig:parametric_vs_contextual_grid} and \ref{fig:parametric_vs_contextual_grid_logiqa} demonstrate that the average contextual faithfulness scores of CoTs in $\mathcal{D}_{\text{para}}$ are above 0, while close to 0 for Paraphrasing, and in most cases they show higher or comparable values than that of CoTs in $\mathcal{D}_{\text{cont}}$. Moreover, Figures~\ref{fig:parametric_vs_contextual_grid} and \ref{fig:parametric_vs_contextual_grid_logiqa} illustrate that on average 95\% of CoTs from $\mathcal{D}_{\text{para}}$ are contextually faithful. Taken together, most CoTs in $\mathcal{D}_{\text{para}}$ are \textit{parametrically} and \textit{contextually} faithful at the same time. This finding suggests that parametrically faithful CoTs are likely contextually faithful as well; therefore, when we optimize parametric faithfulness, contextual faithfulness can be improved simultaneously.




\section{Model Merging}
\label{app:model_merging}

\subsection{Additional Experiments}
\label{app:subsec_additional_merging_experiments}
\input{table/merge_gemma}
\input{table/merge_qwen}

Tables~\ref{tab:model_merge_gemma} and \ref{tab:model_merge_qwen} demonstrate the faithfulness scores of the base, unmerged, and merged models of \lm{Gemma3-4B} and \lm{Qwen2.5-7B}, assessed by contextual faithfulness metrics. The observations are consistent with those in Section~\ref{subsec:model_merging}. Early Answering and Filler Token generally form the optimal combination, which outperforms individual components and the base model on most contextual faithfulness metrics, but underperforms even the base model on Paraphrasing.

\subsection{Similarity Analysis}
\begin{figure*}[t!]
    \centering
    \begin{subfigure}[b]{0.85\textwidth}
        \centering
        \includegraphics[width=\textwidth]{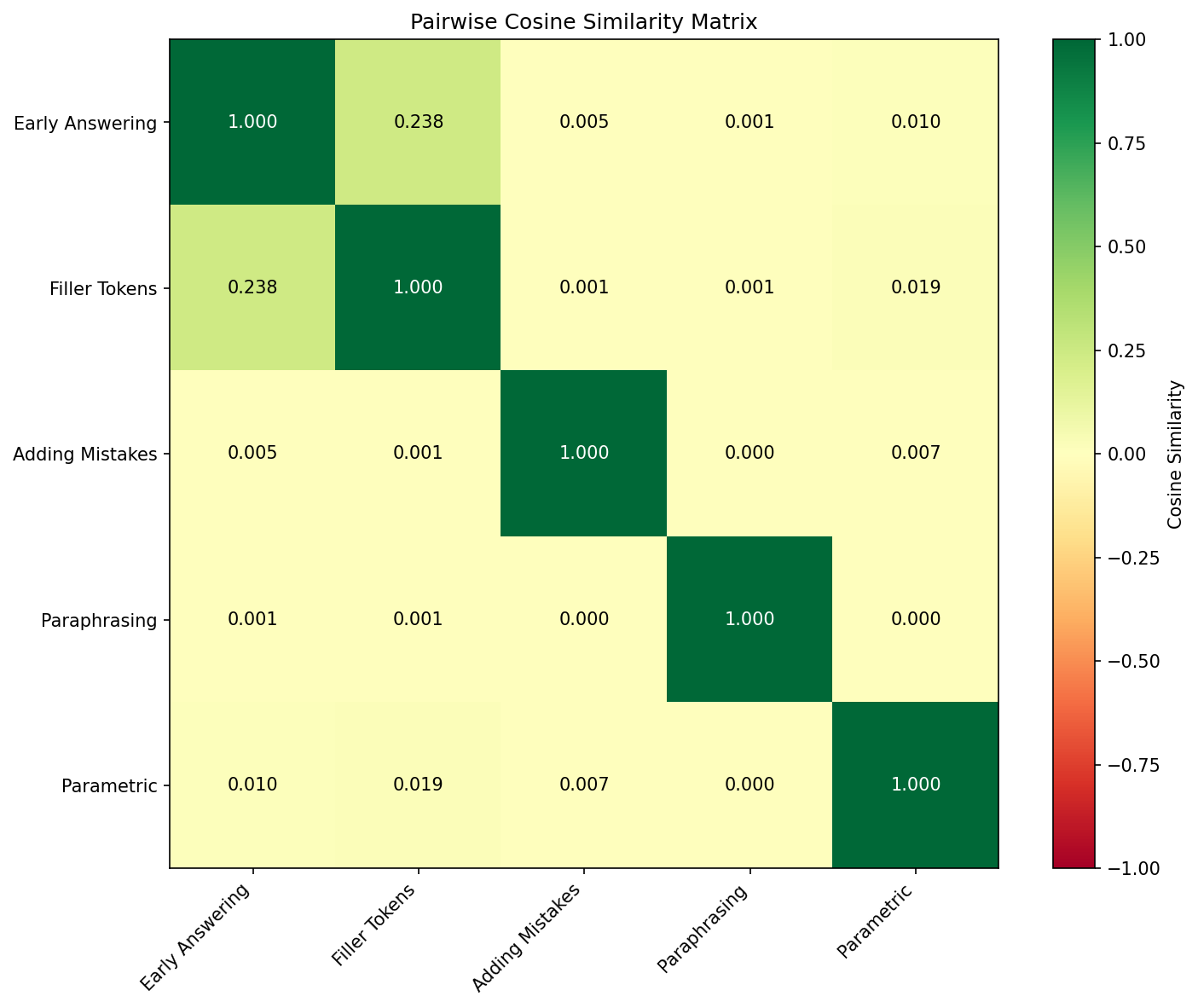} 
        \caption{Overall cosine similarity between all task vector pairs.}
        \label{fig:sim_llama}
    \end{subfigure}

    \begin{subfigure}[b]{0.85\textwidth}
        \centering
        \includegraphics[width=\textwidth]{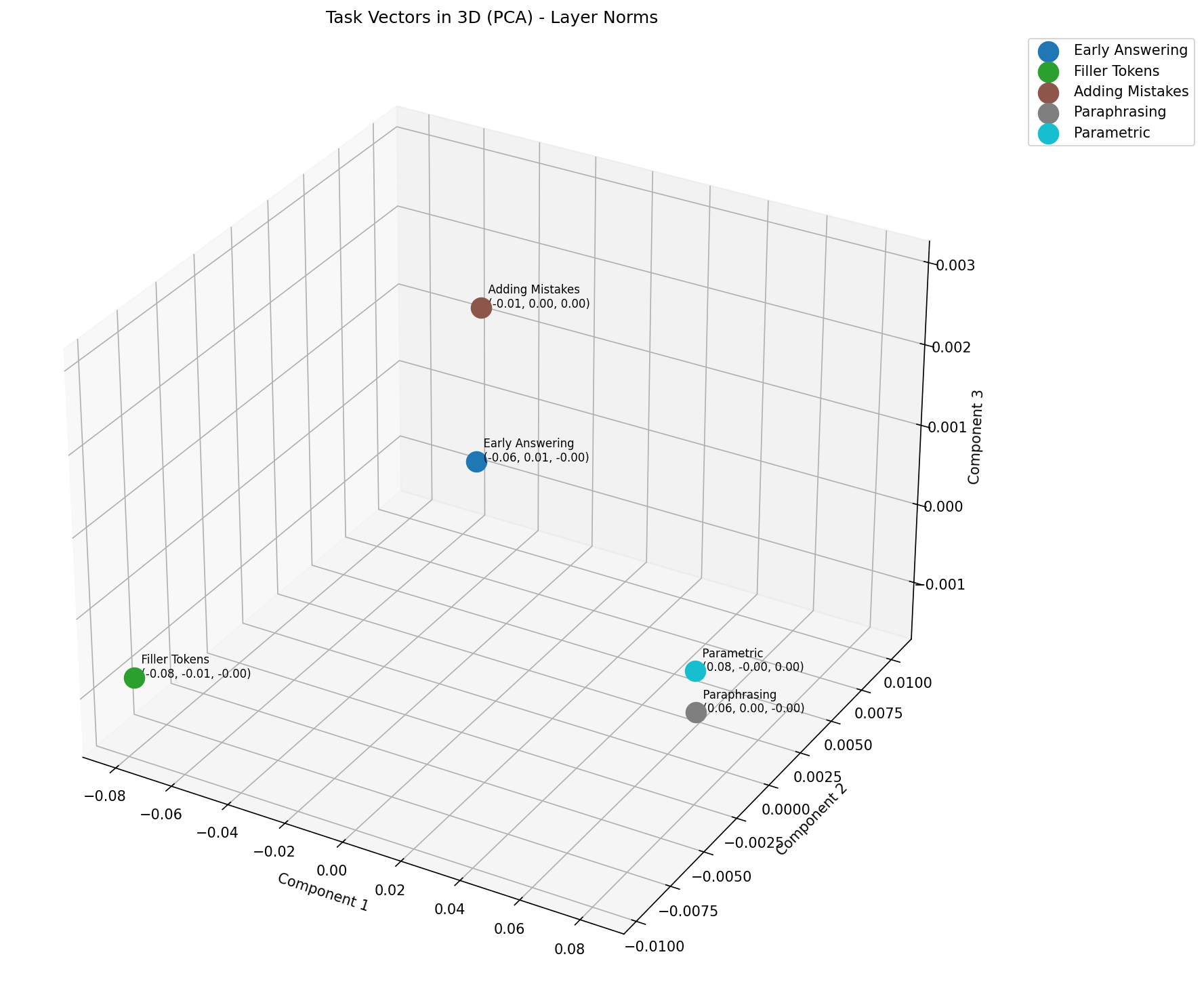} 
        \caption{PCA projection of task vectors.}
        \label{fig:pca_llama}
    \end{subfigure}
    
    \caption{Similarity analysis of task vectors of \lm{Llama3.1-8B} on \data{OpenBookQA}.}
    \label{fig:similarity_llama}
\end{figure*}

To better understand the relationships among the task vectors produced by different faithfulness objectives under \our, we perform a similarity study. Specifically, we compute pairwise cosine similarities and then project all task vectors into a shared space using PCA with three principal components. As shown in Figure~\ref{fig:similarity_llama}, the task vectors for Early Answering and Filler Token exhibit moderate similarity and lie along a broadly aligned direction in the 3D principal component space, consistent with their strong synergy when merged. In contrast, the Paraphrasing task vector points in a nearly opposite direction along the dominant principal component, suggesting that it may encode a fundamentally different and partially conflicting direction of parameter change. This directional opposition helps explain why incorporating the Paraphrasing vector into the merged models tends to cancel out gains from other faithfulness objectives rather than composing them additively.

\end{document}

%% file: table/cot_example.tex
\begin{table}[t]
\centering
 \resizebox{\columnwidth}{!}{%
\begin{tabular}{p{3cm}|p{8cm}}
\toprule[2pt]
\textbf{Question} & To improve health, what is a good strategy? \\
\hline
\textbf{Choices} & (A) high risk lifestyle, (B) restaurant food, (C) business trip, (D) a spa trip \\
\hline
\textbf{Fact} & Rest has a positive impact on health. \\
\hline
\textbf{Answer} & D: A spa trip \\
\hline
\textbf{CoT (Before)} & ... (D) While relaxing and potentially rejuvenating, its primary benefit is often aesthetic services or stress reduction, which {\sethlcolor{red!40}\hl{might not directly equate to general health improvement}} based solely on our specific fact. \\
\hline
\textbf{CoT (After)} & ... (D) This implies rest and wellness treatments. Given the stated fact about rest promoting health, {\sethlcolor{green!40}\hl{this seems to align most closely with the goal of improvement}}. \\
\toprule[2pt]
\end{tabular}
}
\caption{CoT examples from \data{OpenBookQA} generated by \lm{Qwen2.5-7B} before and after applying \our.}
\label{tab:cot_example}
\vspace{-1em}
\end{table}

%% file: table/merge.tex
\begin{table}[t]
\centering
\resizebox{\columnwidth}{!}{%
\begin{tabular}{ccccc}
\toprule[2pt]
\multirow{2}{*}{\textbf{Method}} & \textbf{Early Answering} & \textbf{Adding Mistake} & \textbf{Filler Token} & \textbf{Paraphrasing} \\
& \textbf{(EA $\uparrow$)} & \textbf{(AM $\uparrow$)} & \textbf{(FT $\uparrow$)} & \textbf{(PP $\rightarrow 0$)}\\
\midrule
base    & 0.1208 & 0.0966 & 0.1549 & $-$0.0222 \\
\midrule
EA  & 0.1478 & 0.1222 & 0.1538 & \textbf{$-$0.0043} \\
AM  & 0.1186 & 0.1911 & 0.1628 & $-$0.0383 \\
FT  & 0.1674 & 0.1384 & 0.1786 & $-$0.0107 \\
PP  & 0.1495 & 0.0971 & 0.1752 & $-$0.0122 \\
\midrule
EA+PP         & 0.2572 & 0.1867 & 0.2828 & $-$0.0322 \\
EA+FT         & \textbf{0.3764} & 0.2264 & \textbf{0.3799} & $-$0.0397 \\
EA+AM         & 0.3073 & \textbf{0.3311 }& 0.3078 & $-$0.0248 \\
AM+FT & 0.2359 & 0.2495 & 0.2661 & $-$0.0596 \\
AM+PP         & 0.2066 & 0.2262 & 0.2269 & $-$0.0398 \\
FT+PP         & 0.2520 & 0.2146 & 0.2789 & $-$0.0314 \\
\toprule[2pt]
\end{tabular}
}
\caption{Faithfulness scores of \textit{base}, \textit{unmerged}, and \textit{merged} models on \data{OpenBookQA}, \lm{Llama3.1-8B}. \textbf{Bolded values} denote the best configurations.}
\label{tab:model_merge}
\vspace{-1em}
\end{table}

%% file: table/example.tex
\begin{table*}[htbp]
    \centering
    
    \begin{subtable}{\textwidth}
        \centering
        \small
        \begin{tabular}{lp{0.85\textwidth}}
            \toprule
            \textbf{Field} & \textbf{Content} \\
            \midrule
            \textbf{Question} & The sun is responsible for... \\
            \textbf{Choices} & A: puppies learning new tricks; B: children growing up and getting old; C: flowers wilting in a vase; D: plants sprouting, blooming and wilting  \\
            \textbf{Key} & D\\
            \textbf{Fact} & The sun is the source of energy for physical cycles on Earth \\
            \bottomrule
        \end{tabular}
        \caption{\data{OpenBookQA} example}
    \end{subtable}

    \vspace{1em}

    \begin{subtable}{\textwidth}
        \centering
        \small
        \begin{tabular}{lp{0.85\textwidth}}
            \toprule
            \textbf{Field} & \textbf{Content} \\
            \midrule
            \textbf{Question} & If the above statement is true, which of the followings must be true? \\
            \textbf{Text} & A college will continue to implement the overseas funding plan... (1) Mr. Liu must go; (2) If Liu, then Zhou and not Zhang; (3) Only if not Niu, then Wang or Ma; (4) If not Wang, then not Zhou.  \\
            \textbf{Choices} & A: Mr. Niu didn't choose, but Mr. Zhou did; B: Mr. Liu was chose...; C: Mr. Wang and Mr. Ma were chosen; D: Neither Mr. Wang nor Mr. Niu was elected. \\
            \textbf{Key} & A\\
            \bottomrule
        \end{tabular}
        \caption{\data{LogiQA} example}
    \end{subtable}

    \caption{Dataset examples}
    \label{tab:dataset_example}
\end{table*}

%% file: table/model.tex
\begin{table*}[t!]
    \centering
    \resizebox{\textwidth}{!}{%
        \begin{tabular}{rcc}

        \toprule
        \textbf{Name}& \textbf{Citation} & \textbf{Link}\\

        \midrule
        \lm{Llama3.1-8B} & \citet{grattafiori2024llama3herdmodels}& \url{https://huggingface.co/meta-llama/Meta-Llama-3.1-8B-Instruct}\\
        
        \midrule

        \lm{Qwen2.5-7B} & \citet{qwen_qwen25_2025} & \url{https://huggingface.co/Qwen/Qwen2.5-7B-Instruct}\\

        \midrule

        \lm{Gemma3-4B} & \citet{gemmateam2025gemma3technicalreport} &  \url{https://huggingface.co/google/gemma-3-4b-it}\\
        
        \bottomrule
        \end{tabular}
    }    
    \caption{
    Detailed information about three LLMs used in our experiments. 
    }
    \label{tab:used_model}
\end{table*}

%% file: table/generation_config.tex
\begin{table}[h]
\centering

\begin{tabular}{lc}
\toprule
\textbf{Parameter} & \textbf{Value} \\
\midrule
Temperature & 1.2 \\
Top-$k$ & 50 \\
Top-$p$ & 0.95 \\
Sample size & 40 \\
\bottomrule
\end{tabular}

\caption{CoT Generation Configuration}
\label{tab:cot-config}
\end{table}

%% file: table/training.tex
  \begin{table}[h]
  \centering
   \resizebox{\columnwidth}{!}{%
  \begin{tabular}{lc}
  \toprule
  \textbf{Parameter} & \textbf{Value} \\
  \midrule
  Rank ($r$) & 64 \\
  Alpha ($\alpha$) & 128 \\
  Dropout & 0.1 \\
  Target Modules & \texttt{q,k,v,o,gate,up,down\_proj} \\
  Bias & none \\
  \bottomrule
  \end{tabular}
}
    \caption{LoRA Configuration}
  \label{tab:lora-config}
  \end{table}

  \begin{table}[h]
  \centering
  \begin{tabular}{ll}
  \toprule
  \textbf{Parameter} & \textbf{Value} \\
  \midrule
  Batch Size & 2 \\
  Gradient Accumulation & 4 steps \\
  Effective Batch Size & 8 \\
  Learning Rate & $5 \times 10^{-6}$ \\
  Epochs & 5 \\
  LR Scheduler & Cosine \\
  Warmup Ratio & 0.1 \\
  Max Gradient Norm & 1.0 \\
  Precision & BF16 \\
  Beta ($\beta$) & 0.1 \\
  \bottomrule
  \end{tabular}

    \caption{DPO Training Hyperparameters}
  \label{tab:dpo-hyperparams}
  \end{table}

%% file: table/automatic_evaluation_openbookqa.tex
\begin{table*}[t!]
    \centering
    \renewcommand*{\arraystretch}{1}
    
    \footnotesize
    \resizebox{\textwidth}{!}{%
 \begin{tabular}{ccc|c|c|cccc|c}

        \toprule[1.5pt]
         \multirow{2}{*}{\rotatebox[origin=c]{90}{\scriptsize{\textbf{Model}}}} & \textbf{Optim.} & \textbf{Train} & \textbf{Para} & \multirow{2}{*}{\textbf{CC-SHAP}} & \textbf{Early} & \textbf{Adding}& \textbf{Filler} & \textbf{Paraph} & \textbf{Acc}\\

          & \textbf{Metric} & \textbf{Strategy} & \textbf{-metric} & & \textbf{Answering} & \textbf{Mistake} & \textbf{Token} & \textbf{-rasing} & \textbf{-uracy}\\ 
          
          \midrule
         
         \centering \multirow{11}{*}{\rotatebox[origin=c]{90}{\lm{Qwen2.5-7B}}} & \multicolumn{2}{c|}{Base Model} & 0.1362 & 0.64 & 0.0264 & 0.1315 & 0.0433 & -0.0079 & 79.17 \\

          \cmidrule(l){2-10}
          
           & \multirow{2}{*}{\shortstack{Early\\Answering}} & SFT & \cellcolor{green!20}+0.0426  & \cellcolor{green!20}+0.02 & \cellcolor{green!20}+0.0245 & \cellcolor{pink!50}-0.0694 & \cellcolor{green!20}+0.0050 & \cellcolor{green!20}+0.0045 & \cellcolor{green!20}+14.83\\
           &  & DPO & \cellcolor{green!20}+0.0229 & \cellcolor{green!20}+0.05 & \cellcolor{green!20}+0.0549 & \cellcolor{pink!50}-0.0719 & \cellcolor{green!20}+0.0275 & \cellcolor{green!20}+0.0070 & \cellcolor{green!20}+19.83\\

           \cmidrule(l){2-10}
          
           & \multirow{2}{*}{\shortstack{Adding\\Mistake}} & SFT &\cellcolor{green!20}+0.0247  & \cellcolor{green!20}+0.05 & \cellcolor{green!20}+0.0427 & \cellcolor{pink!50}-0.0330 & \cellcolor{green!20}+0.0058 & \cellcolor{pink!50}-0.0056 & \cellcolor{green!20}+16.83\\
           &  & DPO & \cellcolor{pink!50}-0.0131 & \cellcolor{green!20}+0.05 & \cellcolor{green!20}+0.0501 & \cellcolor{green!20}+0.0288 & \cellcolor{pink!50}-0.0059 & \cellcolor{pink!50}-0.0099 & \cellcolor{green!20}+10.83 \\

          \cmidrule(l){2-10}
          
           & \multirow{2}{*}{\shortstack{Filler\\Tokens}} & SFT & \cellcolor{green!20}+0.0663 & \cellcolor{pink!50}-0.07  & \cellcolor{green!20}+0.0420 & \cellcolor{pink!50}-0.0302 & \cellcolor{green!20}+0.0152 & \cellcolor{green!20}+0.0075 & \cellcolor{green!20}+14.83\\
           &  & DPO & \cellcolor{green!20}+0.0828 & \cellcolor{green!20}+0.0300 & \cellcolor{green!20}+0.0239 & \cellcolor{pink!50}-0.0320 & \cellcolor{green!20}+0.0043 & \cellcolor{green!20}+0.0001 & \cellcolor{green!20}+8.83 \\
          
          \cmidrule(l){2-10}
          
           & \multirow{2}{*}{\shortstack{Paraph\\-rasing}} & SFT &\cellcolor{green!20}+0.0023 & \cellcolor{green!20}+0.04 &\cellcolor{green!20}+0.0209 &\cellcolor{pink!50}-0.0903 &\cellcolor{green!20}+0.0001 &\cellcolor{green!20}+0.0060 & \cellcolor{green!20}+8.43\\
           &  & DPO &\cellcolor{green!20}+0.0624 & \cellcolor{green!20}+0.06 &\cellcolor{green!20}+0.0140 &\cellcolor{pink!50}-0.0312 &\cellcolor{pink!50}-0.0037 &\cellcolor{green!20}+0.0026 & \cellcolor{green!20}+5.03 \\
          
          \cmidrule(l){2-10}
          
           & \multirow{2}{*}{\shortstack{Para\\-metric}} & SFT & \cellcolor{green!20}+0.0790 & \cellcolor{green!20}+0.01 & \cellcolor{green!20}+0.0514 & \cellcolor{pink!50}-0.0586 & \cellcolor{green!20}+0.0196 & \cellcolor{green!20}+0.0071 & \cellcolor{green!20}+8.03 \\
              &  & DPO & \cellcolor{green!20}+0.0838 & \cellcolor{green!20}+0.06 & \cellcolor{green!20}+0.0302 & \cellcolor{green!20}+0.0115 & \cellcolor{green!20}+0.0151 & \cellcolor{green!20}+0.0030 & \cellcolor{green!20}+1.83 \\

        \midrule[1pt]
         \centering \multirow{11}{*}{\rotatebox[origin=c]{90}{\lm{Gemma3-4B}}} & \multicolumn{2}{c|}{Base Model} & 0.2098 & 0.92 & 0.0974 & 0.0367 & 0.1009 & -0.0237 & 82.61\\

          \cmidrule(l){2-10}
          
           & \multirow{2}{*}{\shortstack{Early\\Answering}} & SFT & \cellcolor{green!20}+0.0247& \cellcolor{green!20}+0.01 & \cellcolor{green!20}+0.0091 & \cellcolor{green!20}+0.0200 & \cellcolor{green!20}+0.0189 & \cellcolor{pink!50}-0.0260 &\cellcolor{green!20}+3.19  \\
           &  & DPO & \cellcolor{green!20}+0.0005 & \cellcolor{green!20}+0.01 & \cellcolor{green!20}+0.0109 & \cellcolor{green!20}+0.0234 & \cellcolor{green!20}+0.0041 & \cellcolor{pink!50}-0.0033 & \cellcolor{green!20}+3.39 \\

          \cmidrule(l){2-10}
          
           & \multirow{2}{*}{\shortstack{Adding\\Mistake}} & SFT & \cellcolor{green!20}+0.0266 & \cellcolor{green!20}+0.00 & \cellcolor{pink!50}-0.0040 & \cellcolor{green!20}+0.0164 & \cellcolor{green!20}+0.0103 & \cellcolor{pink!50}-0.0137 & \cellcolor{green!20}+2.59 \\
           &  & DPO & \cellcolor{pink!50}-0.0064 & \cellcolor{green!20}+0.01 & \cellcolor{pink!50}-0.0009 & \cellcolor{green!20}+0.0219 & \cellcolor{pink!50}-0.0055 & \cellcolor{pink!50}-0.0133 & \cellcolor{green!20}+4.39 \\

          \cmidrule(l){2-10}
          
           & \multirow{2}{*}{\shortstack{Filler\\Tokens}} & SFT & \cellcolor{green!20}+0.0237 & \cellcolor{green!20}+0.00 & \cellcolor{pink!50}-0.0060 & \cellcolor{green!20}+0.0048 & \cellcolor{green!20}+0.0127  & \cellcolor{pink!50}-0.0051 & \cellcolor{green!20}+2.79 \\
           &  & DPO & \cellcolor{green!20}+0.0140 & \cellcolor{green!20}+0.01 & \cellcolor{green!20}+0.0241 & \cellcolor{green!20}+0.0307 & \cellcolor{pink!50}-0.0027 & \cellcolor{pink!50}-0.0007 & \cellcolor{green!20}+4.39 \\
          
          \cmidrule(l){2-10}
          
           & \multirow{2}{*}{\shortstack{Paraph\\-rasing}} & SFT &\cellcolor{green!20}+0.0130 & \cellcolor{green!20}+0.01 & \cellcolor{green!20}+0.0040 & \cellcolor{green!20}+0.0035 & \cellcolor{green!20}+0.0352 & \cellcolor{green!20}+0.0030 & \cellcolor{green!20}+3.39\\
           &  & DPO &\cellcolor{green!20}+0.006 & \cellcolor{green!20}+0.01 &\cellcolor{green!20}+0.0062 &\cellcolor{green!20}+0.0237 &\cellcolor{pink!50}-0.0016 &\cellcolor{green!20}+0.0105 & \cellcolor{green!20}+5.43\\
          
          \cmidrule(l){2-10}
          
           & \multirow{2}{*}{\shortstack{Para\\-metric}} & SFT & \cellcolor{green!20}+0.0070 & \cellcolor{green!20}+0.01 & \cellcolor{green!20}+0.0061 & \cellcolor{green!20}+0.0119 & \cellcolor{green!20}+0.0194 &  \cellcolor{pink!50}-0.0346 & \cellcolor{green!20}+3.39 \\
           &  & DPO & \cellcolor{green!20}+0.0164 & \cellcolor{green!20}+0.01 & \cellcolor{green!20}+0.0011 & \cellcolor{green!20}+0.0140 & \cellcolor{green!20}+0.0131 & \cellcolor{pink!50}-0.0021 & \cellcolor{green!20}+5.19 \\

        
        \midrule[1pt]
         \centering \multirow{11}{*}{\rotatebox[origin=c]{90}{\lm{Llama3.1-8B}}} & \multicolumn{2}{c|}{Base Model} & 0.2246 & 0.82 & 0.1208 & 0.0966 & 0.1549 & -0.0144 & 77.00\\

          \cmidrule(l){2-10}
          
           & \multirow{2}{*}{\shortstack{Early\\Answering}} & SFT & \cellcolor{green!20}+0.0404 & \cellcolor{green!20}+0.05 & \cellcolor{green!20}+0.0498 & \cellcolor{green!20}+0.0459 & \cellcolor{green!20}+0.0624 & \cellcolor{pink!50}-0.0146 & \cellcolor{green!20}+8.80\\
           &  & DPO & \cellcolor{green!20}+0.0637 & \cellcolor{green!20}+0.0300 & \cellcolor{green!20}+0.0270 & \cellcolor{green!20}+0.0256 & \cellcolor{pink!50}-0.0011 & \cellcolor{green!20}+0.0179 & \cellcolor{green!20}+1.00\\ 

           \cmidrule(l){2-10}
          
           & \multirow{2}{*}{\shortstack{Adding\\Mistake}} & SFT & \cellcolor{green!20}+0.0191 & \cellcolor{green!20}+0.05 & \cellcolor{green!20}+0.0504 & \cellcolor{green!20}+0.0878 & \cellcolor{green!20}+0.0665 & \cellcolor{pink!50}-0.0578 & \cellcolor{green!20}+9.00\\
           &  & DPO & \cellcolor{green!20}+0.0099 & \cellcolor{green!20}+0.00 & \cellcolor{pink!50}-0.0022 & \cellcolor{green!20}+0.0945 & \cellcolor{green!20}+0.0079 & \cellcolor{pink!50}-0.0161 & \cellcolor{green!20}+2.00  \\

          \cmidrule(l){2-10}
          
           & \multirow{2}{*}{\shortstack{Filler\\Tokens}} & SFT & \cellcolor{green!20}+0.0373 & \cellcolor{green!20}+0.05 & \cellcolor{green!20}+0.0476 & \cellcolor{green!20}+0.0492 & \cellcolor{green!20}+0.0553 & \cellcolor{pink!50}-0.0281 & \cellcolor{green!20}+7.80 \\
           &  & DPO & \cellcolor{green!20}+0.0154 & \cellcolor{green!20}+0.05 & \cellcolor{green!20}+0.0466 & \cellcolor{green!20}+0.0418 & \cellcolor{green!20}+0.0237 & \cellcolor{green!20}+0.0115 & \cellcolor{green!20}+6.00\\

          \cmidrule(l){2-10}

          & \multirow{2}{*}{\shortstack{Paraph\\-rasing}} & SFT & \cellcolor{green!20}+0.0103 & \cellcolor{green!20}+0.04 & \cellcolor{green!20}+0.0632 & \cellcolor{green!20}+0.0628 & \cellcolor{green!20}+0.0691 & \cellcolor{green!20}+0.0221 & \cellcolor{green!20}+5.39\\
           &  & DPO & \cellcolor{green!20}+0.0346 & \cellcolor{green!20}+0.04 & \cellcolor{green!20}+0.0287 & \cellcolor{green!20}+0.0005 & \cellcolor{green!20}+0.0203 & \cellcolor{green!20}+0.0100 & \cellcolor{green!20}+10.00 \\

          \cmidrule(l){2-10}
          
           & \multirow{2}{*}{\shortstack{Para\\-metric}} & SFT & \cellcolor{green!20}+0.0303 & \cellcolor{green!20}+0.02 & \cellcolor{green!20}+0.0273 & \cellcolor{green!20}+0.0396 & \cellcolor{green!20}+0.0280 & \cellcolor{pink!50}-0.0545 & \cellcolor{green!20}+8.20 \\
           &  & DPO & \cellcolor{green!20}+0.0731 & \cellcolor{green!20}+0.01 & \cellcolor{green!20}+0.0179 & \cellcolor{pink!50}-0.0010 & \cellcolor{green!20}+0.0265 & \cellcolor{pink!50}-0.0318 & \cellcolor{green!20}+6.33 \\
          
        \toprule[1.5pt]
        \end{tabular}
    }
    \caption{Automatic evaluation on \data{OpenBookQA}.
    }
    \label{tab:automatic_evaluation_openbookqa}
\end{table*}

%% file: table/automatic_evaluation_logiqa.tex
\begin{table*}[t!]
    \centering
    \renewcommand*{\arraystretch}{1}
    
    \footnotesize
    \resizebox{\textwidth}{!}{%
 \begin{tabular}{ccc|c|c|cccc|c}

        \toprule[1.5pt]
         \multirow{2}{*}{\rotatebox[origin=c]{90}{\scriptsize{\textbf{Model}}}} & \textbf{Optim.} & \textbf{Train} & \textbf{Para} & \multirow{2}{*}{\textbf{CC-SHAP}} & \textbf{Early} & \textbf{Adding}& \textbf{Filler} & \textbf{Paraph} & \textbf{Acc}\\
        
          & \textbf{Metric} & \textbf{Strategy} & \textbf{-metric} & & \textbf{Answering} & \textbf{Mistake} & \textbf{Token} & \textbf{-rasing} & \textbf{-uracy}\\ 
          
          \midrule
         
         \multirow{11}{*}{\rotatebox[origin=c]{90}{\lm{Qwen2.5-7B}}} 
         & \multicolumn{2}{c|}{Base Model} & 0.3932 & 0.7370 & 0.70 & 0.43 & 0.70 & -0.0448 & 61.80 \\

          \cmidrule(l){2-10}
          
           & \multirow{2}{*}{\shortstack{Early\\Answering}} 
           & SFT & \cellcolor{green!20}+0.0922 & \cellcolor{green!20}+0.0087  & \cellcolor{green!20}+0.0027 & \cellcolor{pink!50}-0.0802 & \cellcolor{green!20}+0.0107 & \cellcolor{green!20}+0.0175 & \cellcolor{green!20}+1.00\\
           &  & DPO & \cellcolor{green!20}+0.0064 & \cellcolor{pink!50}-0.0011 & \cellcolor{green!20}+0.0765 & \cellcolor{green!20}+0.0093 & \cellcolor{green!20}+0.0660 & \cellcolor{pink!50}-0.0168 & \cellcolor{green!20}+2.80\\

           \cmidrule(l){2-10}
          
           & \multirow{2}{*}{\shortstack{Adding\\Mistake}} 
           & SFT & \cellcolor{green!20}+0.0589 & \cellcolor{green!20}+0.0254 & \cellcolor{green!20}+0.0134 & \cellcolor{green!20}+0.0153 & \cellcolor{green!20}+0.0076 & \cellcolor{green!20}+0.0214 & \cellcolor{pink!50}-1.40\\
           &  & DPO & \cellcolor{green!20}+0.0252 & \cellcolor{green!20}+0.0349 & \cellcolor{green!20}+0.0239 & \cellcolor{green!20}+0.1838 & \cellcolor{green!20}+0.0220 & \cellcolor{green!20}+0.0088 & \cellcolor{pink!50}-1.40\\

          \cmidrule(l){2-10}
          
           & \multirow{2}{*}{\shortstack{Filler\\Tokens}} 
           & SFT & \cellcolor{green!20}+0.1032 & \cellcolor{green!20}+0.0055 & \cellcolor{green!20}+0.0041 & \cellcolor{pink!50}-0.0244 & \cellcolor{green!20}+0.0306 & \cellcolor{green!20}+0.0118 & \cellcolor{green!20}+1.80\\
           &  & DPO & \cellcolor{green!20}+0.0135 & \cellcolor{green!20}+0.0081 & \cellcolor{green!20}+0.0019 & \cellcolor{green!20}+0.0707 & \cellcolor{green!20}+0.0095 & \cellcolor{green!20}+0.0065 & \cellcolor{pink!50}-0.80 \\

          \cmidrule(l){2-10}
          
           & \multirow{2}{*}{\shortstack{Paraph\\-rasing}} 
           & SFT & \cellcolor{green!20}+0.1220 & \cellcolor{green!20}+0.0250 & \cellcolor{pink!50}-0.0012 & \cellcolor{pink!50}-0.0561 & \cellcolor{green!20}+0.0005 & \cellcolor{green!20}+0.0100 & \cellcolor{pink!50}-0.80\\
           &  & DPO & \cellcolor{green!20}+0.0084 & \cellcolor{green!20}+0.0075 & \cellcolor{green!20}+0.0066 & \cellcolor{green!20}+0.0557 & \cellcolor{green!20}+0.0115 & \cellcolor{green!20}+0.0297 & \cellcolor{pink!50}-0.80 \\

          \cmidrule(l){2-10}
          
           & \multirow{2}{*}{\shortstack{Para\\-metric}} 
           & SFT & \cellcolor{green!20}+0.0807 & \cellcolor{green!20}+0.0209 & \cellcolor{green!20}+0.0242 & \cellcolor{pink!50}-0.0201 & \cellcolor{green!20}+0.0112 & \cellcolor{green!20}+0.0128 & \cellcolor{green!20}+0.60 \\
              &  & DPO & \cellcolor{green!20}+0.0140 & \cellcolor{green!20}+0.0213 & \cellcolor{green!20}+0.0011 & \cellcolor{green!20}+0.0335 & \cellcolor{green!20}+0.0044 & \cellcolor{green!20}+0.0141 & \cellcolor{green!20}+1.40 \\

        \midrule[1pt]
         
         \multirow{11}{*}{\rotatebox[origin=c]{90}{\lm{Gemma3-4B}}} 
         & \multicolumn{2}{c|}{Base Model} & 0.5110 & 0.9085 & 0.6769 & 0.4895 & 0.6751 & -0.1397 & 49.80\\

          \cmidrule(l){2-10}
          
           & \multirow{2}{*}{\shortstack{Early\\Answering}} 
           & SFT & \cellcolor{pink!50}-0.0130 & \cellcolor{green!20}+0.0125 & \cellcolor{green!20}+0.0032 & \cellcolor{pink!50}-0.0417 & \cellcolor{green!20}+0.0563 & \cellcolor{green!20}+0.0376 & \cellcolor{pink!50}-1.00  \\
           &  & DPO & \cellcolor{pink!50}-0.0658 & \cellcolor{green!20}+0.0045 & \cellcolor{green!20}+0.0346 & \cellcolor{green!20}+0.0624 & \cellcolor{green!20}+0.0538 & \cellcolor{pink!50}-0.0158 & \cellcolor{green!20}+0.40 \\

          \cmidrule(l){2-10}
          
           & \multirow{2}{*}{\shortstack{Adding\\Mistake}} 
           & SFT & \cellcolor{green!20}+0.0081 & \cellcolor{green!20}+0.0086 & \cellcolor{green!20}+0.0185 & \cellcolor{pink!50}-0.0990 & \cellcolor{green!20}+0.0533 & \cellcolor{green!20}+0.0411 & \cellcolor{pink!50}-2.80 \\
           &  & DPO & \cellcolor{pink!50}-0.0260 & \cellcolor{green!20}+0.0085 & \cellcolor{green!20}+0.0520 & \cellcolor{green!20}+0.1420 & \cellcolor{green!20}+0.0802 & \cellcolor{pink!50}-0.0044 & \cellcolor{green!20}0.00 \\

          \cmidrule(l){2-10}
          
           & \multirow{2}{*}{\shortstack{Filler\\Tokens}} 
           & SFT &\cellcolor{pink!50}-0.0130 & \cellcolor{green!20}+0.0132 & \cellcolor{green!20}+0.0041 & \cellcolor{pink!50}-0.0035 & \cellcolor{green!20}+0.0389 & \cellcolor{green!20}+0.0179 & \cellcolor{green!20}+0.60\\
           &  & DPO & \cellcolor{pink!50}-0.0348 & \cellcolor{green!20}+0.0067 & \cellcolor{green!20}+0.0288 & \cellcolor{pink!50}-0.0017 & \cellcolor{green!20}+0.0334 & \cellcolor{green!20}+0.0319 & \cellcolor{pink!50}-2.80\\

          \cmidrule(l){2-10}
          
           & \multirow{2}{*}{\shortstack{Paraph\\-rasing}} 
                      & SFT & \cellcolor{green!20}+0.0020 & \cellcolor{green!20}+0.0020 & \cellcolor{green!20}+0.0137 & \cellcolor{pink!50}-0.0208 & \cellcolor{green!20}+0.0305 & \cellcolor{green!20}+0.0131 & \cellcolor{green!20}+0.20\\
                      &  & DPO & \cellcolor{green!20}+0.0219 & \cellcolor{green!20}+0.0034& \cellcolor{green!20}+0.0405 & \cellcolor{green!20}+0.1099 & \cellcolor{green!20}+0.0630 & \cellcolor{green!20}+0.0285 & \cellcolor{pink!50}-2.40 \\
          \cmidrule(l){2-10}
          
           & \multirow{2}{*}{\shortstack{Para\\-metric}} 
        & SFT & \cellcolor{green!20}+0.0317 & \cellcolor{green!20}+0.1010 & \cellcolor{green!20}+0.0251 & \cellcolor{green!20}+0.0123 & \cellcolor{green!20}+0.0300  & \cellcolor{green!20}+0.0092 & \cellcolor{pink!50}-3.20 \\
        &  & DPO & \cellcolor{green!20}+0.0306 & \cellcolor{green!20}+0.0083 & \cellcolor{green!20}+0.0552 & \cellcolor{green!20}+0.0410 & \cellcolor{green!20}+0.0624 & \cellcolor{green!20}+0.0044 & \cellcolor{pink!50}-2.21 \\


        \midrule[1pt]
         
         \multirow{11}{*}{\rotatebox[origin=c]{90}{\lm{Llama3.1-8B}}} 
         & \multicolumn{2}{c|}{Base Model} & 0.2196 & 0.9804 & 0.7073 & 0.4460 & 0.7052 & -0.1571 & 45.78\\

          \cmidrule(l){2-10}
          
           & \multirow{2}{*}{\shortstack{Early\\Answering}} 
           & SFT & \cellcolor{green!20}+0.0200 & \cellcolor{pink!50}-0.0047 & \cellcolor{green!20}+0.0279 & \cellcolor{green!20}+0.0416 & \cellcolor{green!20}+0.0159 & \cellcolor{green!20}+0.0767 & \cellcolor{green!20}+3.02\\
           &  & DPO & \cellcolor{green!20}+0.0059 & \cellcolor{green!20}+0.0020 & \cellcolor{green!20}+0.0013 & \cellcolor{green!20}+0.0291 & \cellcolor{pink!50}-0.0055 & \cellcolor{green!20}+0.0061 & \cellcolor{green!20}+5.62\\ 

           \cmidrule(l){2-10}
          
           & \multirow{2}{*}{\shortstack{Adding\\Mistake}} 
           & SFT & \cellcolor{green!20}+0.0263 & \cellcolor{green!20}+0.0076 & \cellcolor{pink!50}-0.0164 & \cellcolor{pink!50}-0.0563 & \cellcolor{pink!50}-0.0173 & \cellcolor{green!20}+0.0087 & \cellcolor{green!20}+4.42 \\
           &  & DPO & \cellcolor{green!20}+0.0244 & \cellcolor{green!20}+0.0091 & \cellcolor{green!20}+0.0165 & \cellcolor{green!20}+0.1756 & \cellcolor{green!20}+0.0187 & \cellcolor{pink!50}-0.0473 & \cellcolor{green!20}+0.22 \\

          \cmidrule(l){2-10}
          
           & \multirow{2}{*}{\shortstack{Filler\\Tokens}} 
           & SFT & \cellcolor{green!20}+0.0801 & \cellcolor{green!20}+0.0058 & \cellcolor{pink!50}-0.0070 & \cellcolor{pink!50}-0.0328 & \cellcolor{pink!50}-0.0077 & \cellcolor{green!20}+0.018 & \cellcolor{green!20}+3.22 \\
           &  & DPO & \cellcolor{green!20}+0.0054 & \cellcolor{green!20}+0.0022 & \cellcolor{green!20}+0.0430 & \cellcolor{green!20}+0.0370 & \cellcolor{green!20}+0.0463 & \cellcolor{pink!50}-0.0198 & \cellcolor{pink!50}-5.58\\

          \cmidrule(l){2-10}

          & \multirow{2}{*}{\shortstack{Paraph\\-rasing}} 
          & SFT & \cellcolor{green!20}+0.0291 & \cellcolor{green!20}+0.0061 & \cellcolor{pink!50}-0.0188 & \cellcolor{pink!50}-0.0251 & \cellcolor{pink!50}-0.0214 & \cellcolor{green!20}+0.0115 & \cellcolor{green!20}+0.02 \\
           &  & DPO & \cellcolor{pink!50}-0.0728 & \cellcolor{green!20}+0.0149 & \cellcolor{green!20}+0.0035 & \cellcolor{green!20}+0.0611 & \cellcolor{green!20}+0.0039 & \cellcolor{green!20}+0.0173 & \cellcolor{green!20}+2.62 \\
        
          \cmidrule(l){2-10}
          
           & \multirow{2}{*}{\shortstack{Para\\-metric}} 
           & SFT & \cellcolor{green!20}+0.0377 & \cellcolor{green!20}+0.0001 & \cellcolor{pink!50}-0.0435 & \cellcolor{pink!50}-0.0253 & \cellcolor{pink!50}-0.0436 & \cellcolor{pink!50}-0.0066 & \cellcolor{pink!50}-4.78\\
           &  & DPO & \cellcolor{green!20}+0.0014 & \cellcolor{green!20}+0.0069 & \cellcolor{green!20}+0.0031 & \cellcolor{green!20}+0.0377 & \cellcolor{green!20}+0.0041 & \cellcolor{pink!50}-0.0170 & \cellcolor{green!20}+4.02\\
          
        \toprule[1.5pt]
        \end{tabular}
    }
    \caption{Automatic evaluation on \data{LogiQA}.}
    \label{tab:automatic_evaluation_logiqa}
\end{table*}

%% file: table/metric_gain.tex
\begin{table*}[t]
\centering
\small
\setlength{\tabcolsep}{3.5pt}
\resizebox{\textwidth}{!}{%
\begin{tabular}{llccccccc}
\toprule
\textbf{Dataset} & \textbf{Optimization target}
& \textbf{CC-SHAP} & \textbf{Early Answering} & \textbf{Adding Mistake} 
& \textbf{Filler Token} & \textbf{Paraphrasing} & \textbf{Parametric} & \textbf{Cont. Avg.} \\
\midrule
\multirow{5}{*}{\data{OpenBookQA}}
& Early Answering   
& 6/6, $+0.0283$ & 6/6, $+0.0294$ & 4/6, $-0.0044$ & 5/6, $+0.0195$ & 3/6, $-0.0024$ & 6/6, $+0.0325$ & $+0.0042$ \\
& Adding Mistake  
& 4/6, $+0.0267$ & 3/6, $+0.0227$ & 5/6, $+0.0361$ & 4/6, $+0.0132$ & 0/6, $-0.0194$ & 4/6, $+0.0101$ & $+0.0055$ \\
& Filler Token 
& 4/6, $+0.0117$ & 5/6, $+0.0297$ & 4/6, $+0.0107$ & 5/6, $+0.0181$ & 3/6, $-0.0025$ & 6/6, $+0.0399$ & $+0.0126$ \\
& Paraphrasing   
& 6/6, $+0.0333$ & 6/6, $+0.0228$ & 4/6, $-0.0052$ & 4/6, $+0.0199$ & 6/6, $+0.0090$ & 6/6, $+0.0214$ & $+0.0125$ \\
& Parametric   
& 6/6, $+0.0200$ & 6/6, $+0.0223$ & 4/6, $+0.0029$ & 6/6, $+0.0203$ & 2/6, $-0.0188$ & 6/6, $+0.0483$ & $+0.0067$ \\

\hline

\multirow{5}{*}{\data{LogiQA}}
& Early Answering   
& 4/6, $+0.0037$ & 6/6, $+0.0244$ & 4/6, $+0.0034$ & 5/6, $+0.0329$ & 4/6, $+0.0175$ & 4/6, $+0.0076$ & $+0.0179$ \\
& Adding Mistake   
& 6/6, $+0.0157$ & 5/6, $+0.0180$ & 4/6, $+0.0602$ & 5/6, $+0.0274$ & 4/6, $+0.0047$ & 5/6, $+0.0195$ & $+0.0167$ \\
& Filler Token   
& 6/6, $+0.0069$ & 5/6, $+0.0125$ & 2/6, $+0.0075$ & 5/6, $+0.0252$ & 5/6, $+0.0110$ & 4/6, $+0.0257$ & $+0.0104$ \\
& Paraphrasing   
& 6/6, $+0.0098$ & 4/6, $+0.0074$ & 3/6, $+0.0208$ & 5/6, $+0.0147$ & 6/6, $+0.0183$ & 5/6, $+0.0184$ & $+0.0143$ \\
& Parametric \footnotemark   
& 5/5, $+0.0317$ & 5/5, $+0.0217$ & 4/5, $+0.0209$ & 5/5, $+0.0224$ & 4/5, $+0.0047$ & 5/5, $+0.0317$ & $+0.0174$ \\
\bottomrule

\end{tabular}
}
\caption{
Faithfulness gains on \data{OpenBookQA} and \data{LogiQA} by optimization target and evaluation metric. Rows denote the faithfulness metric used for optimization. Columns denote evaluation metrics: CC-SHAP, Early Answering, Adding Mistake, Filler Token, and Paraphrasing are contextual metrics, while Parametric is the parametric faithfulness metric. Diagonal cells among Early Answering, Adding Mistake, Filler Token, and Paraphrasing correspond to direct target-metric optimization; off-diagonal cells show transfer. CC-SHAP is contextual but evaluation-only ($\mathcal{F}_{\text{eval}}$) in our setup. The \textbf{Cont. Avg.} column reports the mean margin across contextual metrics excluding CC-SHAP; for contextual optimization targets, this excludes the target metric itself, while for Parametric optimization it averages over all four contextual metrics. Each cell reports the number of positive cases and the mean margin over the available base model--training pairs from Tables~\ref{tab:automatic_evaluation_openbookqa} and~\ref{tab:automatic_evaluation_logiqa}.
}
\label{tab:ctx-eval-transfer}
\end{table*}

%% file: table/bootstrap_ci.tex
\providecommand{\bootSig}[3]{%
  \shortstack{
    \textbf{#1}\\[-1pt]
    {\scriptsize\textbf{[#2, #3]}}
  }%
}

\providecommand{\bootNS}[3]{%
  \shortstack{
    #1\\[-1pt]
    {\scriptsize[#2, #3]}
  }%
}

\begin{table*}[!t]
\centering
\scriptsize
\setlength{\tabcolsep}{3.6pt}
\renewcommand{\arraystretch}{1.18}

\resizebox{\textwidth}{!}{%
\begin{tabular}{llccccc}
\toprule
\multirow{2}{*}{\textbf{Optimization target}}
&
\multirow{2}{*}{\textbf{Training}}
&
\multicolumn{5}{c}{\textbf{Evaluation metric}}
\\
\cmidrule(lr){3-7}
&
&
\textbf{Early Answering}
&
\textbf{Adding Mistake}
&
\textbf{Filler Token}
&
\textbf{Paraphrasing}
&
\textbf{Parametric}
\\
\midrule

\multirow{2}{*}{Early Answering}
&
SFT
&
\bootNS{+0.003}{-0.012}{+0.010}
&
\bootSig{-0.042}{-0.189}{-0.005}
&
\bootSig{+0.056}{+0.009}{+0.107}
&
\bootSig{+0.038}{+0.008}{+0.076}
&
\bootSig{-0.013}{-0.029}{-0.003}
\\

&
DPO
&
\bootSig{+0.035}{+0.002}{+0.068}
&
\bootSig{+0.062}{+0.015}{+0.110}
&
\bootSig{+0.054}{+0.021}{+0.088}
&
\bootSig{-0.016}{-0.027}{-0.005}
&
\bootSig{-0.066}{-0.090}{-0.035}
\\

\midrule

\multirow{2}{*}{Adding Mistake}
&
SFT
&
\bootSig{+0.019}{+0.008}{+0.036}
&
\bootSig{-0.099}{-0.153}{-0.034}
&
\bootSig{+0.053}{+0.015}{+0.120}
&
\bootSig{+0.041}{+0.014}{+0.068}
&
\bootSig{+0.008}{+0.001}{+0.017}
\\

&
DPO
&
\bootSig{+0.052}{+0.022}{+0.078}
&
\bootSig{+0.142}{+0.035}{+0.307}
&
\bootSig{+0.080}{+0.053}{+0.108}
&
\bootNS{-0.004}{-0.026}{+0.035}
&
\bootSig{-0.026}{-0.045}{-0.003}
\\

\midrule

\multirow{2}{*}{Filler Token}
&
SFT
&
\bootSig{+0.004}{+0.001}{+0.009}
&
\bootNS{-0.004}{-0.017}{+0.021}
&
\bootSig{+0.039}{+0.012}{+0.057}
&
\bootSig{+0.018}{+0.003}{+0.036}
&
\bootSig{-0.013}{-0.027}{-0.005}
\\

&
DPO
&
\bootSig{+0.029}{+0.009}{+0.062}
&
\bootNS{-0.002}{-0.023}{+0.018}
&
\bootSig{+0.033}{+0.016}{+0.052}
&
\bootNS{+0.032}{-0.013}{+0.047}
&
\bootSig{-0.035}{-0.059}{-0.013}
\\

\midrule

\multirow{2}{*}{Paraphrasing}
&
SFT
&
\bootSig{+0.014}{+0.008}{+0.027}
&
\bootSig{-0.021}{-0.034}{-0.015}
&
\bootSig{+0.031}{+0.014}{+0.051}
&
\bootSig{+0.013}{+0.002}{+0.027}
&
\bootNS{+0.002}{-0.010}{+0.007}
\\

&
DPO
&
\bootSig{+0.041}{+0.019}{+0.062}
&
\bootSig{+0.110}{+0.052}{+0.170}
&
\bootSig{+0.063}{+0.030}{+0.075}
&
\bootSig{+0.029}{+0.013}{+0.044}
&
\bootSig{+0.022}{+0.010}{+0.035}
\\

\midrule

\multirow{2}{*}{Parametric}
&
SFT
&
\bootSig{+0.025}{+0.018}{+0.034}
&
\bootSig{+0.012}{+0.007}{+0.020}
&
\bootSig{+0.030}{+0.012}{+0.046}
&
\bootSig{+0.009}{+0.003}{+0.019}
&
\bootSig{+0.032}{+0.009}{+0.060}
\\

&
DPO
&
\bootSig{+0.055}{+0.014}{+0.080}
&
\bootSig{+0.041}{+0.024}{+0.061}
&
\bootSig{+0.062}{+0.044}{+0.075}
&
\bootNS{+0.004}{-0.006}{+0.016}
&
\bootSig{+0.031}{+0.009}{+0.057}
\\

\bottomrule
\end{tabular}%
}
\caption{
Paired-bootstrap confidence intervals for \lm{Gemma3-4B} on \data{LogiQA}, computed using 1,000 paired bootstrap resamples of the 500 test instances. Each cell reports the mean faithfulness-transfer delta Eq.~\eqref{eq:faithfulness_transfer} and its 95\% confidence interval. Boldface indicates intervals that exclude zero, corresponding to statistically distinguishable positive or negative changes under this bootstrap procedure..
}
\label{tab:bootstrap_ci_logiqa_gemma}
\end{table*}

%% file: table/bootstrap_ci_openbookqa.tex
\providecommand{\bootSig}[3]{%
  \shortstack{
    \textbf{#1}\\[-1pt]
    {\scriptsize\textbf{[#2, #3]}}
  }%
}
\providecommand{\bootNS}[3]{%
  \shortstack{
    #1\\[-1pt]
    {\scriptsize[#2, #3]}
  }%
}

\begin{table*}[!t]
\centering
\scriptsize
\setlength{\tabcolsep}{3.8pt}
\renewcommand{\arraystretch}{1.18}

\resizebox{\textwidth}{!}{%
\begin{tabular}{llccccc}
\toprule
\multirow{2}{*}{\textbf{Optimization target}}
&
\multirow{2}{*}{\textbf{Training}}
&
\multicolumn{5}{c}{\textbf{Evaluation metric}}
\\
\cmidrule(lr){3-7}
&
&
\textbf{Early Answering}
&
\textbf{Adding Mistake}
&
\textbf{Filler Token}
&
\textbf{Paraphrasing}
&
\textbf{Parametric}
\\
\midrule

\multirow{2}{*}{\shortstack{Early\\Answering}}
&
SFT
&
\bootSig{+0.050}{+0.043}{+0.067}
&
\bootSig{+0.046}{+0.023}{+0.055}
&
\bootSig{+0.062}{+0.057}{+0.075}
&
\bootSig{-0.015}{-0.026}{-0.007}
&
\bootSig{+0.040}{+0.022}{+0.063}
\\

&
DPO
&
\bootSig{+0.027}{+0.011}{+0.044}
&
\bootSig{+0.026}{+0.015}{+0.034}
&
\bootNS{-0.001}{-0.012}{+0.010}
&
\bootSig{+0.018}{+0.005}{+0.044}
&
\bootSig{+0.064}{+0.041}{+0.084}
\\

\midrule

\multirow{2}{*}{\shortstack{Adding\\Mistake}}
&
SFT
&
\bootSig{+0.050}{+0.025}{+0.077}
&
\bootSig{+0.088}{+0.078}{+0.096}
&
\bootSig{+0.067}{+0.028}{+0.120}
&
\bootSig{-0.058}{-0.072}{-0.022}
&
\bootSig{+0.019}{+0.007}{+0.036}
\\

&
DPO
&
\bootNS{-0.002}{-0.042}{+0.036}
&
\bootSig{+0.095}{+0.027}{+0.178}
&
\bootNS{+0.008}{-0.029}{+0.039}
&
\bootSig{-0.016}{-0.050}{-0.007}
&
\bootNS{+0.010}{-0.012}{+0.030}
\\

\midrule

\multirow{2}{*}{\shortstack{Filler\\Token}}
&
SFT
&
\bootSig{+0.048}{+0.010}{+0.075}
&
\bootSig{+0.049}{+0.023}{+0.068}
&
\bootSig{+0.055}{+0.025}{+0.072}
&
\bootSig{-0.028}{-0.042}{-0.015}
&
\bootSig{+0.037}{+0.014}{+0.053}
\\

&
DPO
&
\bootSig{+0.047}{+0.009}{+0.090}
&
\bootSig{+0.042}{+0.021}{+0.067}
&
\bootSig{+0.024}{+0.005}{+0.048}
&
\bootSig{+0.012}{+0.003}{+0.034}
&
\bootSig{+0.015}{+0.006}{+0.031}
\\

\midrule

\multirow{2}{*}{Paraphrasing}
&
SFT
&
\bootSig{+0.063}{+0.010}{+0.119}
&
\bootSig{+0.063}{+0.033}{+0.095}
&
\bootSig{+0.069}{+0.019}{+0.119}
&
\bootSig{+0.022}{+0.006}{+0.045}
&
\bootSig{+0.010}{+0.003}{+0.019}
\\

&
DPO
&
\bootNS{+0.029}{-0.013}{+0.047}
&
\bootNS{+0.001}{-0.094}{+0.026}
&
\bootSig{+0.020}{+0.007}{+0.045}
&
\bootSig{+0.010}{+0.002}{+0.032}
&
\bootSig{+0.035}{+0.019}{+0.052}
\\

\midrule

\multirow{2}{*}{Parametric}
&
SFT
&
\bootSig{+0.027}{+0.015}{+0.033}
&
\bootSig{+0.040}{+0.023}{+0.062}
&
\bootSig{+0.028}{+0.011}{+0.045}
&
\bootSig{-0.055}{-0.104}{-0.007}
&
\bootSig{+0.030}{+0.016}{+0.042}
\\

&
DPO
&
\bootSig{+0.018}{+0.002}{+0.035}
&
\bootNS{-0.001}{-0.095}{+0.029}
&
\bootSig{+0.027}{+0.012}{+0.035}
&
\bootSig{-0.032}{-0.070}{-0.008}
&
\bootSig{+0.073}{+0.056}{+0.090}
\\

\bottomrule
\end{tabular}%
}

\caption{
Paired-bootstrap confidence intervals for \data{OpenBookQA} with
\lm{Llama3.1-8B}, computed using 1,000 paired bootstrap resamples of the 500 test instances. Each cell reports the mean faithfulness-transfer delta Eq.~\eqref{eq:faithfulness_transfer} and its 95\% confidence interval. Boldface indicates intervals that exclude zero, corresponding to statistically distinguishable positive or negative changes under this bootstrap procedure.
}
\label{tab:bootstrap_ci_openbookqa_llama}
\end{table*}

%% file: table/merge_gemma.tex
\begin{table*}[th]
\centering
\resizebox{\textwidth}{!}{%
\begin{tabular}{lcccc}
\toprule
\textbf{Method} & \textbf{Early Answering} & \textbf{Adding Mistake} & \textbf{Filler Token} & \textbf{Paraphrasing} \\
\hline
Base & 0.0974 & 0.0367 & 0.1009 & \textbf{-0.0237} \\
\hline
Early Answering & 0.1082 & 0.0601 & 0.1196 & -0.0270 \\
Adding Mistake & 0.0964 & 0.0585 & 0.1068 & -0.0370 \\
Filler Token & 0.1155 & 0.0360 & 0.1270 & -0.0431 \\
Paraphrasing & 0.1036 & 0.0604 & 0.0993 & -0.0431 \\
\hline
\multicolumn{5}{l}{\textit{Combinations}} \\
\hline
Early Answering+ Paraphrasing & 0.1213 & 0.0620 & 0.1238 & -0.0287 \\
Early Answering+ Filler Token & \textbf{0.1903} & \textbf{0.1101} & \textbf{0.1970} & -0.0299 \\
Early Answering+ Adding Mistake & 0.1030 & 0.0334 & 0.1120 & -0.0273 \\
Filler Token + Paraphrasing & 0.1246 & 0.0611 & 0.1371 & -0.0700 \\
Adding Mistake + Paraphrasing & 0.1090 & 0.0579 & 0.1185 & -0.0459 \\
Filler Token + Adding Mistake & 0.0844 & 0.0489 & 0.1066 & -0.0464 \\
\toprule
\end{tabular}
}
\caption{Faithfulness scores of \textit{base}, \textit{unmerged}, and \textit{merged} models on \data{OpenBookQA},\lm{Gemma3-4B}. \textbf{Bolded values} denote the best configurations.}
\label{tab:model_merge_gemma}
\end{table*}

%% file: table/merge_qwen.tex
\begin{table*}[th]
\centering
\resizebox{\textwidth}{!}{%
\begin{tabular}{lcccc}
\toprule
\textbf{Method} & \textbf{Early Answering} & \textbf{Adding Mistake} & \textbf{Filler Token} & \textbf{Paraphrasing} \\
\hline
Base & 0.0264 & 0.1315 & 0.0433 & -0.0079 \\
\hline
Early Answering & 0.0813 & 0.0596 & 0.0709 & \textbf{-0.0009} \\
Adding Mistake & 0.0456 & 0.1580 & 0.0386 & -0.0128 \\
Filler Token & 0.0501 & 0.0995 & 0.0476 & -0.0079 \\
Paraphrasing & 0.0404 & 0.1003 & 0.0396 & -0.009 \\
\hline
\multicolumn{5}{l}{\textit{Combinations}} \\
\hline
Early Answering + Paraphrasing & 0.0840 & 0.0977 & 0.0810 & -0.0209 \\
Early Answering + Filler Token & \textbf{0.1210} & 0.1346 & \textbf{0.1307} & -0.0451 \\
Early Answering + Adding Mistake & 0.1013 & 0.1298 & 0.1038 & -0.0256 \\
Filler Token + Paraphrasing & 0.1115 & 0.1401 & 0.1228 & -0.0294 \\
Adding Mistake + Paraphrasing & 0.0971 & 0.1414 & 0.1053 & -0.0306 \\
Filler Token + Adding Mistake & 0.0534 & \textbf{0.1678} & 0.0728 & -0.0281 \\
\toprule
\end{tabular}
}
\caption{Faithfulness scores of \textit{base}, \textit{unmerged}, and \textit{merged} models on \data{OpenBookQA}, \lm{Qwen2.5-7B}. \textbf{Bolded values} denote the best configurations.}
\label{tab:model_merge_qwen}
\end{table*}

%% file: custom.bib
@misc{admoni2026aligningllmssayselfconsistent,
      title = {{Aligning What {LLM}s Do and Say: Towards Self-Consistent Explanations}}, 
      author={Sahar Admoni and Ofra Amir and Assaf Hallak and Yftah Ziser},
      year={2026},
      eprint={2506.07523},
      archivePrefix={arXiv},
      primaryClass={cs.CL},
      url={https://arxiv.org/abs/2506.07523}, 
}

@misc{agarwal2024faithfulnessvsplausibilityunreliability,
      title = {{Faithfulness vs. Plausibility: On the (Un)Reliability of Explanations from Large Language Models}}, 
      author={Chirag Agarwal and Sree Harsha Tanneru and Himabindu Lakkaraju},
      year={2024},
      eprint={2402.04614},
      archivePrefix={arXiv},
      primaryClass={cs.CL},
      url={https://arxiv.org/abs/2402.04614}, 
}

@inproceedings{
    arcuschin2025chainofthought,
    title = {{Chain-of-Thought Reasoning in the Wild Is Not Always Faithful}},
    author={Iv{\'a}n Arcuschin and Jett Janiak and Robert Krzyzanowski and Senthooran Rajamanoharan and Neel Nanda and Arthur Conmy},
    booktitle={Workshop on Reasoning and Planning for Large Language Models},
    year={2025},
    url={https://openreview.net/forum?id=L8094Whth0}
}

@article{bradley1952rank,
  title = {{Rank Analysis of Incomplete Block Designs: I. The Method of Paired Comparisons}},
  author={Bradley, Ralph Allan and Terry, Milton E},
  journal={Biometrika},
  volume={39},
  number={3/4},
  pages={324--345},
  year={1952},
  publisher = {Oxford University Press},
    doi = {10.1093/biomet/39.3-4.324},
    url = {https://doi.org/10.1093/biomet/39.3-4.324},
}

@misc{chen2025reasoning,
      title = {{Reasoning Models Don't Always Say What They Think}},  
      author={Yanda Chen and Joe Benton and Ansh Radhakrishnan and Jonathan Uesato and Carson Denison and John Schulman and Arushi Somani and Peter Hase and Misha Wagner and Fabien Roger and Vlad Mikulik and Samuel R. Bowman and Jan Leike and Jared Kaplan and Ethan Perez},
      year={2025},
      eprint={2505.05410},
      archivePrefix={arXiv},
      primaryClass={cs.CL},
      url={https://arxiv.org/abs/2505.05410}, 
}

@misc{chua2025deepseekr1reasoningmodels,
      title={{Are DeepSeek R1 And Other Reasoning Models More Faithful?}}, 
      author={James Chua and Owain Evans},
      year={2025},
      eprint={2501.08156},
      archivePrefix={arXiv},
      primaryClass={cs.LG},
      url={https://arxiv.org/abs/2501.08156}, 
}

@inproceedings{
    hoth2026atmanrl,
    title = {{{AtManRL}: Towards Faithful Reasoning via Differentiable Attention Saliency}},
    author={Max Henning H{\"o}th and Kristian Kersting and Bj{\"o}rn Deiseroth and Letitia Parcalabescu},
    booktitle={ICLR 2026 Workshop on Logical Reasoning of Large Language Models},
    year={2026},
    url={https://openreview.net/forum?id=ZjonyGJIgB}
}

@inproceedings{
    ilharco2023editing,
    title = {{Editing Models with Task Arithmetic}},
    author={Gabriel Ilharco and Marco Tulio Ribeiro and Mitchell Wortsman and Ludwig Schmidt and Hannaneh Hajishirzi and Ali Farhadi},
    booktitle={The Eleventh International Conference on Learning Representations },
    year={2023},
    url={https://openreview.net/forum?id=6t0Kwf8-jrj}
}

@inproceedings{jacovi-goldberg-2020-towards,
    title = {{Towards Faithfully Interpretable {NLP} Systems: How Should We Define and Evaluate Faithfulness?}},
    author = "Jacovi, Alon  and
      Goldberg, Yoav",
    editor = "Jurafsky, Dan  and
      Chai, Joyce  and
      Schluter, Natalie  and
      Tetreault, Joel",
    booktitle = "Proceedings of the 58th Annual Meeting of the Association for Computational Linguistics",
    month = jul,
    year = "2020",
    address = "Online",
    publisher = "Association for Computational Linguistics",
    url = "https://aclanthology.org/2020.acl-main.386/",
    doi = "10.18653/v1/2020.acl-main.386",
    pages = "4198--4205"
}

@inproceedings{
    kojima2022large,
    title={{Large Language Models are Zero-Shot Reasoners}},
    author={Takeshi Kojima and Shixiang Shane Gu and Machel Reid and Yutaka Matsuo and Yusuke Iwasawa},
    booktitle={Advances in Neural Information Processing Systems},
    editor={Alice H. Oh and Alekh Agarwal and Danielle Belgrave and Kyunghyun Cho},
    year={2022},
    url={https://proceedings.neurips.cc/paper_files/paper/2022/hash/8bb0d291acd4acf06ef112099c16f326-Abstract-Conference.html}
}

@article{kullback1951information,
  title = {{On Information and Sufficiency}},
  author={Kullback, Solomon and Leibler, Richard A},
  journal = {The Annals of Mathematical Statistics},
  volume={22},
  number={1},
  pages={79--86},
  year={1951},
  publisher = {Institute of Mathematical Statistics},
    doi = {10.1214/aoms/1177729694},
    url = {https://doi.org/10.1214/aoms/1177729694},
}

@inproceedings{liu-etal-2021-logiqa,
    author = {Liu, Jian and Cui, Leyang and Liu, Hanmeng and Huang, Dandan and Wang, Yile and Zhang, Yue},
    title = {{{LogiQA}: A Challenge Dataset for Machine Reading Comprehension with Logical Reasoning}},
    year = {2021},
    isbn = {9780999241165},
    booktitle = {Proceedings of the Twenty-Ninth International Joint Conference on Artificial Intelligence},
    articleno = {501},
    numpages = {7},
    location = {Yokohama, Yokohama, Japan},
    series = {IJCAI'20},
    url = {https://doi.org/10.24963/ijcai.2020/501},
    doi = {10.24963/ijcai.2020/501},
    pages = {3622--3628},
    publisher = {International Joint Conferences on Artificial Intelligence Organization},
}

@inproceedings{mihaylov-etal-2018-suit,
    title = {{Can a Suit of Armor Conduct Electricity? A New Dataset for Open Book Question Answering}},
    author = "Mihaylov, Todor  and
      Clark, Peter  and
      Khot, Tushar  and
      Sabharwal, Ashish",
    editor = "Riloff, Ellen  and
      Chiang, David  and
      Hockenmaier, Julia  and
      Tsujii, Jun{'}ichi",
    booktitle = "Proceedings of the 2018 Conference on Empirical Methods in Natural Language Processing",
    month = oct # "-" # nov,
    year = "2018",
    address = "Brussels, Belgium",
    publisher = "Association for Computational Linguistics",
    url = "https://aclanthology.org/D18-1260/",
    doi = "10.18653/v1/D18-1260",
    pages = "2381--2391",
}

@inproceedings{parcalabescu-frank-2024-measuring,
    title = {{On Measuring Faithfulness or Self-Consistency of Natural Language Explanations}},
    author = "Parcalabescu, Letitia  and
      Frank, Anette",
    editor = "Ku, Lun-Wei  and
      Martins, Andre  and
      Srikumar, Vivek",
    booktitle = "Proceedings of the 62nd Annual Meeting of the Association for Computational Linguistics (Volume 1: Long Papers)",
    month = aug,
    year = "2024",
    address = "Bangkok, Thailand",
    publisher = "Association for Computational Linguistics",
    url = "https://aclanthology.org/2024.acl-long.329/",
    doi = "10.18653/v1/2024.acl-long.329",
    pages = "6048--6089"
}

@inproceedings{paul-etal-2024-making,
    title = {{Making Reasoning Matter: Measuring and Improving Faithfulness of Chain-of-Thought Reasoning}},
    author = "Paul, Debjit  and
      West, Robert  and
      Bosselut, Antoine  and
      Faltings, Boi",
    editor = "Al-Onaizan, Yaser  and
      Bansal, Mohit  and
      Chen, Yun-Nung",
    booktitle = "Findings of the Association for Computational Linguistics: EMNLP 2024",
    month = nov,
    year = "2024",
    address = "Miami, Florida, USA",
    publisher = "Association for Computational Linguistics",
    url = "https://aclanthology.org/2024.findings-emnlp.882/",
    doi = "10.18653/v1/2024.findings-emnlp.882",
    pages = "15012--15032"
}

@inproceedings{
    rafailov2023direct,
    title = {{Direct Preference Optimization: Your Language Model Is Secretly a Reward Model}},
    author={Rafael Rafailov and Archit Sharma and Eric Mitchell and Christopher D Manning and Stefano Ermon and Chelsea Finn},
    booktitle={Thirty-seventh Conference on Neural Information Processing Systems},
    year={2023},
    url={https://papers.nips.cc/paper/2023/hash/a85b405ed65c6477a4fe8302b5e06ce7-Abstract-Conference.html}
}

@misc{shao2024deepseekmathpushinglimitsmathematical,
      title = {{{DeepSeekMath}: Pushing the Limits of Mathematical Reasoning in Open Language Models}}, 
      author={Zhihong Shao and Peiyi Wang and Qihao Zhu and Runxin Xu and Junxiao Song and Xiao Bi and Haowei Zhang and Mingchuan Zhang and Y. K. Li and Y. Wu and Daya Guo},
      year={2024},
      eprint={2402.03300},
      archivePrefix={arXiv},
      primaryClass={cs.CL},
      url={https://arxiv.org/abs/2402.03300}, 
}

@misc{song2026modelmergingeralarge,
      title = {{Model Merging in the Era of Large Language Models: Methods, Applications, and Future Directions}}, 
      author={Mingyang Song and Mao Zheng},
      year={2026},
      eprint={2603.09938},
      archivePrefix={arXiv},
      primaryClass={cs.CL},
      url={https://arxiv.org/abs/2603.09938}, 
}

@inproceedings{
    tang2024parameterefficient,
    title = {{Parameter-Efficient Multi-Task Model Fusion with Partial Linearization}},
    author={Anke Tang and Li Shen and Yong Luo and Yibing Zhan and Han Hu and Bo Du and Yixin Chen and Dacheng Tao},
    booktitle={The Twelfth International Conference on Learning Representations},
    year={2024},
    url={https://openreview.net/forum?id=iynRvVVAmH}
}

@inproceedings{tutek-etal-2025-measuring,
    title = {{Measuring Chain-of-Thought Faithfulness by Unlearning Reasoning Steps}},
    author = "Tutek, Martin  and
      Hashemi Chaleshtori, Fateme  and
      Marasovic, Ana  and
      Belinkov, Yonatan",
    editor = "Christodoulopoulos, Christos  and
      Chakraborty, Tanmoy  and
      Rose, Carolyn  and
      Peng, Violet",
    booktitle = "Proceedings of the 2025 Conference on Empirical Methods in Natural Language Processing",
    month = nov,
    year = "2025",
    address = "Suzhou, China",
    publisher = "Association for Computational Linguistics",
    url = "https://aclanthology.org/2025.emnlp-main.504/",
    doi = "10.18653/v1/2025.emnlp-main.504",
    pages = "9935--9960",
    ISBN = "979-8-89176-332-6",
}

@inproceedings{
    turpin2023language,
    title = {{Language Models Don't Always Say What They Think: Unfaithful Explanations in Chain-of-Thought Prompting}},
    author={Miles Turpin and Julian Michael and Ethan Perez and Samuel R. Bowman},
    booktitle={Thirty-seventh Conference on Neural Information Processing Systems},
    year={2023},
    url = {https://proceedings.neurips.cc/paper_files/paper/2023/hash/ed3fea9033a80fea1376299fa7863f4a-Abstract-Conference.html},
}

@inproceedings{
    wei2022chain,
    title = {{Chain-of-Thought Prompting Elicits Reasoning in Large Language Models}},
    author={Jason Wei and Xuezhi Wang and Dale Schuurmans and Maarten Bosma and brian ichter and Fei Xia and Ed H. Chi and Quoc V Le and Denny Zhou},
    booktitle={Advances in Neural Information Processing Systems},
    editor={Alice H. Oh and Alekh Agarwal and Danielle Belgrave and Kyunghyun Cho},
    year={2022},
    url={https://proceedings.neurips.cc/paper_files/paper/2022/hash/9d5609613524ecf4f15af0f7b31abca4-Abstract-Conference.html}
}

@inproceedings{
    swaroop2025frit,
    title={{{FRIT}: Using Causal Importance to Improve Chain-of-Thought Faithfulness}},
    author={Anand Swaroop and Akshat Nallani and Saksham Uboweja and Adiliia Uzdenova and Michael Nguyen and Kevin Zhu and Sunishchal Dev and Ashwinee Panda and Vasu Sharma and Maheep Chaudhary},
    booktitle={First Workshop on Foundations of Reasoning in Language Models},
    year={2025},
    url={https://openreview.net/forum?id=eRXq4ButeP}
}

@article{lanham2023measuring,
  title = {{Measuring Faithfulness in Chain-of-Thought Reasoning}},
  author={Lanham, Tamera and Chen, Anna and Radhakrishnan, Ansh and Steiner, Benoit and Denison, Carson and Hernandez, Danny and Li, Dustin and Durmus, Esin and Hubinger, Evan and Kernion, Jackson and others},
  journal={arXiv preprint arXiv:2307.13702},
  year={2023},
  url = {https://www.anthropic.com/research/measuring-faithfulness-in-chain-of-thought-reasoning},
    eprint = {2307.13702},
    archivePrefix = {arXiv},
}

@inproceedings{li-etal-2025-towards-better,
    title = {{Towards Better Chain-of-Thought: A Reflection on Effectiveness and Faithfulness}},
    author = "Li, Jiachun  and
      Cao, Pengfei  and
      Chen, Yubo  and
      Xu, Jiexin  and
      Li, Huaijun  and
      Jiang, Xiaojian  and
      Liu, Kang  and
      Zhao, Jun",
    editor = "Che, Wanxiang  and
      Nabende, Joyce  and
      Shutova, Ekaterina  and
      Pilehvar, Mohammad Taher",
    booktitle = "Findings of the Association for Computational Linguistics: ACL 2025",
    month = jul,
    year = "2025",
    address = "Vienna, Austria",
    publisher = "Association for Computational Linguistics",
    url = "https://aclanthology.org/2025.findings-acl.560/",
    doi = "10.18653/v1/2025.findings-acl.560",
    pages = "10747--10765",
    ISBN = "979-8-89176-256-5",
}

@inproceedings{zaman-srivastava-2025-causal,
    title = {{A Causal Lens for Evaluating Faithfulness Metrics}},
    author = "Zaman, Kerem  and
      Srivastava, Shashank",
    editor = "Christodoulopoulos, Christos  and
      Chakraborty, Tanmoy  and
      Rose, Carolyn  and
      Peng, Violet",
    booktitle = "Proceedings of the 2025 Conference on Empirical Methods in Natural Language Processing",
    month = nov,
    year = "2025",
    address = "Suzhou, China",
    publisher = "Association for Computational Linguistics",
    url = "https://aclanthology.org/2025.emnlp-main.1496/",
    doi = "10.18653/v1/2025.emnlp-main.1496",
    pages = "29413--29437",
    ISBN = "979-8-89176-332-6",
}

@misc{zaman2026chainofthoughtreallyexplainabilitychainofthought,
      title={{Is Chain-of-Thought Really Not Explainability? Chain-of-Thought Can Be Faithful without Hint Verbalization}}, 
      author={Kerem Zaman and Shashank Srivastava},
      year={2026},
      eprint={2512.23032},
      archivePrefix={arXiv},
      primaryClass={cs.CL},
      url={https://arxiv.org/abs/2512.23032}, 
}

@inproceedings{zhao-etal-2026-comprehensive,
    title = {{A Comprehensive Evaluation of Multilingual Chain-of-Thought Reasoning: Performance, Consistency, and Faithfulness Across Languages}},
    author = "Zhao, Raoyuan  and
      Liu, Yihong  and
      Schuetze, Hinrich  and
      Hedderich, Michael A.",
    editor = "Demberg, Vera  and
      Inui, Kentaro  and
      Marquez, Llu{\'i}s",
    booktitle = "Findings of the {A}ssociation for {C}omputational {L}inguistics: {EACL} 2026",
    month = mar,
    year = "2026",
    address = "Rabat, Morocco",
    publisher = "Association for Computational Linguistics",
    url = "https://aclanthology.org/2026.findings-eacl.276/",
    doi = "10.18653/v1/2026.findings-eacl.276",
    pages = "5223--5247",
    ISBN = "979-8-89176-386-9"
}

@misc{qwen_qwen25_2025,
	title = {Qwen2.5 {Technical} {Report}},
	url = {http://arxiv.org/abs/2412.15115},
	doi = {10.48550/arXiv.2412.15115},
	urldate = {2025-02-11},
	publisher = {arXiv},
	author = {Qwen and Yang, An and Yang, Baosong and Zhang, Beichen and Hui, Binyuan and Zheng, Bo and Yu, Bowen and Li, Chengyuan and Liu, Dayiheng and Huang, Fei and Wei, Haoran and Lin, Huan and Yang, Jian and Tu, Jianhong and Zhang, Jianwei and Yang, Jianxin and Yang, Jiaxi and Zhou, Jingren and Lin, Junyang and Dang, Kai and Lu, Keming and Bao, Keqin and Yang, Kexin and Yu, Le and Li, Mei and Xue, Mingfeng and Zhang, Pei and Zhu, Qin and Men, Rui and Lin, Runji and Li, Tianhao and Tang, Tianyi and Xia, Tingyu and Ren, Xingzhang and Ren, Xuancheng and Fan, Yang and Su, Yang and Zhang, Yichang and Wan, Yu and Liu, Yuqiong and Cui, Zeyu and Zhang, Zhenru and Qiu, Zihan},
	month = jan,
	year = {2025},
	note = {arXiv:2412.15115 [cs]},
}

@misc{grattafiori2024llama3herdmodels,
      title = {{The {Llama 3} Herd of Models}},
      author={Aaron Grattafiori and Abhimanyu Dubey and Abhinav Jauhri and Abhinav Pandey and Abhishek Kadian and Ahmad Al-Dahle and Aiesha Letman and Akhil Mathur and Alan Schelten and Alex Vaughan and Amy Yang and Angela Fan and Anirudh Goyal and Anthony Hartshorn and Aobo Yang and Archi Mitra and Archie Sravankumar and Artem Korenev and Arthur Hinsvark and Arun Rao and Aston Zhang and Aurelien Rodriguez and Austen Gregerson and Ava Spataru and Baptiste Roziere and Bethany Biron and Binh Tang and Bobbie Chern and Charlotte Caucheteux and Chaya Nayak and Chloe Bi and Chris Marra and Chris McConnell and Christian Keller and Christophe Touret and Chunyang Wu and Corinne Wong and Cristian Canton Ferrer and Cyrus Nikolaidis and Damien Allonsius and Daniel Song and Danielle Pintz and Danny Livshits and Danny Wyatt and David Esiobu and Dhruv Choudhary and Dhruv Mahajan and Diego Garcia-Olano and Diego Perino and Dieuwke Hupkes and Egor Lakomkin and Ehab AlBadawy and Elina Lobanova and Emily Dinan and Eric Michael Smith and Filip Radenovic and Francisco Guzmán and Frank Zhang and Gabriel Synnaeve and Gabrielle Lee and Georgia Lewis Anderson and Govind Thattai and Graeme Nail and Gregoire Mialon and Guan Pang and Guillem Cucurell and Hailey Nguyen and Hannah Korevaar and Hu Xu and Hugo Touvron and Iliyan Zarov and Imanol Arrieta Ibarra and Isabel Kloumann and Ishan Misra and Ivan Evtimov and Jack Zhang and Jade Copet and Jaewon Lee and Jan Geffert and Jana Vranes and Jason Park and Jay Mahadeokar and Jeet Shah and Jelmer van der Linde and Jennifer Billock and Jenny Hong and Jenya Lee and Jeremy Fu and Jianfeng Chi and Jianyu Huang and Jiawen Liu and Jie Wang and Jiecao Yu and Joanna Bitton and Joe Spisak and Jongsoo Park and Joseph Rocca and Joshua Johnstun and Joshua Saxe and Junteng Jia and Kalyan Vasuden Alwala and Karthik Prasad and Kartikeya Upasani and Kate Plawiak and Ke Li and Kenneth Heafield and Kevin Stone and Khalid El-Arini and Krithika Iyer and Kshitiz Malik and Kuenley Chiu and Kunal Bhalla and Kushal Lakhotia and Lauren Rantala-Yeary and Laurens van der Maaten and Lawrence Chen and Liang Tan and Liz Jenkins and Louis Martin and Lovish Madaan and Lubo Malo and Lukas Blecher and Lukas Landzaat and Luke de Oliveira and Madeline Muzzi and Mahesh Pasupuleti and Mannat Singh and Manohar Paluri and Marcin Kardas and Maria Tsimpoukelli and Mathew Oldham and Mathieu Rita and Maya Pavlova and Melanie Kambadur and Mike Lewis and Min Si and Mitesh Kumar Singh and Mona Hassan and Naman Goyal and Narjes Torabi and Nikolay Bashlykov and Nikolay Bogoychev and Niladri Chatterji and Ning Zhang and Olivier Duchenne and Onur Çelebi and Patrick Alrassy and Pengchuan Zhang and Pengwei Li and Petar Vasic and Peter Weng and Prajjwal Bhargava and Pratik Dubal and Praveen Krishnan and Punit Singh Koura and Puxin Xu and Qing He and Qingxiao Dong and Ragavan Srinivasan and Raj Ganapathy and Ramon Calderer and Ricardo Silveira Cabral and Robert Stojnic and Roberta Raileanu and Rohan Maheswari and Rohit Girdhar and Rohit Patel and Romain Sauvestre and Ronnie Polidoro and Roshan Sumbaly and Ross Taylor and Ruan Silva and Rui Hou and Rui Wang and Saghar Hosseini and Sahana Chennabasappa and Sanjay Singh and Sean Bell and Seohyun Sonia Kim and Sergey Edunov and Shaoliang Nie and Sharan Narang and Sharath Raparthy and Sheng Shen and Shengye Wan and Shruti Bhosale and Shun Zhang and Simon Vandenhende and Soumya Batra and Spencer Whitman and Sten Sootla and Stephane Collot and Suchin Gururangan and Sydney Borodinsky and Tamar Herman and Tara Fowler and Tarek Sheasha and Thomas Georgiou and Thomas Scialom and Tobias Speckbacher and Todor Mihaylov and Tong Xiao and Ujjwal Karn and Vedanuj Goswami and Vibhor Gupta and Vignesh Ramanathan and Viktor Kerkez and Vincent Gonguet and Virginie Do and Vish Vogeti and Vítor Albiero and Vladan Petrovic and Weiwei Chu and Wenhan Xiong and Wenyin Fu and Whitney Meers and Xavier Martinet and Xiaodong Wang and Xiaofang Wang and Xiaoqing Ellen Tan and Xide Xia and Xinfeng Xie and Xuchao Jia and Xuewei Wang and Yaelle Goldschlag and Yashesh Gaur and Yasmine Babaei and Yi Wen and Yiwen Song and Yuchen Zhang and Yue Li and Yuning Mao and Zacharie Delpierre Coudert and Zheng Yan and Zhengxing Chen and Zoe Papakipos and Aaditya Singh and Aayushi Srivastava and Abha Jain and Adam Kelsey and Adam Shajnfeld and Adithya Gangidi and Adolfo Victoria and Ahuva Goldstand and Ajay Menon and Ajay Sharma and Alex Boesenberg and Alexei Baevski and Allie Feinstein and Amanda Kallet and Amit Sangani and Amos Teo and Anam Yunus and Andrei Lupu and Andres Alvarado and Andrew Caples and Andrew Gu and Andrew Ho and Andrew Poulton and Andrew Ryan and Ankit Ramchandani and Annie Dong and Annie Franco and Anuj Goyal and Aparajita Saraf and Arkabandhu Chowdhury and Ashley Gabriel and Ashwin Bharambe and Assaf Eisenman and Azadeh Yazdan and Beau James and Ben Maurer and Benjamin Leonhardi and Bernie Huang and Beth Loyd and Beto De Paola and Bhargavi Paranjape and Bing Liu and Bo Wu and Boyu Ni and Braden Hancock and Bram Wasti and Brandon Spence and Brani Stojkovic and Brian Gamido and Britt Montalvo and Carl Parker and Carly Burton and Catalina Mejia and Ce Liu and Changhan Wang and Changkyu Kim and Chao Zhou and Chester Hu and Ching-Hsiang Chu and Chris Cai and Chris Tindal and Christoph Feichtenhofer and Cynthia Gao and Damon Civin and Dana Beaty and Daniel Kreymer and Daniel Li and David Adkins and David Xu and Davide Testuggine and Delia David and Devi Parikh and Diana Liskovich and Didem Foss and Dingkang Wang and Duc Le and Dustin Holland and Edward Dowling and Eissa Jamil and Elaine Montgomery and Eleonora Presani and Emily Hahn and Emily Wood and Eric-Tuan Le and Erik Brinkman and Esteban Arcaute and Evan Dunbar and Evan Smothers and Fei Sun and Felix Kreuk and Feng Tian and Filippos Kokkinos and Firat Ozgenel and Francesco Caggioni and Frank Kanayet and Frank Seide and Gabriela Medina Florez and Gabriella Schwarz and Gada Badeer and Georgia Swee and Gil Halpern and Grant Herman and Grigory Sizov and Guangyi and Zhang and Guna Lakshminarayanan and Hakan Inan and Hamid Shojanazeri and Han Zou and Hannah Wang and Hanwen Zha and Haroun Habeeb and Harrison Rudolph and Helen Suk and Henry Aspegren and Hunter Goldman and Hongyuan Zhan and Ibrahim Damlaj and Igor Molybog and Igor Tufanov and Ilias Leontiadis and Irina-Elena Veliche and Itai Gat and Jake Weissman and James Geboski and James Kohli and Janice Lam and Japhet Asher and Jean-Baptiste Gaya and Jeff Marcus and Jeff Tang and Jennifer Chan and Jenny Zhen and Jeremy Reizenstein and Jeremy Teboul and Jessica Zhong and Jian Jin and Jingyi Yang and Joe Cummings and Jon Carvill and Jon Shepard and Jonathan McPhie and Jonathan Torres and Josh Ginsburg and Junjie Wang and Kai Wu and Kam Hou U and Karan Saxena and Kartikay Khandelwal and Katayoun Zand and Kathy Matosich and Kaushik Veeraraghavan and Kelly Michelena and Keqian Li and Kiran Jagadeesh and Kun Huang and Kunal Chawla and Kyle Huang and Lailin Chen and Lakshya Garg and Lavender A and Leandro Silva and Lee Bell and Lei Zhang and Liangpeng Guo and Licheng Yu and Liron Moshkovich and Luca Wehrstedt and Madian Khabsa and Manav Avalani and Manish Bhatt and Martynas Mankus and Matan Hasson and Matthew Lennie and Matthias Reso and Maxim Groshev and Maxim Naumov and Maya Lathi and Meghan Keneally and Miao Liu and Michael L. Seltzer and Michal Valko and Michelle Restrepo and Mihir Patel and Mik Vyatskov and Mikayel Samvelyan and Mike Clark and Mike Macey and Mike Wang and Miquel Jubert Hermoso and Mo Metanat and Mohammad Rastegari and Munish Bansal and Nandhini Santhanam and Natascha Parks and Natasha White and Navyata Bawa and Nayan Singhal and Nick Egebo and Nicolas Usunier and Nikhil Mehta and Nikolay Pavlovich Laptev and Ning Dong and Norman Cheng and Oleg Chernoguz and Olivia Hart and Omkar Salpekar and Ozlem Kalinli and Parkin Kent and Parth Parekh and Paul Saab and Pavan Balaji and Pedro Rittner and Philip Bontrager and Pierre Roux and Piotr Dollar and Polina Zvyagina and Prashant Ratanchandani and Pritish Yuvraj and Qian Liang and Rachad Alao and Rachel Rodriguez and Rafi Ayub and Raghotham Murthy and Raghu Nayani and Rahul Mitra and Rangaprabhu Parthasarathy and Raymond Li and Rebekkah Hogan and Robin Battey and Rocky Wang and Russ Howes and Ruty Rinott and Sachin Mehta and Sachin Siby and Sai Jayesh Bondu and Samyak Datta and Sara Chugh and Sara Hunt and Sargun Dhillon and Sasha Sidorov and Satadru Pan and Saurabh Mahajan and Saurabh Verma and Seiji Yamamoto and Sharadh Ramaswamy and Shaun Lindsay and Shaun Lindsay and Sheng Feng and Shenghao Lin and Shengxin Cindy Zha and Shishir Patil and Shiva Shankar and Shuqiang Zhang and Shuqiang Zhang and Sinong Wang and Sneha Agarwal and Soji Sajuyigbe and Soumith Chintala and Stephanie Max and Stephen Chen and Steve Kehoe and Steve Satterfield and Sudarshan Govindaprasad and Sumit Gupta and Summer Deng and Sungmin Cho and Sunny Virk and Suraj Subramanian and Sy Choudhury and Sydney Goldman and Tal Remez and Tamar Glaser and Tamara Best and Thilo Koehler and Thomas Robinson and Tianhe Li and Tianjun Zhang and Tim Matthews and Timothy Chou and Tzook Shaked and Varun Vontimitta and Victoria Ajayi and Victoria Montanez and Vijai Mohan and Vinay Satish Kumar and Vishal Mangla and Vlad Ionescu and Vlad Poenaru and Vlad Tiberiu Mihailescu and Vladimir Ivanov and Wei Li and Wenchen Wang and Wenwen Jiang and Wes Bouaziz and Will Constable and Xiaocheng Tang and Xiaojian Wu and Xiaolan Wang and Xilun Wu and Xinbo Gao and Yaniv Kleinman and Yanjun Chen and Ye Hu and Ye Jia and Ye Qi and Yenda Li and Yilin Zhang and Ying Zhang and Yossi Adi and Youngjin Nam and Yu and Wang and Yu Zhao and Yuchen Hao and Yundi Qian and Yunlu Li and Yuzi He and Zach Rait and Zachary DeVito and Zef Rosnbrick and Zhaoduo Wen and Zhenyu Yang and Zhiwei Zhao and Zhiyu Ma},
      year={2024},
      eprint={2407.21783},
      archivePrefix={arXiv},
      primaryClass={cs.AI},
      url={https://arxiv.org/abs/2407.21783}, 
}

@misc{gemmateam2025gemma3technicalreport,
      title = {{{Gemma 3} Technical Report}}, 
      author={Gemma and Aishwarya Kamath and Johan Ferret and Shreya Pathak and Nino Vieillard and Ramona Merhej and Sarah Perrin and Tatiana Matejovicova and Alexandre Ramé and Morgane Rivière and Louis Rouillard and Thomas Mesnard and Geoffrey Cideron and Jean-bastien Grill and Sabela Ramos and Edouard Yvinec and Michelle Casbon and Etienne Pot and Ivo Penchev and Gaël Liu and Francesco Visin and Kathleen Kenealy and Lucas Beyer and Xiaohai Zhai and Anton Tsitsulin and Robert Busa-Fekete and Alex Feng and Noveen Sachdeva and Benjamin Coleman and Yi Gao and Basil Mustafa and Iain Barr and Emilio Parisotto and David Tian and Matan Eyal and Colin Cherry and Jan-Thorsten Peter and Danila Sinopalnikov and Surya Bhupatiraju and Rishabh Agarwal and Mehran Kazemi and Dan Malkin and Ravin Kumar and David Vilar and Idan Brusilovsky and Jiaming Luo and Andreas Steiner and Abe Friesen and Abhanshu Sharma and Abheesht Sharma and Adi Mayrav Gilady and Adrian Goedeckemeyer and Alaa Saade and Alex Feng and Alexander Kolesnikov and Alexei Bendebury and Alvin Abdagic and Amit Vadi and András György and André Susano Pinto and Anil Das and Ankur Bapna and Antoine Miech and Antoine Yang and Antonia Paterson and Ashish Shenoy and Ayan Chakrabarti and Bilal Piot and Bo Wu and Bobak Shahriari and Bryce Petrini and Charlie Chen and Charline Le Lan and Christopher A. Choquette-Choo and CJ Carey and Cormac Brick and Daniel Deutsch and Danielle Eisenbud and Dee Cattle and Derek Cheng and Dimitris Paparas and Divyashree Shivakumar Sreepathihalli and Doug Reid and Dustin Tran and Dustin Zelle and Eric Noland and Erwin Huizenga and Eugene Kharitonov and Frederick Liu and Gagik Amirkhanyan and Glenn Cameron and Hadi Hashemi and Hanna Klimczak-Plucińska and Harman Singh and Harsh Mehta and Harshal Tushar Lehri and Hussein Hazimeh and Ian Ballantyne and Idan Szpektor and Ivan Nardini and Jean Pouget-Abadie and Jetha Chan and Joe Stanton and John Wieting and Jonathan Lai and Jordi Orbay and Joseph Fernandez and Josh Newlan and Ju-yeong Ji and Jyotinder Singh and Kat Black and Kathy Yu and Kevin Hui and Kiran Vodrahalli and Klaus Greff and Linhai Qiu and Marcella Valentine and Marina Coelho and Marvin Ritter and Matt Hoffman and Matthew Watson and Mayank Chaturvedi and Michael Moynihan and Min Ma and Nabila Babar and Natasha Noy and Nathan Byrd and Nick Roy and Nikola Momchev and Nilay Chauhan and Noveen Sachdeva and Oskar Bunyan and Pankil Botarda and Paul Caron and Paul Kishan Rubenstein and Phil Culliton and Philipp Schmid and Pier Giuseppe Sessa and Pingmei Xu and Piotr Stanczyk and Pouya Tafti and Rakesh Shivanna and Renjie Wu and Renke Pan and Reza Rokni and Rob Willoughby and Rohith Vallu and Ryan Mullins and Sammy Jerome and Sara Smoot and Sertan Girgin and Shariq Iqbal and Shashir Reddy and Shruti Sheth and Siim Põder and Sijal Bhatnagar and Sindhu Raghuram Panyam and Sivan Eiger and Susan Zhang and Tianqi Liu and Trevor Yacovone and Tyler Liechty and Uday Kalra and Utku Evci and Vedant Misra and Vincent Roseberry and Vlad Feinberg and Vlad Kolesnikov and Woohyun Han and Woosuk Kwon and Xi Chen and Yinlam Chow and Yuvein Zhu and Zichuan Wei and Zoltan Egyed and Victor Cotruta and Minh Giang and Phoebe Kirk and Anand Rao and Kat Black and Nabila Babar and Jessica Lo and Erica Moreira and Luiz Gustavo Martins and Omar Sanseviero and Lucas Gonzalez and Zach Gleicher and Tris Warkentin and Vahab Mirrokni and Evan Senter and Eli Collins and Joelle Barral and Zoubin Ghahramani and Raia Hadsell and Yossi Matias and D. Sculley and Slav Petrov and Noah Fiedel and Noam Shazeer and Oriol Vinyals and Jeff Dean and Demis Hassabis and Koray Kavukcuoglu and Clement Farabet and Elena Buchatskaya and Jean-Baptiste Alayrac and Rohan Anil and Dmitry and Lepikhin and Sebastian Borgeaud and Olivier Bachem and Armand Joulin and Alek Andreev and Cassidy Hardin and Robert Dadashi and Léonard Hussenot},
      year={2025},
      eprint={2503.19786},
      archivePrefix={arXiv},
      primaryClass={cs.CL},
      url={https://arxiv.org/abs/2503.19786}, 
}

@article{young2026measuring,
  title = {{Measuring Faithfulness Depends on How You Measure: Classifier Sensitivity in {LLM} Chain-of-Thought Evaluation}},
  author  = {Young, Richard J.},
  journal = {arXiv preprint arXiv:2603.20172},
  year    = {2026},
  url     = {https://arxiv.org/abs/2603.20172}
}
